\documentclass[lettersize,journal]{IEEEtran}

\PassOptionsToPackage{numbers, compress}{natbib}

\usepackage[utf8]{inputenc} %
\usepackage[T1]{fontenc}    %
\usepackage{hyperref}       %
\usepackage{url}            %
\usepackage{booktabs}       %
\usepackage{amsfonts}       %
\usepackage{nicefrac}       %
\usepackage{microtype}      %
\usepackage{xcolor}         %
\usepackage{graphicx}       %
\usepackage{tabularx}       %
\usepackage{ragged2e}       %
\usepackage{cite}
\usepackage{amsmath}        %
\usepackage{makecell}       %
\usepackage{multirow}
\usepackage{adjustbox}      %
\usepackage{rotating}
\usepackage{cellspace}
\usepackage{nicematrix}
\usepackage{booktabs}
\usepackage{xparse}
\usepackage{placeins}
\usepackage{tikz}

\usepackage{algorithmic}
\usepackage{array}
\usepackage[caption=false,font=normalsize,labelfont=sf,textfont=sf]{subfig}
\usepackage{textcomp}
\usepackage{stfloats}
\usepackage{verbatim}
\def\BibTeX{{\rm B\kern-.05em{\sc i\kern-.025em b}\kern-.08em
    T\kern-.1667em\lower.7ex\hbox{E}\kern-.125emX}}
\usepackage{balance}

\usetikzlibrary{calc}

\newcolumntype{C}{>{\Centering\arraybackslash}X} %

\newcommand{\unaryminus}{\scalebox{1.0}[1.0]{\($-$\)}} %

\makeatletter
\renewcommand\@makefntext[1]{\leftskip=2em\hskip-2em\@makefnmark#1}
\makeatother

\newcolumntype{R}[2]{%
    >{\adjustbox{angle=#1,lap={1ex}{\width-#2}}\bgroup}%
    l%
    <{\egroup}%
}

\title{The BUTTER Zone: An Empirical Study of Training Dynamics in Fully Connected Neural Networks}

\author{
\IEEEmembership{Member, IEEE}, Charles Edison Tripp,
Jordan Perr-Sauer,
Lucas Hayne,
Monte Lunacek,
Jamil Gafur
}

\begin{document}

\maketitle

\begin{abstract}

We present an empirical dataset surveying the deep learning phenomenon on fully-connected feed-forward multilayer perceptron neural networks.
The dataset, which is now freely available online, records the per-epoch training and generalization performance of 483 thousand distinct hyperparameter choices of architectures, tasks, depths, network sizes (number of parameters), learning rates, batch sizes, and regularization penalties.
Repeating each experiment an average of 24 times resulted in 11 million total training runs and 40 billion epochs recorded.
Accumulating this 1.7 TB dataset utilized 11 thousand CPU core-years, 72.3 GPU-years, and 163 node-years.
In surveying the dataset, we observe durable patterns persisting across tasks and topologies.
We aim to spark scientific study of machine learning techniques as a catalyst for the theoretical discoveries needed to progress the field beyond energy-intensive and heuristic practices.

\end{abstract}


\section{Introduction}
\label{sec:Introduction}

\IEEEPARstart{H}{istorically}, the governing dynamics of machine learning methods have been derived from first principles analysis.
However, increasingly complex methods, such as deep learning, have proven resistant to such theoretical analysis.
Without substantial insight into the governing dynamics of deep learning, practitioners are left with trial-and-error heuristics and energy-intensive hyperparameter optimization schemes.
We hypothesize that deep learning methods have attained sufficient complexity to justify their empirical study as a complex computational phenomena, and that scientific study of the behavior of deep learning can provide a guiding light for theoretical analysis.
In this spirit, we collected and present an empirical dataset, which we name ``BUTTER'', surveying the deep learning phenomenon on fully-connected feed-forward multilayer perceptron neural networks (FC MLPs), encompassing the training and test performance of network topologies sweeping across 13 learning tasks, 8 network shapes, 14 depths, 23 parameter counts, 4 learning rates, 6 batch sizes, 4 label noise levels, 27 regularization penalties, 3 optimizers, and two momentum levels.
The dataset tests 483 thousand distinct experiments with an average of 24 repetitions each, yielding 11 million training runs covering 20 performance metrics for each of the 40 billion training epochs observed.
Accumulating this 1.7 TB dataset utilized 11 thousand CPU core-years, 72.3 GPU-years, and 163 node-years across four high-performance computing (HPC) systems.

BUTTER studies MLPs as they are a fundamental starting point for deep learning research.
As discussed in Section \ref{sec:background}, an empirical deep learning dataset of BUTTER's scale is novel, even for MLPs.
As findings about MLPs can inform investigations into other other salient network classes including convolutional neural networks (CNNs), Transformers, recurrent neural networks (RNNs), and graph neural networks (GNNs), BUTTER forms a guide-stone for generating and analyzing empirical datasets studying such architectures, which could extend the BUTTER dataset in the future.
We hope this analysis contributes to a call-to-arms for researchers to dig deeper and to conduct their own empirical studies to gain a better understanding of this computational phenomenon and to improve deep learning engineering practices.

\section{Background}
\label{sec:background}

\noindent We present this dataset to promote investigation into the expressivity, generalization, and trainability of deep learning models, three properties identified by Chen et al. \cite{chen2021understanding} as important for successful deep learning.
Early researchers focused primarily on expressivity, proving that FC MLPs serve as universal approximators \cite{hornik1989multilayer} and bounded the width of networks needed to approximate certain functions \cite{barron1993universal}.
Inspired by practical results suggesting that MLPs can overcome the curse of dimensionality, recent theoretical work identified scaling law bounds for rates of convergence when training on restricted forms of binary classification \cite{zhang2017learnability} and regression
\cite{kohler2021rate, LANGER2021104695}. 
The neural tangent kernel (NTK) provides an analytic expression for the first order approximation of training dynamics for single-layer MLPs in the infinite-width case \cite{jacot_neural_2020}.
Other recent theoretical results include proving that functions can be constructed that are expressible by 3-layer MLPs that are not well approximated by 2-layer MLPs with similar sizes \cite{eldan2016power}, Kolmogorov-optimal approximation bounds for MLPs using infinite training data \cite{elbrachter2021deep}, lower bounds on the error of ReLU MLP networks for smooth functions as a function of width and depth \cite{lu2021deep}, a proof that CNNs can outperform one-layer MLPs \cite{deep_vs_shallow_theory}, a proof that rectangular ReLU FC MLPs with a width proportional to their number of weights can achieve an approximation rate for which cannot be attained by shallower networks \cite{yarotsky2018optimal}, and bounds on the approximation rates of polynomial and periodic activation function based MLPs \cite{yarotsky2020phase}.
Theories like these outline what is possible in deep learning, but practitioners need reliable predictions of how their design choices will affect their model training and generalization given their practical constraints.
A practitioner might ask \textit{``If I train this network on my dataset with these hyperparameters, what test loss will it achieve and how many epochs will it take?''}

Recent empirical work has begun to address this challenge.
Pezeshki et al. \cite{pezeshki2021multi} analytically reproduce double descent curves at various regularization strengths and posit the existence of slow and fast features that are learned at different rates.
Chen et al. \cite{chen2021multiple} show that for linear regression, the distribution of new features added to the training set determines the location of descents and ascents in the generalization error curve.
Adlam et al. \cite{adlam2020quadratic} present a mathematical foundation for the double descent phenomenon using random matrix theory and predict a complex relationship between the size and generalizability of a model.
In 2015 and elaborated in 2017, Neyshabur et al. presented empirical evidence that the classical u-shaped bias-variance trade off \cite{geman1992_bias_variance_in_nn, hastie2009elements} is not exactly observed in deep learning \cite{neyshabur2015_deep_descent_evidence, neyshabur2017_deep_descent_evidence}.
Neal et al. \cite{neal2018_double_descent} and Belkin et al. \cite{belkin_2019_double_descent} discussed empirical and theoretical evidence that the classical bias-variance trade-off may not apply to overparameterized models; beyond the interpolation threshold, we may observe paradoxically decreasing generalization error.
In 2019 and 2021 Nakkiran et al. presented empirical evidence of double descent in terms of number of parameters and epochs \cite{nakkiran2019deep, nakkiran2021deep}.

\begin{table*}
{
    \centering
    \caption{
    \textbf{A comparison of BUTTER to related datasets.}
    }
    \label{tab:datasets}
    \centering
    {
    \begin{minipage}{\textwidth}
    \renewcommand\footnoterule{}     %
    \centering
    \begin{tabular}{lrrrrrrrrrc}
        Dataset & 
        Experiments  &
        Repetitions  &
        \makecell{Tasks}  &
        \makecell{Epochs}   &
        \makecell{Width\\Range}  &
        \makecell{Depth\\Range}  &
        \makecell{Batch\\Sizes}  &
        \makecell{Learning\\Rates}  &
        \makecell{Regularization\\Levels}  &
        \makecell{Network\\Class}  \\%
        \midrule
        \textbf{BUTTER} \rule{0pt}{2.4ex} & \textbf{483k} & \textbf{30} & \textbf{13} & \textbf{3k}\footnote{\scriptsize Experiments in BUTTER run for 3000 epochs, except the 30k and 300 epoch runs as indicated in Table \ref{tab:sweeps}.} & \textbf{1-7M} & \textbf{2-20} & \textbf{6} & \textbf{4} & 31 & \textbf{MLP}\\
        NAS-HPO-Bench \cite{NAS_HPO_Bench}  & 62k & 4 & 1 & 100 &  16-512 & 2 & 4 & 1 & 1 & MLP\\ 
        NAS-HPO-Bench-II \cite{hirose2021hpo} & 192k & 3 & 1 & 12 & 64 & 15 &  6 & 8 & 1 & CNN \\
        LCBench \cite{LCBench} & 2k & 1 & 35 & 12-50 & 64-1.0k & 1-5 & n/a\footnote{\scriptsize LCBench randomly sampled batch sizes, learning rates, and regularization levels.} & n/a & n/a & MLP\\
        NATS-Bench \cite{Dong2020NAS-Bench-201, Dong2022NATSBenchBN} & 6k & 1 & 3 & 200 & 8-16 & 15 & 1 & 1 & 1 & CNN\\ 
        NAS-Bench-101 \cite{ying2019bench} & 423k & 3 & 1 & 108 & 128 & 9 & 1 & 1 & 1 & CNN \\ 
        NAS-Bench-ASR \cite{mehrotra2021nasbenchasr} & 8k & 3 & 1 & 40 & 600-1.2k & 12-24 & 2 & 1 & 1 & LSTM\\
        TransNAS-Bench \cite{trans-nas-bench} & 7k  & 1 & 7 & 100 &  64 & 8-12 & 1 & 1 & 1 & CNN \\
         \midrule
    \end{tabular}
    \vspace{-2ex}
    \end{minipage}
    }
}
\end{table*}

In 2020, Kaplan et al. empirically studied scaling laws of neural language models given unbounded training data \cite{kaplan2020_scaling_laws} and Henighan et al. empirically studied scaling laws of Autoregressive Generative Modeling \cite{henighan2020_scaling_laws}.
In 2022, Tay et al. derived scaling laws across inductive biases and model architectures, and summarized many scaling law findings \cite{tay2022_scaling_laws}.
Although focused on the unbounded data regime, empirical studies of scaling laws suggest performance scales as a power-law with dataset size, training effort, and parameter count.
Further, these studies suggest that width, depth, and model shape have only small impacts on performance.
Our dataset corroborates these findings for MLPs and finite datasets.

Universal approximation bounds and the double deep descent hypothesis relate relatively simple measures (layer width and effective model complexity) to a model's generalization but, in practice, researchers are concerned with many more architectural choices than these two.
Jiang et al. \cite{jiang2019fantastic} conducted an empirical study of complexity measures, or ``properties of the trained model, optimizer, and possibly training data'' that ``monotonically relate to some aspect of generalization.''
Across 10,000 trained models, they found promising PAC-Bayes and optimization-based metrics for predicting generalization performance.
Dziugaite et al. \cite{dziugaite2020search} conducted a similar study seeking distributionally robust generalization measures. 
Additionally, Novak et al.. \cite{novak2018sensitivity} describe a sensitivity metric based on the norm of the Jacobian of the loss function, and find that robustness in this metric around training data is predictive of generalizability. 
To help explain which networks can be trained to achieve good generalization scores (trainability), Frankle and Carbin \cite{frankle2018lottery} established the lottery ticket hypothesis (LTH) from empirical study.
This hypothesis suggests that ``SGD seeks out and trains a well-initialized subnetwork'' and that ``overparameterized networks are easier to train because they have more combinations of subnetworks that are potential winning tickets.''

To enable offline benchmarking of Neural Architecture Search (NAS) algorithms, several related datasets have been generated.
Table \ref{tab:datasets} compares the BUTTER dataset to seven other NAS benchmarking datasets that provide raw ``tabular'' results (leaving out benchmarks which only provide a surrogate model and no raw results) \cite{mehta2022bench}.
Tabular NAS benchmarks exist for image classification \cite{ying2019bench,dong2020bench,zela2020bench,dong2021nats,su2021prioritized,wu2019fbnet}, automatic speech recognition \cite{mehrotra2020bench}, and other tasks \cite{duan2021transnas}.
In contrast to the goals of NAS research, we present a large empirical dataset to stimulate discoveries of patterns underlying the expressivity, generalization, and trainability of deep learning models.
To this end, BUTTER contains more experiments than the NAS datasets, and repeats each experiment using unique random seeds an order of magnitude more times.
Repetitions allow us to access information about convergence and performance variability, enabling dataset consumers to study the stochastic nature of deep learning including the LTH.
Ignoring the variability of the training process exposes users to the risk of spurious and incorrect interpretations due to sample bias.
In some cases, BUTTER experiments minimize test loss far beyond that which is achieved in the 12 to 200 epochs covered by other datasets.
The long training runs in BUTTER (one to two orders of magnitude more epochs than other datasets) allow us to observe the scaling laws of over-fitting \cite{kaplan2020_scaling_laws}, to escape local minima in the training curve, and to probe for deep double descent \cite{nakkiran2019deep}.
Large detailed sweeps over the configuration space enables dataset consumers to analyze scaling laws as a function of these parameters and to determine how these laws might generalize across tasks.
Finally, BUTTER covers 31 regularization levels, a dimension not studied in any of the comparable datasets listed in Table \ref{tab:datasets}.

\section{Contributions}

\label{sec:Contributions}
\begin{enumerate}

    \item We present \href{https://data.openei.org/submissions/5708}{BUTTER}, an empirical deep learning dataset recording 20 performance metrics at each epoch of an average of 24 repetitions of 483 thousand distinct experiments totalling 11 million training runs and 40 billion epochs. Accumulating this 1.7 TB dataset utilized 11 thousand CPU core-years, 72.3 GPU-years, and 163 node-years.
    We describe this dataset in Section \ref{sec:Dataset} and make it freely available under the  \href{https://creativecommons.org/licenses/by-sa/4.0/}{Creative Commons Attribution-ShareAlike 4.0 International Public License} \cite{BUTTERDataset}.
    \item We describe the dataset with tables enumerating the hyperparameter dimensions in section \ref{sec:Dataset}. We visualize and survey the dataset through the use of partial dependence plots of the marginals for each hyperparameter in Section \ref{sec:observations}, making observations and drawing comparisons to findings in related work cited above. We observe certain trends which persist in the data across tasks and topologies, and present these findings in a preliminary analysis in Appendix \ref{sec:analysis}.
    \item We make the \href{https://github.com/NREL/BUTTER-Empirical-Deep-Learning-Experimental-Framework}{source code for our empirical machine learning framework} \cite{EmpiricalFramework}, \href{https://github.com/NREL/E-Queue-HPC}{distributed job queue} \cite{EQueue} which distributes the experiments over as many systems as desired, and \href{https://github.com/NREL/BUTTER-Better-Understanding-of-Training-Topologies-through-Empirical-Results}{the code to reproduce our analysis figures} freely available under the MIT Licence. \footnote{\href{https://github.com/NREL/BUTTER-Better-Understanding-of-Training-Topologies-through-Empirical-Results}{github.com/NREL/BUTTER-Better-Understanding-of-Training-Topologies-through-Empirical-Results}}
\end{enumerate}

\begin{table*}
{
    \caption{
    \textbf{A summary of the sweeps composing the dataset.} 
    To conserve space, the bounds of each range are listed.
    Salient differences from the Primary Sweep are shown in \textbf{bold}.
    Task \#'s and shape \#'s are indexed in Tables \ref{tab:tasks} and \ref{tab:shapes}.
    }
    \label{tab:sweeps}
    \centering
    \begin{minipage}{\textwidth}
    \centering
    \renewcommand\footnoterule{}     %
    \begin{tabular}{l*{9}r}
        & \multicolumn{8}{c}{Sweep} \\ \cmidrule{2-9}
        Attribute & {Primary} & {300 Epoch} & {30k Epoch} & {Learning Rate} & {Label Noise} & {Batch Size} & {Regularization} & {Optimizer} & \textbf{Overall} \\
         \midrule
         Experiments & 38.8k & 2.88k & 22.9k & 59.5k & 102k & 15.9k & 79.1k & 215k & \textbf{483k}\\
         Runs & 1.47M & 28.8k & 239k & 1.32M & 3.69M & 223k & 850k & 4.96M & \textbf{11.2M}\\ 
         Total Epochs & 4.42B & 4.23M & 7.17B & 3.96B & 11.1B & 669M & 14.9B & 9.81B & \textbf{40.0B}\\
         Repetitions & 30 & 10 & 10 & 20 & 30 & 10 & 20 & 30 & \textbf{10..30} \\
         Epochs / Run & 3k & \textbf{300} & \textbf{30k} & 3k & 3k & 3k & 3k & 3k &  \textbf{300..30k}\\
         Task \# & $1..13$& $\mathbf{1..9}$& $\mathbf{1..9}$& $\mathbf{1..9}$ &  $\mathbf{1..7}$ & $\mathbf{1..9}$ & $\mathbf{1..9}$ & $1..13$ & $\mathbf{1..13}$\\
         Shape \# & $1..8$ & $\mathbf{1..6}$ & $\mathbf{1..6}$ & $\mathbf{1..8}$& $\mathbf{1..4}$& $\mathbf{1..4}$& $\mathbf{1}$& $\mathbf{1}$ & $\mathbf{1..8}$\\
         \# Parameters & $2^{5..25}$ & $\mathbf{2^{25..27}}$ & $\mathbf{2^{5..18}}$& $2^{5..24}$& $2^{5..24}$& $2^{5..24}$& $2^{5..24}$& $2^{5..25}$ & $\mathbf{2^{5..27}}$\\
         Depth & $2..20$& $\mathbf{2..10}$& $2..20$& $2..20$& $2..20$& $2..10$& $2..10$& $2..6$ & $\mathbf{2..20}$\\
         Learning Rate & $10^{\unaryminus 4}$& $10^{\unaryminus 4}$& $10^{\unaryminus 4}$& $\mathbf{10^{\unaryminus 2..\unaryminus 5}}$& $\mathbf{10^{\unaryminus 4..\unaryminus 5}}$& $10^{\unaryminus 4}$& $10^{\unaryminus 4}$ & $\mathbf{10^{\unaryminus 2..\unaryminus 5}}$& $\mathbf{10^{\unaryminus 2..\unaryminus 5}}$\\
         Batch Size & $2^8$ & $2^8$ & $2^8$ & $2^8$ &  $2^8$& $\mathbf{2^{5..10}}$ & $2^8$ & $\mathbf{2^{5..8}}$ & $\mathbf{2^{5..10}}$\\
         Label Noise & $0$& $0$& $0$& $0$& \textbf{$\mathbf{0 .. 0.2}$} & $0$ & $0$& $0$&  \textbf{$\mathbf{0 .. 0.2}$}\\
         Regularization & $0$ & $0$& $0$& $0$& $0$& $0$& {$\mathbf{0 .. 0.32}$} & $0$ &  {$\mathbf{0 .. 0.32}$}\\
         Momentum & $0$ & $0$ & $0$ & $0$ & $0$ & $0$ & $0$ & $\mathbf{0, 0.9}$ & $\mathbf{0, 0.9}$ \\
         Optimizer & 1 & 1 & 1 & 1 & 1 & 1 & 1 & \makecell[r]{\textbf{1, 2, 3}} & \makecell[r]{\textbf{1, 2, 3}} \footnote{\scriptsize Optimizers are numbered (1) Adam, (2) SGD, and (3) RMSProp}\\
         \midrule
    \end{tabular}
    \end{minipage}
}
\end{table*}

\section{Description of the BUTTER dataset}
\label{sec:Dataset}

\begin{table}
    \caption{\textbf{The network shapes tested.}
    Examples indicate layer widths from first to last hidden layer.
    }
    \label{tab:shapes}
    \centering
    \setlength\tabcolsep{0.8mm}
    \begin{tabular}{llll}
        \midrule
         \# & Name & Example Widths & Description\\
         \midrule
         1 & rectangle & $[32,32,32,32]$ & All layers are the same width. \\
         2 & trapezoid & $[50,40,30,20]$ & Widths decrease linearly with depth. \\
         3 & exponential & $[128,64,32,16]$ & Widths decrease exponentially. \\
         4 & wide\_first\_2x &$[64,32,32,32]$ & Rectangle + 2x-wide first layer. \\
         5 & wide\_first\_4x &$[128,32,32,32]$ & Rectangular + 4x-wide first layer. \\
         6 & wide\_first\_8x & $[256,32,32,32]$ & Rectangular + 8x-wide first layer. \\
         7 & wide\_first\_16x & $[512,32,32,32]$ & Rectangular + 16x-wide first layer. \\
         8 & rectangle\_residual & $[32,32,32,32]$ & Rectangular + residual connections. \\
         \midrule
    \end{tabular}
\end{table}

\begin{table}
    \caption{\textbf{The learning tasks used to train networks in the dataset.}}
    \label{tab:tasks}
    \centering
    \setlength\tabcolsep{0.9mm}
    \begin{tabular}{rllrrr}
        \midrule
         Task \# & Name & Type & Observations & Features & Classes\\
         \midrule
         1 & 201\_pol & regression & 15000 & 48 & - \\
         2 & 529\_pollen & regression & 3848 & 4 & - \\
         3 & 537\_houses & regression & 20640 & 8 & - \\
         4 & connect\_4 & classification & 67557 & 42 & 3 \\
         5 & mnist & classification & 70000 & 784 & 10 \\
         6 & sleep & classification & 105908 & 13 & 5 \\
         7 & wine\_quality\_white & classification & 4898 & 11 & 7 \\
         8 & adult & classification & 48842 & 14 & 2 \\
         9 & nursery & classification & 12958 & 8 & 4 \\
         10 & 294\_satellite\_image & regression & 6435 & 36 & - \\
         11 & 505\_tecator & regression & 240 & 124 & - \\
         12 & banana & regression & 5300 & 2 & - \\
         13 & splice & classification & 3188 & 60 & 3 \\
         \midrule
    \end{tabular}
\end{table}

\begin{figure*}[t]
    \centering
    \includegraphics[width=\linewidth]{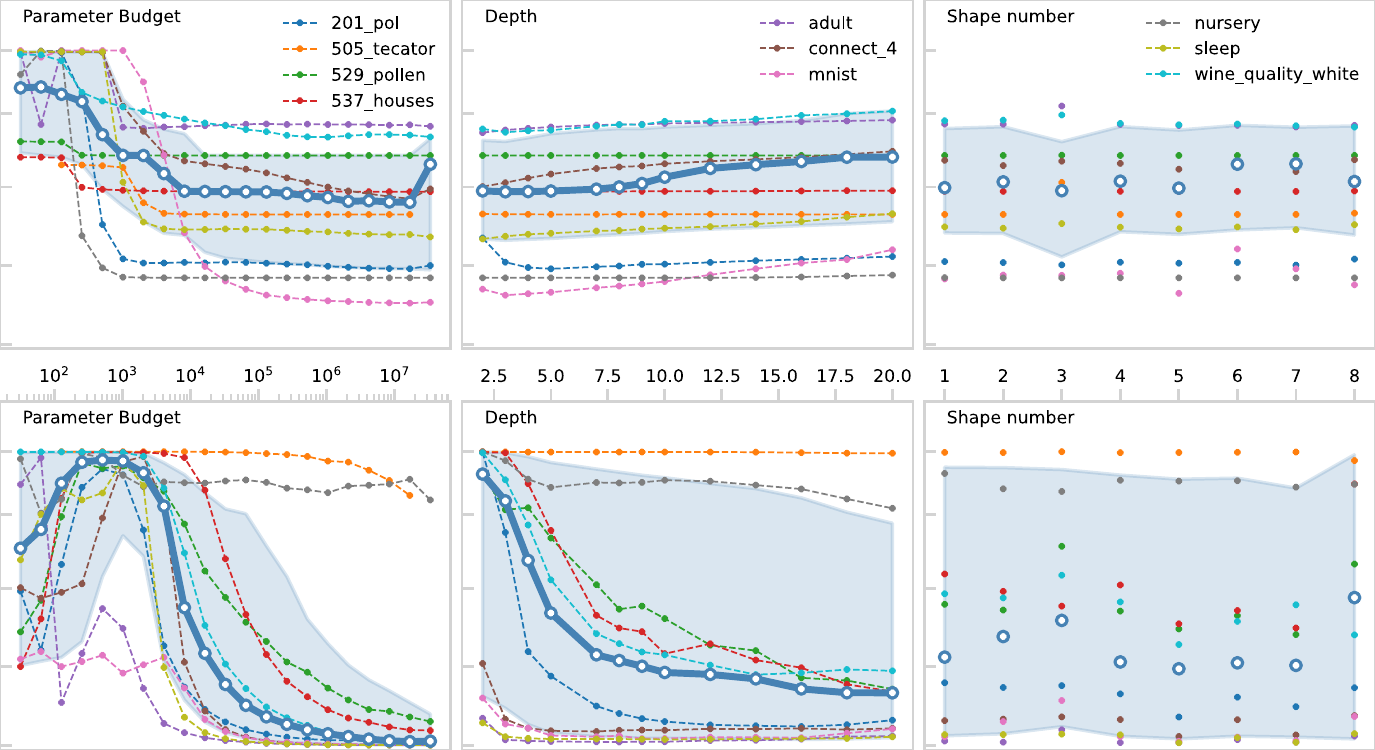}
    \caption{\textbf{Marginal effects plots for number of parameters, depth, and shape on the Primary Sweep.}
    These marginals plots show the median minimum test loss achieved and the median number of epochs to reach it (y-axes) on a per-task (dashed lines) and overall (thick line) basis as a function of a particular hyperparameter (x-axis).
    The overall interquartile range is shaded to indicate how much variation in performance might be due to other hyperparameter choices.
    A thin variation band indicates that the hyperparameter value tightly controls performance and other parameter settings have little impact.
    A wide band indicates that the setting's effect is not robust and performance may be largely dictated by other factors.
    Shape numbers are indexed in Table \ref{tab:shapes}.
    }
    \label{fig:marginals1}
\end{figure*}

\begin{figure}
    \centering
    \includegraphics[width=3in]{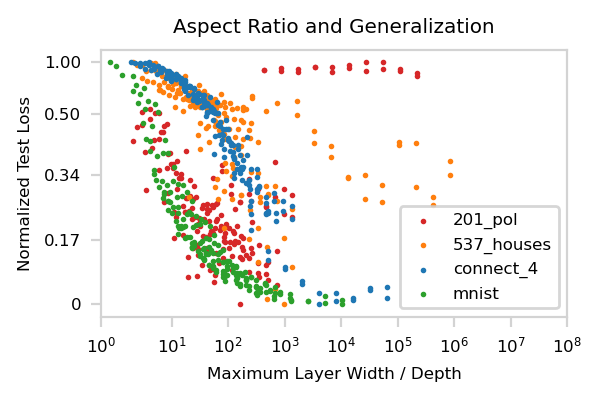}
    \caption{\textbf{Generalization performance depends on the aspect ratio}. We trained networks with varying widths and depths to study the effects of these model architecture choices on generalization. 
    Test losses are min-max normalized and log scaled for each task in the Primary Sweep.
    }
    \label{fig:aspect_ratio}
\end{figure}

\noindent The BUTTER dataset consists of seven hyperparameter sweeps over the MLP design space, detailed in Table \ref{tab:sweeps}.
Each sweep is designed to study the impacts of one or more design choices on training and model performance.
We repeat each experiment multiple times to capture variations in performance due to the stochastic initialization and optimization.
Each repetition of an experiment used a different random seed, and the seed, software versions, and environment used are recorded in the dataset to ensure reproducibility.
We refer to each distinct combination of parameters as an ``experiment'' and each repetition of an experiment as a ``repetition''.
Due to overlaps between sweeps, the total number of experiments and repetitions in the dataset is less than the sums of these quantities over all sweeps.
Network shapes are listed in Table \ref{tab:shapes} and define how the width of each layer changes with depth.
Learning tasks were drawn from the Penn Machine Learning Benchmarks (PMLB) \cite{PMLB} because PMLB provides a single source MIT Licensed library with a simple API enabling us to study a wide variety of tasks.\footnote{See https://epistasislab.github.io/pmlb/ for detailed summaries of PMLB datasets, including those used to generate this dataset.} 
We chose the tasks indexed in Table \ref{tab:tasks} to test both classification (7 tasks) and regression (6 tasks) tasks covering a range of observation (sleep's 106k to 505\_tecator's 240 observations) and feature (banana's 2 to mnist's 784 features) counts.
By drawing this variety of tasks from PMLB, we aim for patterns identified in our dataset to generalize to a meaningful range of tasks, and to increase the possibility of identifying dynamics specific to particular dataset types. 
Our dataset does not include any offensive or personally identifiable information.
The exact topology used in an experiment is explicitly stored, but is defined by the combination of shape, depth, and number of trainable parameters.
Task data was randomly shuffled  on every repetition using an 80\%/20\% training-test split.
The Adaptive Moment Estimation optimizer (Adam) \cite{kingma2014adam} was used as the optimizer for all runs outside of the optimizer sweep, and rectified linear unit (ReLU) activation functions \cite{ReLU1, ReLU2} were used for all hidden layers in all networks.
We used Adam because it has proven effective across many network configurations, choice of learning rate, and tasks due to its step-size adaptation.
Adam occupies a middle ground between Momentum, Root Mean Square Propagation (RMSProp) \cite{rmsprop1, rmsprop2}, and AdaGrad \cite{adagrad} optimizers.
Adam enjoys a long tenure as a popular optimizer and has garnered over 11k citations \cite{karpathy2017peek}.

\begin{figure*}
    \centering
    \makebox[\textwidth][c]{
        \centering
        \includegraphics[width=3.76in]{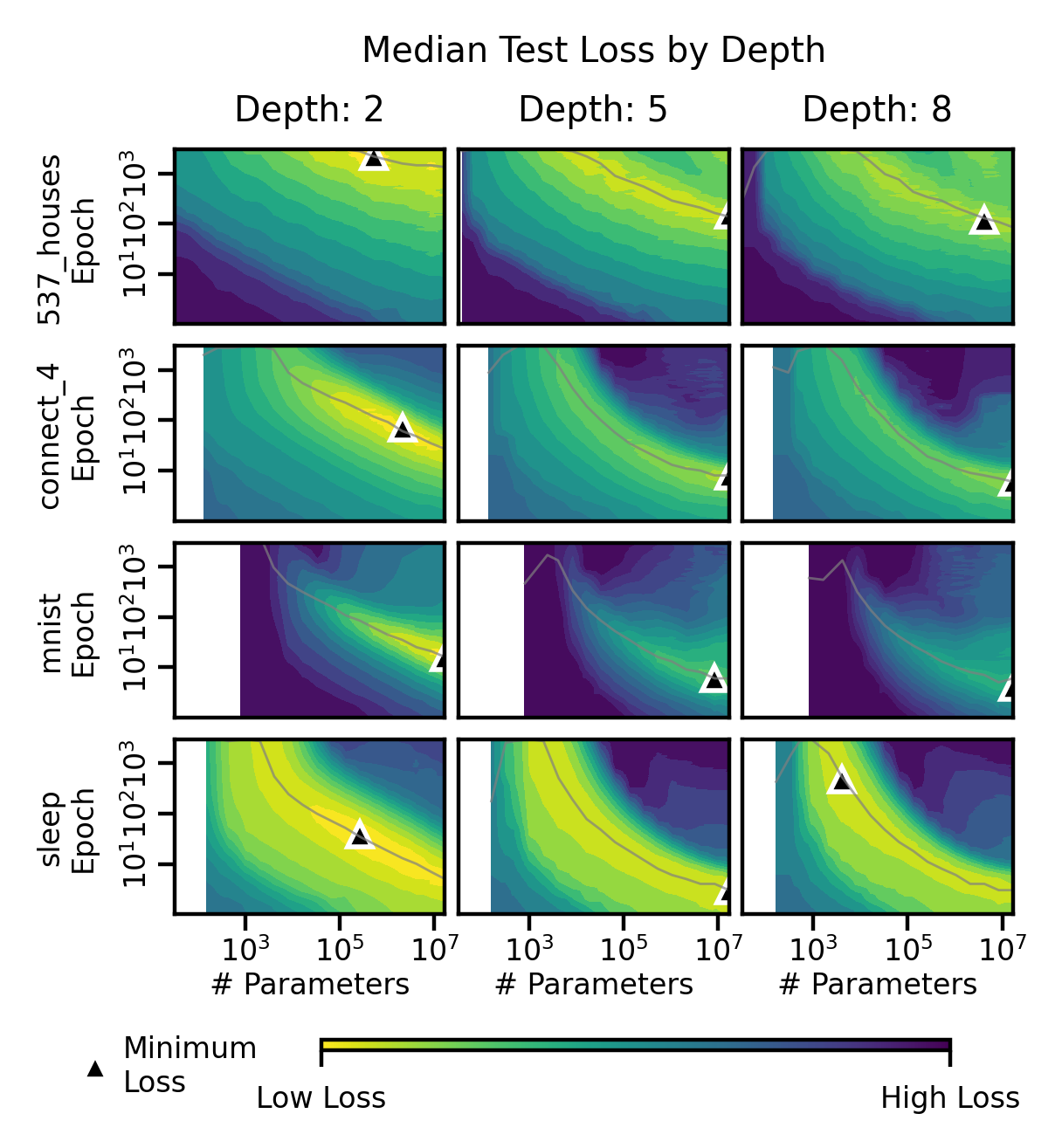}
        \hspace{-3mm}
        \includegraphics[width=3.54in]{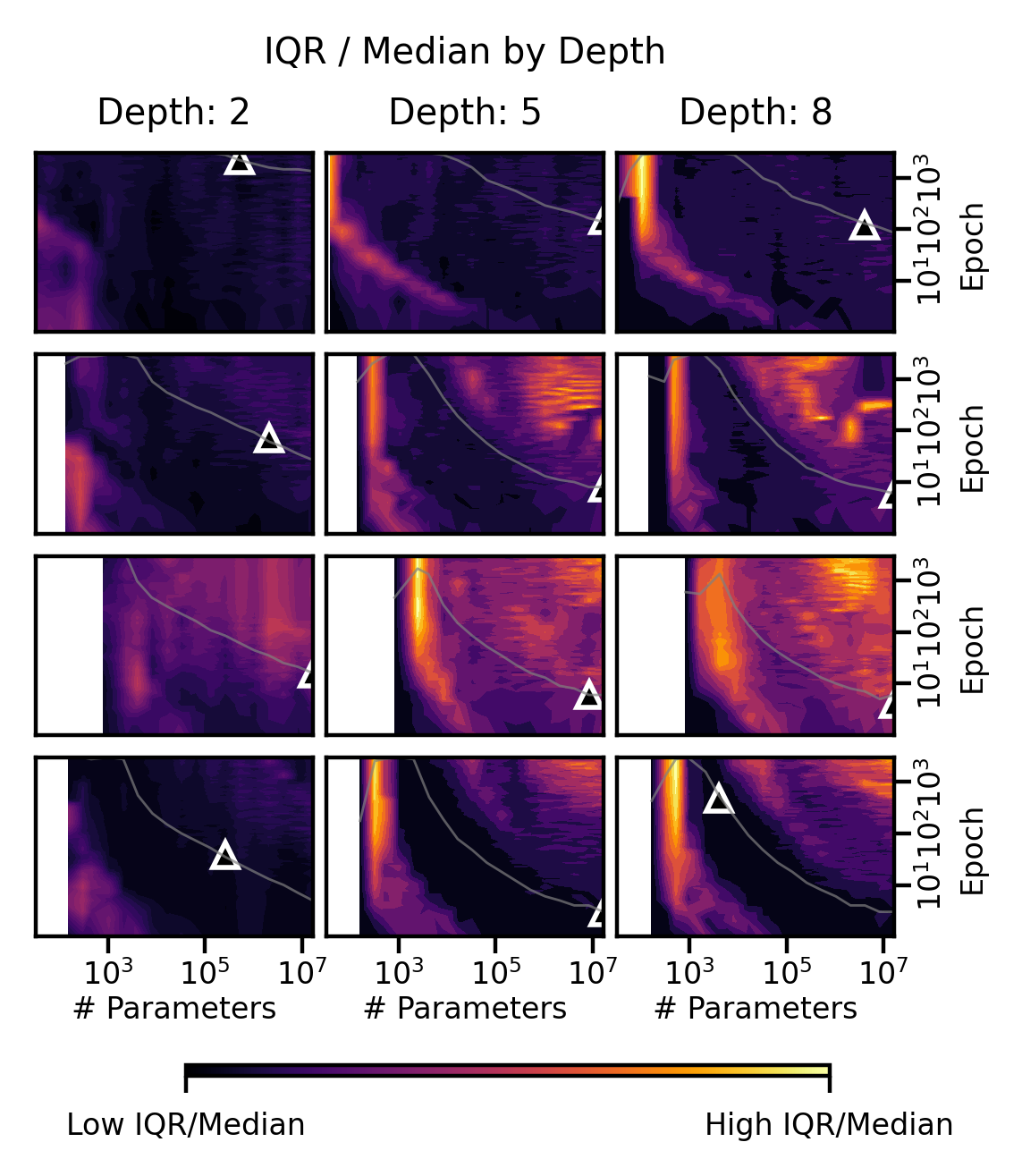}
    } 
    \caption{\textbf{Contour plots of median test loss (left) and interquartile range relative to the median test loss (right)} plotted over the number of parameters vs. training epoch for rectangular networks from the Primary Sweep trained on four tasks (rows) at three depth levels (number of hidden layers, columns).
    To highlight the ``\textit{butter-zone}'' of lowest median test losses, we restrict the contours to losses less than two times the minimum loss.
    The triangle marks the minimum median test loss and is located at the same position in both subfigures.
    A gray line marks the median epoch at which test loss was minimized across all repetitions of that experiment.
    Colors are logarithmically scaled to show detail and are normalized by task.
    Expanded versions of this figure over more hyperparameters and datasets are included in the Appendix \ref{sec:heatmaps}.
    }
    \label{fig:heatmap}
\end{figure*}

\subsection{Motivations for and observations from each sweep}
\label{sec:observations}

\begin{figure*}[t]
    \centering
    \includegraphics[width=\linewidth]{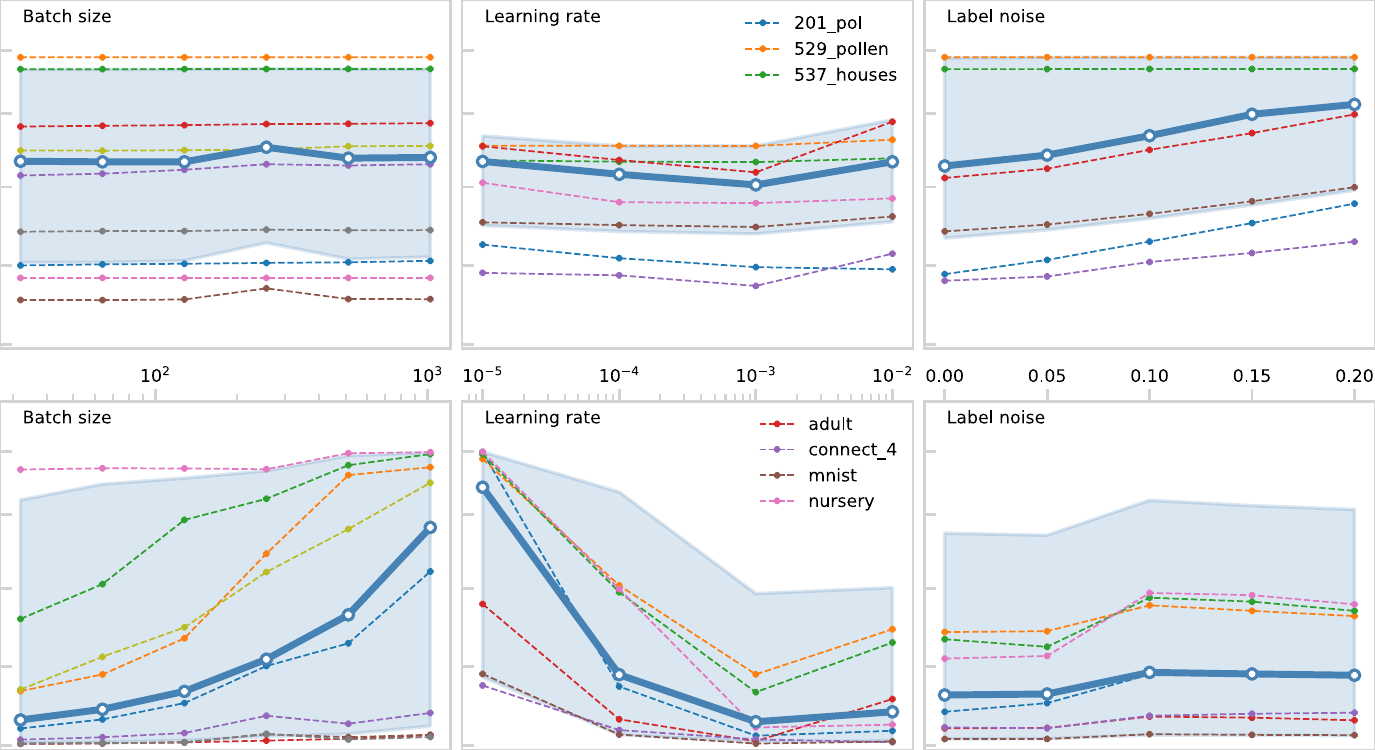}
    \caption{\textbf{Marginal effects plots for batch size, learning rate, and label noise using each parameter's corresponding sweep.} See  Figure \ref{fig:marginals1} for a description of the axes, lines, and shaded region.
    }
    \label{fig:marginals2}
\end{figure*}

\noindent We began with the ``Primary Sweep'' to investigate the influence of parameter count, depth, width, shape, and number of epochs on model performance. %
Given the substantial body of work in NAS, we suspected these factors play critical roles in model performance. %
This sweep is a grid search spanning 13 learning tasks, 20 different numbers of trainable parameters (network sizes), 8 network shapes, and 14 depths with 30 repetitions of each combination.
Each repetition in the Primary Sweep was trained for 3 thousand epochs.
Consistent with recent work \cite{hestness2017deep,rosenfeld2019constructive,sellam2020bleurt} suggesting that scaling laws of neural networks are roughly power-law relationships, we found that many patterns were better visualized by plotting means and quartiles on semilog- or log-scaled axes rather than on linear scaled axes.

As shown in Figure \ref{fig:marginals1}, and Appendix \ref{sec:heatmaps}, Figures \ref{fig:shape1_median}-\ref{fig:shape2_cov}, network shape had a negligible impact on training performance.
This finding is consistent with \cite{kaplan2020_scaling_laws,rosenfeld2021scaling,bansal2022data,tay2022_scaling_laws} and further study of BUTTER and other datasets could be critical to improving the efficiency of NAS approaches. 
We also see that parameter count strongly influences the generalization performance of a network, with larger networks performing better, even well beyond the interpolation threshold.
And, larger networks maximize performance in fewer epochs, but both effects have diminishing returns (note the log-scaled x-axes).
Depth has a much milder effect on performance, with a flat optimum laying between three and seven layers.
However, deeper networks typically maximized performance in fewer epochs than shallow networks did.
As in \cite{kaplan2020_scaling_laws}, we find that networks perform well at a wide range of width-depth ratios, a finding somewhat inconsistent with \cite{goodfellow2013multi}.
Figure \ref{fig:aspect_ratio} shows that, for four tasks in the Primary Sweep, networks with width-to-depth aspect ratios across many orders of magnitude can achieve good test losses.
However, performance degrades for deep and narrow networks and, to a lesser extent, wide and shallow networks. 
Each data point in this figure depicts the median minimum test loss for a single experiment.
Figure \ref{fig:heatmap} visualizes how relative training variation is high surrounding the interpolation threshold: sometimes a training run is ``luckier'' than others; further study could strengthen our understanding of the LTH.

We observed the greatest differences in generalization error between parameter count and epoch, and see hints of epoch-wise double descent at high parameter counts (see depth 8 in Figure \ref{fig:heatmap} and Appendix \ref{sec:heatmaps}).
We were curious to see if epoch-wise double descent could be triggered by either training longer, using larger networks, or adding label noise (as suggested in \cite{nakkiran2021deep}).
So, we ran the 300 Epoch Sweep on larger networks than in the Primary Sweep, the 30k Epoch Sweep on the smaller half of parameter counts from the Primary Sweep, and the Label Noise Sweep repeating the Primary Sweep with increasing levels of label noise.
For regression tasks, additive Gaussian noise with a standard deviation equal to a percentage of the standard deviation of the response variable was used in place of label permutation.
While these sweeps did not show strong evidence of epoch-wise double descent, they provide a broader view into the impacts of training time and parameter count (see Appendix \ref{sec:heatmaps} Figures \ref{fig:label_noise_median} and \ref{fig:label_noise_cov}).
The 30k sweep can act as a validation set when compared to the Primary Sweep, and the 300 sweep extends the primary into larger parameter counts (see Appendix \ref{sec:analysis}). %
As visualized in Figure \ref{fig:marginals2} and Appendix \ref{sec:heatmaps} Figures \ref{fig:label_noise_median} and \ref{fig:label_noise_cov}, high levels of label noise predictably degrade performance.

\begin{figure}[!t]
    \centering
    \includegraphics[width=3in]{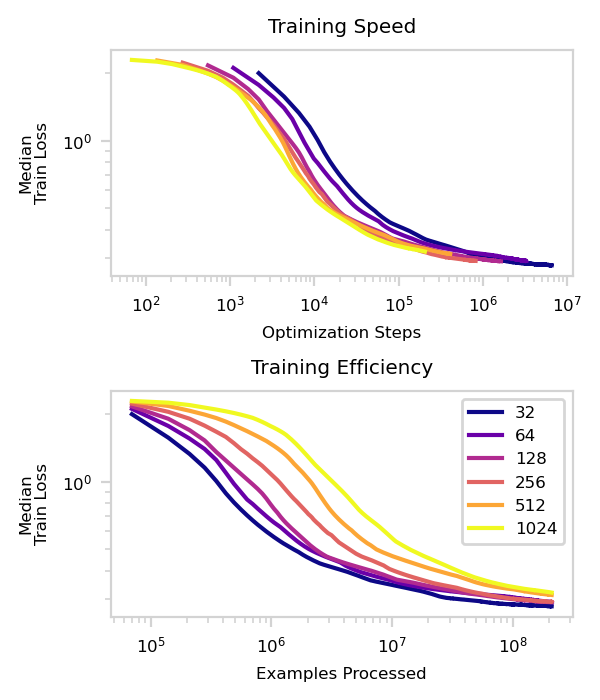}
    \caption{\textbf{Training speed and efficiency depend on batch size.} 
    These plots visualize the median train loss for a 4096 parameter network trained on mnist.
    As in \cite{mccandlish2018empirical}, we find that larger batch sizes require fewer optimization steps to train while smaller batch sizes require fewer examples. 
    }
    \label{fig:batch_sweep}
\end{figure}

Next, we were curious to capture how different learning rates and batch sizes modulate training performance.
Figure \ref{fig:marginals2} suggests batch size has a weak effect on generalization performance and training epochs, but larger batch sizes can increase the number of training epochs to minimize test loss.
Figure \ref{fig:batch_sweep} shows that larger batch sizes required fewer optimization steps but smaller batches are more sample-efficient (also see Appendix \ref{sec:heatmaps} Figures \ref{fig:batch_size_median} and \ref{fig:batch_size_cov}), consistent with \cite{mccandlish2018empirical}, 
Learning rate had a moderate impact on the best test loss achieved, with optimal rates near $10^-3$.
Appendix \ref{sec:heatmaps} Figures \ref{fig:lr_median}-\ref{fig:batch_size_cov} further visualize the Learning Rate and Batch Size Sweeps.

\begin{figure*}[!t]
    \centering
    \includegraphics[width=\linewidth]{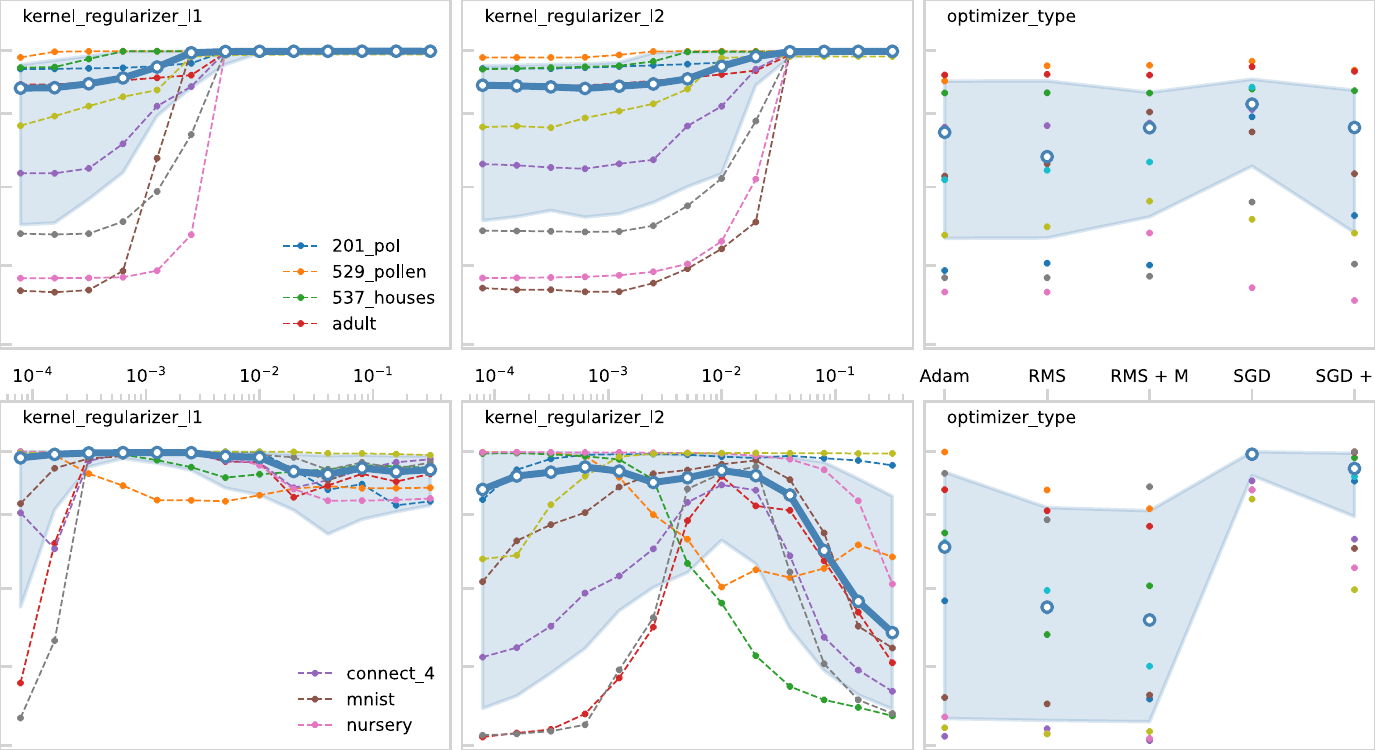}
    \caption{\textbf{Marginal effects plots of regularization strength using the Regularization Sweep (left) and optimizer choice using the Optimizer Sweep (right).} 
    For the regularization plots we use the unpenalized test loss.
    ``+ M'' cases used $0.9$ momentum, other cases used none.
    See Figure \ref{fig:marginals1} for a description of the axes, lines, and shaded region.
    }
    \label{fig:marginals3}
\end{figure*}

Regularization impacts generalization performance, so we performed a ``Regularization Sweep'' to measure how L1 and L2 regularization mediate training and overtraining.
This sweep trains rectangular networks of the Primary Sweep at 14 levels of L1 and L2 kernel regularization each.
Figure \ref{fig:marginals3} shows that heavy regularization can increase test losses and sometimes mildly increase training epochs but, as visualized in Appendix  \ref{sec:heatmaps} Figures \ref{fig:l1_median}-\ref{fig:l2_cov}, it can also delay the onset and severity of overtraining.
Regularization did not induce deep double descent in our experiments.

Lastly, to verify that Adam was not biasing the dataset, we conducted an Optimizer Sweep which repeated experiments on rectangular networks from the Primary Sweep using RMSProp and SGD at several batch sizes and learning rates.
Figure \ref{fig:marginals3} reveals that choice of optimizer moderately impacts training speed and quality but does not result in wholly different dynamics (see Appendix \ref{sec:heatmaps} Figures \ref{fig:optimizer_median} and \ref{fig:optimizer_cov}).
Therefore, we believe patterns seen in this dataset may also manifest when using other optimizers.
SGD shows milder decreases in epochs to minimize test loss as parameter count increases beyond the interpolation threshold.
This may be due to its fixed step-size; further investigation is required.

\subsection{How we made BUTTER: a high-throughput distributed empirical machine learning framework}

\noindent As detailed in Appendix \ref{sec:dataset-datasheet}, we collected training and test performance statistics at each training epoch of each repetition as reported by Tensorflow and stored them in the Postgres database from which BUTTER was extracted.
20 statistics were collected at each epoch including training and test loss, accuracy, mean squared error, and KL-divergence.
To coordinate these experiments, we developed  \href{https://github.com/NREL/BUTTER-Empirical-Deep-Learning-Experimental-Framework}{an object-oriented Python-based framework} that supported the simultaneous coordinated execution of the neural network training runs in this dataset across four HPC clusters \cite{EmpiricalFramework}.
We hope that other researchers study and expand this framework to generate additional large empirical machine learning datasets to inform both theoretical and practical research.
The framework allows for specifications of computational experiments to be serialized and enqueued in a  database-backed task queue which is used to distribute runs to worker nodes which later record the results into the same database.
Table \ref{tab:hpc} lists the specifications of the systems we utilized.

\begin{table}
    \caption{\textbf{Specifications of the computing systems utilized to execute experimental runs.}}
    \label{tab:hpc}
    \centering
    {
    \setlength\tabcolsep{1.0mm}
    \begin{tabular}{lrlrrc}
        \midrule
         System & Nodes & \multicolumn{1}{c}{CPU} & Cores & RAM & GPU \\
         \midrule
         \multirow{2}{*}{Cori} & 9,688 & 1x 1.4GHz Xeon Phi 7250 & 68 & 112GB & -\\
         & 2,388 & 2x 2.3GHz Xeon E5-2698v3 & 64 & 128GB & - \\
         \midrule
         \multirow{2}{*}{Eagle}& 2,568 &  2x 3.0GHz Xeon Gold 6154 & 36 & 96GB & - \\
         & 50 &  2x 3.0GHz Xeon Gold 6154 & 36 & 96GB & 2x V100 \\
         \midrule
         \multirow{2}{*}{Perlmutter} & 3,072 & 2x 2.5GHz EPYC 7763 & 128 & 512GB & - \\
         & 1,536 & 1x 2.5GHz EPYC 7763 & 64 & 256GB & 4x A100 \\
         \midrule
         \multirow{3}{*}{Vermilion} & 62 & 1x 2.4GHz EPYC 7532 & 32 & 128GB & - \\
         & 18 & 2x 2.4GHz EPYC 7532 & 64 & 256GB & - \\
         & 6 & 1x 2.4GHz EPYC 7532 & 32 & 128GB & 1x A100 \\
         \midrule
    \end{tabular}
    }
\end{table}

\subsection{Data publication}

\noindent \href{https://data.openei.org/submissions/5708}{The entire 1.7 TB experimental dataset is available from the Open Energy Data Initiative (OEDI) data repository} as partitioned parquet files \cite{BUTTERDataset}.
Each repetition record contains test and training loss, accuracy, mean squared error, mean absolute error, root mean squared error, squared logarithmic error, hinge loss, squared hinge loss, cosine similarity, and Kullback-Leibler divergence measured at each training epoch.
We also record the random seed; version numbers of the framework, Python, and Tensorflow; hostname; and operating system used to execute the run.
We also provide a smaller summary dataset aggregated over all repetitions of each experiment including statistics such as average, standard deviation, minimum, maximum, and median test and training losses at each epoch.

\section{Discussion}
\label{sec:Discussion}

\noindent We presented BUTTER, a empirical deep learning dataset to further our understanding of deep learning phenomenon, and hopefully to guide theoretical research and inform practical application of neural networks.
It is critical to expand empirical study of machine learning beyond this dataset to different types and aspects of neural network architectures including sparse, convolutional, recurrent, and transformer networks.
To facilitate such future studies, we also publish our distributed experimental framework capable of high-throughput machine learning experimentation across multiple systems.
Further analysis into learning rate, label noise, and regularization could generate compelling findings and a theoretical explanation for the trends we described (see Appendix \ref{sec:analysis}).
If such patterns are shown to hold across more datasets and hyperparameters, it may be possible to optimize the energy-intensive process of hyperparameter tuning by leveraging these analytic relationships.
We acknowledge that this research contributes to the energy consumption and carbon footprint of deep learning.
However, by providing this dataset as a public resource we hope to reduce future energy costs that would have otherwise been incurred by others in recreating these results and we believe this dataset promotes equity by providing this data to those without the resources to generate it.

\section*{Acknowledgements}
This work was authored by the National Renewable Energy Laboratory (NREL), operated by Alliance for Sustainable Energy, LLC, for the U.S. Department of Energy (DOE) under Contract No. DE-AC36-08GO28308. This work was supported by the Laboratory Directed Research and Development (LDRD) Program at NREL. The views expressed in the article do not necessarily represent the views of the DOE or the U.S. Government. The U.S. Government retains and the publisher, by accepting the article for publication, acknowledges that the U.S. Government retains a nonexclusive, paid-up, irrevocable, worldwide license to publish or reproduce the published form of this work, or allow others to do so, for U.S. Government purposes. 

The authors declare that they have no conflict of interest.

\bibliography{main}

\begin{thebibliography}{10}
\providecommand{\url}[1]{#1}
\csname url@samestyle\endcsname
\providecommand{\newblock}{\relax}
\providecommand{\bibinfo}[2]{#2}
\providecommand{\BIBentrySTDinterwordspacing}{\spaceskip=0pt\relax}
\providecommand{\BIBentryALTinterwordstretchfactor}{4}
\providecommand{\BIBentryALTinterwordspacing}{\spaceskip=\fontdimen2\font plus
\BIBentryALTinterwordstretchfactor\fontdimen3\font minus
  \fontdimen4\font\relax}
\providecommand{\BIBforeignlanguage}[2]{{%
\expandafter\ifx\csname l@#1\endcsname\relax
\typeout{** WARNING: IEEEtran.bst: No hyphenation pattern has been}%
\typeout{** loaded for the language `#1'. Using the pattern for}%
\typeout{** the default language instead.}%
\else
\language=\csname l@#1\endcsname
\fi
#2}}
\providecommand{\BIBdecl}{\relax}
\BIBdecl

\bibitem{chen2021understanding}
W.~Chen, X.~Gong, Y.~Wei, H.~Shi, Z.~Yan, Y.~Yang, and Z.~Wang, ``Understanding
  and accelerating neural architecture search with training-free and
  theory-grounded metrics,'' \emph{arXiv preprint arXiv:2108.11939}, 2021.

\bibitem{hornik1989multilayer}
K.~Hornik, M.~Stinchcombe, and H.~White, ``Multilayer feedforward networks are
  universal approximators,'' \emph{Neural networks}, vol.~2, no.~5, pp.
  359--366, 1989.

\bibitem{barron1993universal}
A.~R. Barron, ``Universal approximation bounds for superpositions of a
  sigmoidal function,'' \emph{IEEE Transactions on Information theory},
  vol.~39, no.~3, pp. 930--945, 1993.

\bibitem{zhang2017learnability}
Y.~Zhang, J.~Lee, M.~Wainwright, and M.~I. Jordan, ``On the learnability of
  fully-connected neural networks,'' in \emph{Artificial Intelligence and
  Statistics}.\hskip 1em plus 0.5em minus 0.4em\relax PMLR, 2017, pp. 83--91.

\bibitem{kohler2021rate}
M.~Kohler and S.~Langer, ``On the rate of convergence of fully connected deep
  neural network regression estimates,'' \emph{The Annals of Statistics},
  vol.~49, no.~4, pp. 2231--2249, 2021.

\bibitem{LANGER2021104695}
\BIBentryALTinterwordspacing
S.~Langer, ``Analysis of the rate of convergence of fully connected deep neural
  network regression estimates with smooth activation function,'' \emph{Journal
  of Multivariate Analysis}, vol. 182, p. 104695, 2021. [Online]. Available:
  \url{https://www.sciencedirect.com/science/article/pii/S0047259X20302761}
\BIBentrySTDinterwordspacing

\bibitem{jacot_neural_2020}
\BIBentryALTinterwordspacing
A.~Jacot, F.~Gabriel, and C.~Hongler, ``\BIBforeignlanguage{en}{Neural
  {Tangent} {Kernel}: {Convergence} and {Generalization} in {Neural}
  {Networks}},'' Feb. 2020, arXiv:1806.07572 [cs, math, stat]. [Online].
  Available: \url{http://arxiv.org/abs/1806.07572}
\BIBentrySTDinterwordspacing

\bibitem{eldan2016power}
R.~Eldan and O.~Shamir, ``The power of depth for feedforward neural networks,''
  in \emph{Conference on learning theory}.\hskip 1em plus 0.5em minus
  0.4em\relax PMLR, 2016, pp. 907--940.

\bibitem{elbrachter2021deep}
D.~Elbr{\"a}chter, D.~Perekrestenko, P.~Grohs, and H.~B{\"o}lcskei, ``Deep
  neural network approximation theory,'' \emph{IEEE Transactions on Information
  Theory}, vol.~67, no.~5, pp. 2581--2623, 2021.

\bibitem{lu2021deep}
J.~Lu, Z.~Shen, H.~Yang, and S.~Zhang, ``Deep network approximation for smooth
  functions,'' \emph{SIAM Journal on Mathematical Analysis}, vol.~53, no.~5,
  pp. 5465--5506, 2021.

\bibitem{deep_vs_shallow_theory}
\BIBentryALTinterwordspacing
H.~N. Mhaskar and T.~Poggio, ``Deep vs. shallow networks: An approximation
  theory perspective,'' \emph{Analysis and Applications}, vol.~14, no.~06, pp.
  829--848, 2016. [Online]. Available:
  \url{https://doi.org/10.1142/S0219530516400042}
\BIBentrySTDinterwordspacing

\bibitem{yarotsky2018optimal}
D.~Yarotsky, ``Optimal approximation of continuous functions by very deep relu
  networks,'' in \emph{Conference on learning theory}.\hskip 1em plus 0.5em
  minus 0.4em\relax PMLR, 2018, pp. 639--649.

\bibitem{yarotsky2020phase}
D.~Yarotsky and A.~Zhevnerchuk, ``The phase diagram of approximation rates for
  deep neural networks,'' \emph{Advances in neural information processing
  systems}, vol.~33, pp. 13\,005--13\,015, 2020.

\bibitem{pezeshki2021multi}
M.~Pezeshki, A.~Mitra, Y.~Bengio, and G.~Lajoie, ``Multi-scale feature learning
  dynamics: Insights for double descent,'' \emph{arXiv preprint
  arXiv:2112.03215}, 2021.

\bibitem{chen2021multiple}
L.~Chen, Y.~Min, M.~Belkin, and A.~Karbasi, ``Multiple descent: Design your own
  generalization curve,'' \emph{Advances in Neural Information Processing
  Systems}, vol.~34, 2021.

\bibitem{adlam2020quadratic}
\BIBentryALTinterwordspacing
B.~Adlam and J.~Pennington, ``The neural tangent kernel in high dimensions:
  Triple descent and a multi-scale theory of generalization,'' in
  \emph{Proceedings of the 37th International Conference on Machine Learning},
  ser. Proceedings of Machine Learning Research, H.~D. III and A.~Singh, Eds.,
  vol. 119.\hskip 1em plus 0.5em minus 0.4em\relax PMLR, 13--18 Jul 2020, pp.
  74--84. [Online]. Available:
  \url{https://proceedings.mlr.press/v119/adlam20a.html}
\BIBentrySTDinterwordspacing

\bibitem{geman1992_bias_variance_in_nn}
S.~Geman, E.~Bienenstock, and R.~Doursat, ``Neural networks and the
  bias/variance dilemma,'' \emph{Neural computation}, vol.~4, no.~1, pp. 1--58,
  1992.

\bibitem{hastie2009elements}
T.~Hastie, R.~Tibshirani, J.~H. Friedman, and J.~H. Friedman, \emph{The
  elements of statistical learning: data mining, inference, and
  prediction}.\hskip 1em plus 0.5em minus 0.4em\relax Springer, 2009, vol.~2.

\bibitem{neyshabur2015_deep_descent_evidence}
B.~Neyshabur, R.~Tomioka, and N.~Srebro, ``In search of the real inductive
  bias: On the role of implicit regularization in deep learning.'' in
  \emph{ICLR (Workshop)}, 2015.

\bibitem{neyshabur2017_deep_descent_evidence}
B.~Neyshabur, ``Implicit regularization in deep learning,'' \emph{arXiv
  preprint arXiv:1709.01953}, 2017.

\bibitem{neal2018_double_descent}
B.~Neal, S.~Mittal, A.~Baratin, V.~Tantia, M.~Scicluna, S.~Lacoste-Julien, and
  I.~Mitliagkas, ``A modern take on the bias-variance tradeoff in neural
  networks,'' \emph{arXiv preprint arXiv:1810.08591}, 2018.

\bibitem{belkin_2019_double_descent}
\BIBentryALTinterwordspacing
M.~Belkin, D.~Hsu, S.~Ma, and S.~Mandal, ``Reconciling modern machine-learning
  practice and the classical bias\&\#x2013;variance trade-off,''
  \emph{Proceedings of the National Academy of Sciences}, vol. 116, no.~32, pp.
  15\,849--15\,854, 2019. [Online]. Available:
  \url{https://www.pnas.org/doi/abs/10.1073/pnas.1903070116}
\BIBentrySTDinterwordspacing

\bibitem{nakkiran2019deep}
\BIBentryALTinterwordspacing
P.~Nakkiran, G.~Kaplun, Y.~Bansal, T.~Yang, B.~Barak, and I.~Sutskever, ``Deep
  {Double} {Descent}: {Where} {Bigger} {Models} and {More} {Data} {Hurt},''
  \emph{International Conference on Learning Representations}, Dec. 2019.
  [Online]. Available: \url{http://arxiv.org/abs/1912.02292}
\BIBentrySTDinterwordspacing

\bibitem{nakkiran2021deep}
------, ``Deep double descent: Where bigger models and more data hurt,''
  \emph{Journal of Statistical Mechanics: Theory and Experiment}, vol. 2021,
  no.~12, p. 124003, 2021.

\bibitem{NAS_HPO_Bench}
A.~Klein and F.~Hutter, ``Tabular benchmarks for joint architecture and
  hyperparameter optimization,'' \emph{arXiv preprint arXiv:1905.04970}, 2019.

\bibitem{hirose2021hpo}
Y.~Hirose, N.~Yoshinari, and S.~Shirakawa, ``Nas-hpo-bench-ii: A benchmark
  dataset on joint optimization of convolutional neural network architecture
  and training hyperparameters,'' in \emph{Asian Conference on Machine
  Learning}.\hskip 1em plus 0.5em minus 0.4em\relax PMLR, 2021, pp. 1349--1364.

\bibitem{LCBench}
L.~Zimmer, M.~Lindauer, and F.~Hutter, ``Auto-pytorch: Multi-fidelity
  metalearning for efficient and robust autodl,'' \emph{IEEE Transactions on
  Pattern Analysis and Machine Intelligence}, vol.~43, no.~9, pp. 3079--3090,
  2021.

\bibitem{Dong2020NAS-Bench-201}
\BIBentryALTinterwordspacing
X.~Dong and Y.~Yang, ``Nas-bench-201: Extending the scope of reproducible
  neural architecture search,'' in \emph{International Conference on Learning
  Representations}, 2020. [Online]. Available:
  \url{https://openreview.net/forum?id=HJxyZkBKDr}
\BIBentrySTDinterwordspacing

\bibitem{Dong2022NATSBenchBN}
X.~Dong, L.~Liu, K.~Musial, and B.~Gabrys, ``Nats-bench: Benchmarking nas
  algorithms for architecture topology and size,'' \emph{IEEE Transactions on
  Pattern Analysis and Machine Intelligence}, vol.~44, pp. 3634--3646, 2022.

\bibitem{ying2019bench}
C.~Ying, A.~Klein, E.~Christiansen, E.~Real, K.~Murphy, and F.~Hutter,
  ``Nas-bench-101: Towards reproducible neural architecture search,'' in
  \emph{International Conference on Machine Learning}.\hskip 1em plus 0.5em
  minus 0.4em\relax PMLR, 2019, pp. 7105--7114.

\bibitem{mehrotra2021nasbenchasr}
\BIBentryALTinterwordspacing
A.~Mehrotra, A.~G. C.~P. Ramos, S.~Bhattacharya, {\L}.~Dudziak, R.~Vipperla,
  T.~Chau, M.~S. Abdelfattah, S.~Ishtiaq, and N.~D. Lane,
  ``{\{}NAS{\}}-bench-{\{}asr{\}}: Reproducible neural architecture search for
  speech recognition,'' in \emph{International Conference on Learning
  Representations}, 2021. [Online]. Available:
  \url{https://openreview.net/forum?id=CU0APx9LMaL}
\BIBentrySTDinterwordspacing

\bibitem{trans-nas-bench}
\BIBentryALTinterwordspacing
Y.~Duan, X.~Chen, H.~Xu, Z.~Chen, X.~Liang, T.~Zhang, and Z.~Li,
  ``Transnas-bench-101: Improving transferability and generalizability of
  cross-task neural architecture search,'' 2021. [Online]. Available:
  \url{https://arxiv.org/abs/2105.11871}
\BIBentrySTDinterwordspacing

\bibitem{kaplan2020_scaling_laws}
J.~Kaplan, S.~McCandlish, T.~Henighan, T.~B. Brown, B.~Chess, R.~Child,
  S.~Gray, A.~Radford, J.~Wu, and D.~Amodei, ``Scaling laws for neural language
  models,'' \emph{arXiv preprint arXiv:2001.08361}, 2020.

\bibitem{henighan2020_scaling_laws}
T.~Henighan, J.~Kaplan, M.~Katz, M.~Chen, C.~Hesse, J.~Jackson, H.~Jun, T.~B.
  Brown, P.~Dhariwal, S.~Gray \emph{et~al.}, ``Scaling laws for autoregressive
  generative modeling,'' \emph{arXiv preprint arXiv:2010.14701}, 2020.

\bibitem{tay2022_scaling_laws}
Y.~Tay, M.~Dehghani, S.~Abnar, H.~W. Chung, W.~Fedus, J.~Rao, S.~Narang, V.~Q.
  Tran, D.~Yogatama, and D.~Metzler, ``Scaling laws vs model architectures: How
  does inductive bias influence scaling?'' \emph{arXiv preprint
  arXiv:2207.10551}, 2022.

\bibitem{jiang2019fantastic}
\BIBentryALTinterwordspacing
Y.~Jiang*, B.~Neyshabur*, H.~Mobahi, D.~Krishnan, and S.~Bengio, ``Fantastic
  generalization measures and where to find them,'' in \emph{International
  Conference on Learning Representations}, 2020. [Online]. Available:
  \url{https://openreview.net/forum?id=SJgIPJBFvH}
\BIBentrySTDinterwordspacing

\bibitem{dziugaite2020search}
G.~K. Dziugaite, A.~Drouin, B.~Neal, N.~Rajkumar, E.~Caballero, L.~Wang,
  I.~Mitliagkas, and D.~M. Roy, ``In search of robust measures of
  generalization,'' \emph{Advances in Neural Information Processing Systems},
  vol.~33, pp. 11\,723--11\,733, 2020.

\bibitem{novak2018sensitivity}
R.~Novak, Y.~Bahri, D.~A. Abolafia, J.~Pennington, and J.~Sohl-Dickstein,
  ``Sensitivity and generalization in neural networks: an empirical study,'' in
  \emph{International Conference on Learning Representations}, 2018.

\bibitem{frankle2018lottery}
\BIBentryALTinterwordspacing
J.~Frankle and M.~Carbin, ``The lottery ticket hypothesis: Finding sparse,
  trainable neural networks,'' in \emph{7th International Conference on
  Learning Representations, {ICLR} 2019, New Orleans, LA, USA, May 6-9,
  2019}.\hskip 1em plus 0.5em minus 0.4em\relax OpenReview.net, 2019. [Online].
  Available: \url{https://openreview.net/forum?id=rJl-b3RcF7}
\BIBentrySTDinterwordspacing

\bibitem{mehta2022bench}
\BIBentryALTinterwordspacing
Y.~Mehta, C.~White, A.~Zela, A.~Krishnakumar, G.~Zabergja, S.~Moradian,
  M.~Safari, K.~Yu, and F.~Hutter, ``{NAS}-bench-suite: {NAS} evaluation is
  (now) surprisingly easy,'' in \emph{International Conference on Learning
  Representations}, 2022. [Online]. Available:
  \url{https://openreview.net/forum?id=0DLwqQLmqV}
\BIBentrySTDinterwordspacing

\bibitem{dong2020bench}
\BIBentryALTinterwordspacing
X.~Dong and Y.~Yang, ``Nas-bench-201: Extending the scope of reproducible
  neural architecture search,'' in \emph{International Conference on Learning
  Representations}, 2020. [Online]. Available:
  \url{https://openreview.net/forum?id=HJxyZkBKDr}
\BIBentrySTDinterwordspacing

\bibitem{zela2020bench}
\BIBentryALTinterwordspacing
A.~Zela, J.~Siems, and F.~Hutter, ``Nas-bench-1shot1: Benchmarking and
  dissecting one-shot neural architecture search,'' in \emph{International
  Conference on Learning Representations}, 2020. [Online]. Available:
  \url{https://openreview.net/forum?id=SJx9ngStPH}
\BIBentrySTDinterwordspacing

\bibitem{dong2021nats}
X.~Dong, L.~Liu, K.~Musial, and B.~Gabrys, ``Nats-bench: Benchmarking nas
  algorithms for architecture topology and size,'' \emph{IEEE transactions on
  pattern analysis and machine intelligence}, 2021.

\bibitem{su2021prioritized}
X.~Su, T.~Huang, Y.~Li, S.~You, F.~Wang, C.~Qian, C.~Zhang, and C.~Xu,
  ``Prioritized architecture sampling with monto-carlo tree search,'' in
  \emph{Proceedings of the IEEE/CVF Conference on Computer Vision and Pattern
  Recognition}, 2021, pp. 10\,968--10\,977.

\bibitem{wu2019fbnet}
B.~Wu, X.~Dai, P.~Zhang, Y.~Wang, F.~Sun, Y.~Wu, Y.~Tian, P.~Vajda, Y.~Jia, and
  K.~Keutzer, ``Fbnet: Hardware-aware efficient convnet design via
  differentiable neural architecture search,'' in \emph{Proceedings of the
  IEEE/CVF Conference on Computer Vision and Pattern Recognition}, 2019, pp.
  10\,734--10\,742.

\bibitem{mehrotra2020bench}
A.~Mehrotra, A.~G.~C. Ramos, S.~Bhattacharya, {\L}.~Dudziak, R.~Vipperla,
  T.~Chau, M.~S. Abdelfattah, S.~Ishtiaq, and N.~D. Lane, ``Nas-bench-asr:
  Reproducible neural architecture search for speech recognition,'' in
  \emph{International Conference on Learning Representations}, 2020.

\bibitem{duan2021transnas}
Y.~Duan, X.~Chen, H.~Xu, Z.~Chen, X.~Liang, T.~Zhang, and Z.~Li,
  ``Transnas-bench-101: Improving transferability and generalizability of
  cross-task neural architecture search,'' in \emph{Proceedings of the IEEE/CVF
  Conference on Computer Vision and Pattern Recognition}, 2021, pp. 5251--5260.

\bibitem{BUTTERDataset}
\BIBentryALTinterwordspacing
C.~Tripp, J.~Perr-Sauer, M.~Lunacek, and L.~Hayne, ``Butter - empirical deep
  learning dataset,'' 05 2022, the BUTTER Empirical Deep Learning Dataset
  contains epoch-wise training, validation, loss and error scores that
  represent millions of training runs for a variety of deep neural network
  architectures trained with various hyper-paramter configurations. The data
  are available as H5 files for download. [Online]. Available:
  \url{https://data.openei.org/submissions/5708}
\BIBentrySTDinterwordspacing

\bibitem{EmpiricalFramework}
\BIBentryALTinterwordspacing
------, ``Empirical deep learning experimental framework aspect-test,''
  [Computer Software] \url{https://doi.org/10.11578/dc.20220608.2}, may 2022.
  [Online]. Available: \url{https://doi.org/10.11578/dc.20220608.2}
\BIBentrySTDinterwordspacing

\bibitem{EQueue}
\BIBentryALTinterwordspacing
------, ``Job queue (a task queue for coordinating varied tasks across multiple
  hpc resources and hpc jobs),'' [Computer Software]
  \url{https://doi.org/10.11578/dc.20220608.1}, may 2022. [Online]. Available:
  \url{https://doi.org/10.11578/dc.20220608.1}
\BIBentrySTDinterwordspacing

\bibitem{PMLB}
\BIBentryALTinterwordspacing
J.~D. Romano, T.~T. Le, W.~La~Cava, J.~T. Gregg, D.~J. Goldberg,
  P.~Chakraborty, N.~L. Ray, D.~Himmelstein, W.~Fu, and J.~H. Moore, ``{PMLB
  v1.0: an open-source dataset collection for benchmarking machine learning
  methods},'' \emph{Bioinformatics}, vol.~38, no.~3, pp. 878--880, 10 2021.
  [Online]. Available: \url{https://doi.org/10.1093/bioinformatics/btab727}
\BIBentrySTDinterwordspacing

\bibitem{kingma2014adam}
\BIBentryALTinterwordspacing
D.~P. Kingma and J.~Ba, ``Adam: {A} method for stochastic optimization,'' in
  \emph{3rd International Conference on Learning Representations, {ICLR} 2015,
  San Diego, CA, USA, May 7-9, 2015, Conference Track Proceedings}, Y.~Bengio
  and Y.~LeCun, Eds., 2015. [Online]. Available:
  \url{http://arxiv.org/abs/1412.6980}
\BIBentrySTDinterwordspacing

\bibitem{ReLU1}
R.~H. Hahnloser, R.~Sarpeshkar, M.~A. Mahowald, R.~J. Douglas, and H.~S. Seung,
  ``Digital selection and analogue amplification coexist in a cortex-inspired
  silicon circuit,'' \emph{nature}, vol. 405, no. 6789, pp. 947--951, 2000.

\bibitem{ReLU2}
R.~Hahnloser and H.~S. Seung, ``Permitted and forbidden sets in symmetric
  threshold-linear networks,'' \emph{Advances in neural information processing
  systems}, vol.~13, 2000.

\bibitem{rmsprop1}
T.~Tieleman and G.~Hinton, ``Lecture 6.5-rmsprop, coursera: Neural networks for
  machine learning,'' \emph{University of Toronto, Technical Report}, vol.~6,
  2012.

\bibitem{rmsprop2}
A.~Graves, ``Generating sequences with recurrent neural networks,'' \emph{arXiv
  preprint arXiv:1308.0850}, 2013.

\bibitem{adagrad}
J.~Duchi, E.~Hazan, and Y.~Singer, ``Adaptive subgradient methods for online
  learning and stochastic optimization.'' \emph{Journal of machine learning
  research}, vol.~12, no.~7, 2011.

\bibitem{karpathy2017peek}
A.~Karpathy, ``A peek at trends in machine learning,'' \emph{Medium. com},
  2017.

\bibitem{hestness2017deep}
J.~Hestness, S.~Narang, N.~Ardalani, G.~Diamos, H.~Jun, H.~Kianinejad,
  M.~Patwary, M.~Ali, Y.~Yang, and Y.~Zhou, ``Deep learning scaling is
  predictable, empirically,'' \emph{arXiv preprint arXiv:1712.00409}, 2017.

\bibitem{rosenfeld2019constructive}
J.~S. Rosenfeld, A.~Rosenfeld, Y.~Belinkov, and N.~Shavit, ``A constructive
  prediction of the generalization error across scales,'' in
  \emph{International Conference on Learning Representations}, 2019.

\bibitem{sellam2020bleurt}
T.~Sellam, D.~Das, and A.~Parikh, ``Bleurt: Learning robust metrics for text
  generation,'' in \emph{Proceedings of the 58th Annual Meeting of the
  Association for Computational Linguistics}, 2020, pp. 7881--7892.

\bibitem{rosenfeld2021scaling}
J.~S. Rosenfeld, ``Scaling laws for deep learning,'' Ph.D. dissertation,
  Massachusetts Institute of Technology, 2021.

\bibitem{bansal2022data}
Y.~Bansal, B.~Ghorbani, A.~Garg, B.~Zhang, C.~Cherry, B.~Neyshabur, and
  O.~Firat, ``Data scaling laws in nmt: The effect of noise and architecture,''
  in \emph{International Conference on Machine Learning}.\hskip 1em plus 0.5em
  minus 0.4em\relax PMLR, 2022, pp. 1466--1482.

\bibitem{goodfellow2013multi}
I.~J. Goodfellow, Y.~Bulatov, J.~Ibarz, S.~Arnoud, and V.~Shet, ``Multi-digit
  number recognition from street view imagery using deep convolutional neural
  networks,'' \emph{arXiv preprint arXiv:1312.6082}, 2013.

\bibitem{mccandlish2018empirical}
S.~McCandlish, J.~Kaplan, D.~Amodei, and O.~D. Team, ``An empirical model of
  large-batch training,'' \emph{arXiv preprint arXiv:1812.06162}, 2018.

\bibitem{Hunter2007matplotlib}
J.~D. Hunter, ``Matplotlib: A 2d graphics environment,'' \emph{Computing in
  Science \& Engineering}, vol.~9, no.~3, pp. 90--95, 2007.

\bibitem{savitzky1964smoothing}
A.~Savitzky and M.~J. Golay, ``Smoothing and differentiation of data by
  simplified least squares procedures.'' \emph{Analytical chemistry}, vol.~36,
  no.~8, pp. 1627--1639, 1964.

\bibitem{Scipy}
P.~Virtanen, R.~Gommers, T.~E. Oliphant, M.~Haberland, T.~Reddy, D.~Cournapeau,
  E.~Burovski, P.~Peterson, W.~Weckesser, J.~Bright, S.~J. {van der Walt},
  M.~Brett, J.~Wilson, K.~J. Millman, N.~Mayorov, A.~R.~J. Nelson, E.~Jones,
  R.~Kern, E.~Larson, C.~J. Carey, {\.I}.~Polat, Y.~Feng, E.~W. Moore,
  J.~{VanderPlas}, D.~Laxalde, J.~Perktold, R.~Cimrman, I.~Henriksen, E.~A.
  Quintero, C.~R. Harris, A.~M. Archibald, A.~H. Ribeiro, F.~Pedregosa, P.~{van
  Mulbregt}, and {SciPy 1.0 Contributors}, ``{{SciPy} 1.0: Fundamental
  Algorithms for Scientific Computing in Python},'' \emph{Nature Methods},
  vol.~17, pp. 261--272, 2020.

\bibitem{datasetDatasheets}
\BIBentryALTinterwordspacing
T.~Gebru, J.~Morgenstern, B.~Vecchione, J.~W. Vaughan, H.~Wallach, H.~Daumé,
  and K.~Crawford, ``Datasheets for datasets,'' 2018. [Online]. Available:
  \url{https://arxiv.org/abs/1803.09010}
\BIBentrySTDinterwordspacing

\end{thebibliography}

\begin{IEEEbiography}[{\includegraphics[width=1in,height=1.25in,clip,keepaspectratio]{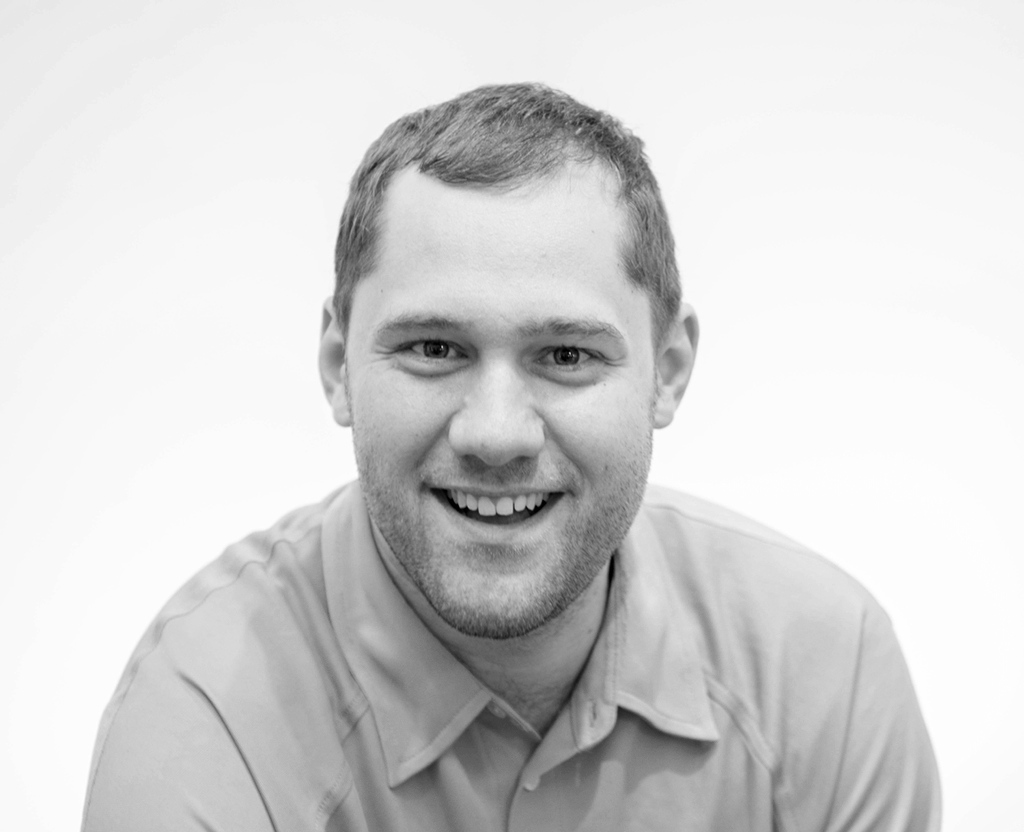}}]{Charles Edison Tripp}
(Member, IEEE) received the B.S. in Electrical Engineering from Rice University in 2006, and the M.S. and Ph.D. degrees in Electrical Engineering from Stanford University in 2008 and 2013 respectively. 
He is a senior scientist in the Artificial Intelligence, Learning, and Intelligent Systems group of the Computational Science Center at the National Renewable Energy Laboratory in Golden, Colorado, USA.
His research interests include machine learning, reinforcement learning, Bayesian methods, and decision science.
\end{IEEEbiography}

\begin{IEEEbiography}[{\includegraphics[width=1in,height=1.25in,clip,keepaspectratio]{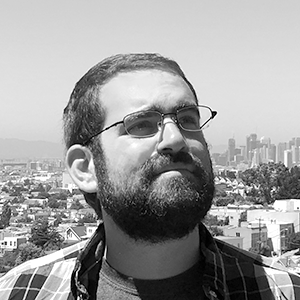}}]{Jordan Perr-Sauer}
received his B.S. in Applied Mathematics from University of Colorado, and his M.S. in Computer Science from the University of Colorado, Boulder in 2022. 
He is a data science researcher in the Data, Analysis, and Visualization group of the Computational Science Center at the National Renewable Energy Laboratory in Golden, Colorado, USA.
His research interests include explainable and interpretable machine learning and software engineering for science.
\end{IEEEbiography}

\begin{IEEEbiography}[{\includegraphics[width=1in,height=1.25in,clip,keepaspectratio]{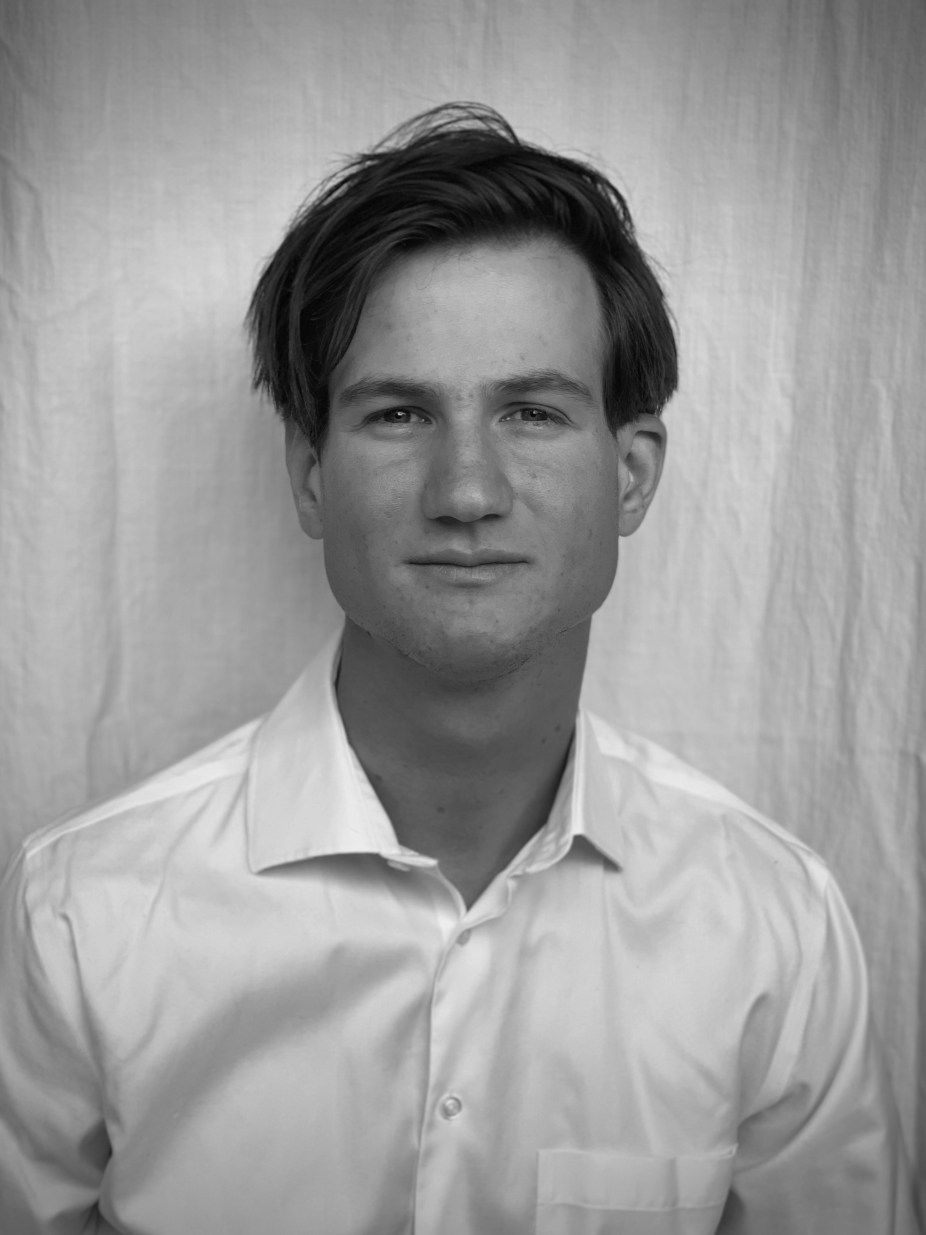}}]{Lucas Hayne}
(Member, IEEE) received a B.S. in Computer Science and Psychology in 2017, an M.S. in Computer Science in 2021, and plans on graduating with his Ph.D. in 2023 from the University of Colorado Boulder.
He is a Data Analyst in the Artificial Intelligence, Learning, and Intelligent Systems group of the Computational Science Center at the National Renewable Energy Laboratory in Golden, Colorado, USA.
His research interests include machine learning interpretability and cognitive science.
\end{IEEEbiography}

\vspace{-10mm}
\begin{IEEEbiography}[{\includegraphics[width=1in,height=1.25in,clip,keepaspectratio]{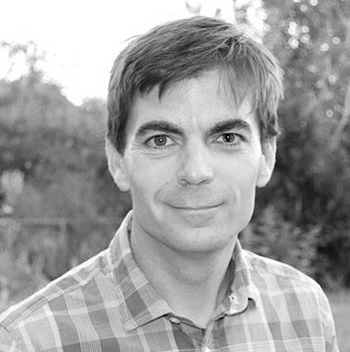}}]{Monte Lunacek}
received his B.S. in Applied Mathematics from the University of Colorado, Colorado Springs, and his M.S. and Ph.D. degrees in Computer Science from Colorado State University.
He is a Senior Scientist in the Computational Science Center at the National Renewable Energy Laboratory.
His research interests include distributed computing, machine learning, data optimization and analysis, and web visualization.
\end{IEEEbiography}

\vspace{-10mm}
\begin{IEEEbiography}[{\includegraphics[width=1in,height=1.25in,clip,keepaspectratio]{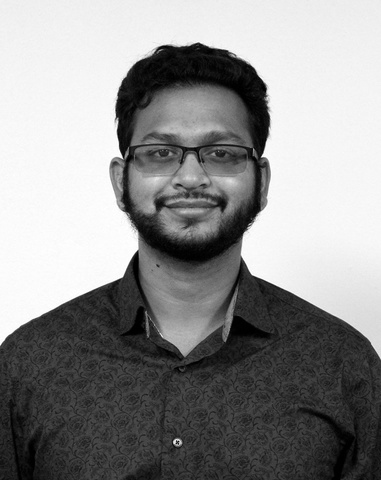}}]{Jamil Gafur}
 received his B.S. in Computer Science from CUNY Lehman College in 2018. Currently, a Ph.D. Student in Computer Science at the University of Iowa  working on localized explanations of neural networks through swarm algorithms. He is a Computational
Scientist in the Artificial Intelligence, Learning, and Intelligent Systems group of the Computational Science Center at the National Renewable Energy Laboratory in Golden, Colorado, USA. His research interests include adversarial attacks, machine learning, and heuristic optimization.\end{IEEEbiography}
\FloatBarrier
\clearpage

\appendix

\section{Appendix}

\noindent This Appendix contains additional details and links that can be used to access and reproduce the BUTTER dataset, as well as a number of plots visualizing several dimensions of interest within the dataset. Each plot visualizes attributes from each sweep of the dataset.
In each plot, we vary one attribute at a time while other attributes remain constant. 
\textbf{Unless otherwise stated we use the following as default values for each attribute: shape: rectangle, depth: 3, learning rate: 0.0001, batch size: 256, label noise: 0.0, and regularization penalty: 0.0.} 
For all plots depicting median losses, we restrict the contours to losses less than 10 times the minimum loss unless otherwise stated. 
Triangles mark the minimum median test losses. 
A gray line marks the median epoch at which test loss was minimized across all repetitions of that experiment.
Colors are logarithmically scaled to show detail and are normalized by task. To reveal details in regions with less variation in plots depicting coefficients of variation, we offset the interquartile range / median slightly before scaling.

Appendix Contents:
\begin{itemize}
    \item Appendix \ref{sec:analysis} A Preliminary Analysis of The Dataset
    \item Appendix \ref{sec:heatmaps} Sweep Overview Plots
    \begin{itemize}
        \item Figures \ref{fig:shape1_median}-\ref{fig:shape2_cov} Network Shape Plots
        \item Figures \ref{fig:depth1_median}-\ref{fig:depth2_cov} Network Depth Plots
        \item Figures \ref{fig:lr_median}, \ref{fig:lr_cov} Learning Rate Plots
        \item Figures \ref{fig:batch_size_median}, \ref{fig:batch_size_cov} Minibatch Size Plots
        \item Figures \ref{fig:l1_median}, \ref{fig:l1_cov} $L^1$     Regularization Plots
        \item Figures \ref{fig:l2_median}, \ref{fig:l2_cov}  $L^2$ Regularization Plots
        \item Figures \ref{fig:label_noise_median}, \ref{fig:label_noise_cov}  Label Noise Plots
        \item Figures \ref{fig:optimizer_median}, \ref{fig:optimizer_cov} Optimizer Comparison Plots
    \end{itemize}
    \item Appendix \ref{sec:dataset-information} Dataset Information
    \item Appendix \ref{sec:dataset-datasheet} Dataset Datasheet
\end{itemize}

\FloatBarrier
\clearpage
\subsection{Preliminary analysis}
\label{sec:analysis}

\noindent Here we present a preliminary analysis of patterns in the test loss observed across \# parameters, training epoch, learning task, and network shape.
The purpose of this analysis is to demonstrate some consistent patterns that we observe in the data, and evaluate how well those patterns generalize to data with larger numbers of parameters and epochs.
Such a result could strengthen our understanding of deep learning and inform the hyperparameter tuning process.
This is only a peek at what future analysis could illuminate, and uses only a fraction of the data contained within the dataset.

Figure \ref{fig:heatmap} provides contour plot visualization of the median test loss and interquartile range / median over repetitions of each experiment as a function of epoch and the number of free parameters for seven datasets and three network depths from the Primary Sweep.
The contour plots were created using the tricontourf function of Matplotlib utilizing the Delaunay triangulation method \cite{Hunter2007matplotlib}.
Faint vertical banding or jagged contours in this figure may be artifacts of the visualization technique due to the regular spacing of data along the horizontal axis.
Each data point plotted is aggregated over at least twenty repetitions of that experiment with distinct random seeds.
In these plots we observed a similar curvilinear shape of the region of lowest test loss across all training datasets and network depths.
The variation seen in smaller parameter counts may be due in part to initialization stochasticity as suggested by the lottery ticket hypothesis \cite{frankle2018lottery}.
Variation in the epochs before and after the lowest test loss region is may be due to initialization and minibatch stochasticity.

\subsubsection{Fitting and extrapolating analytic curves}
\begin{figure*}[p]
    \makebox[\textwidth][c]{\includegraphics[width=0.85\textwidth]{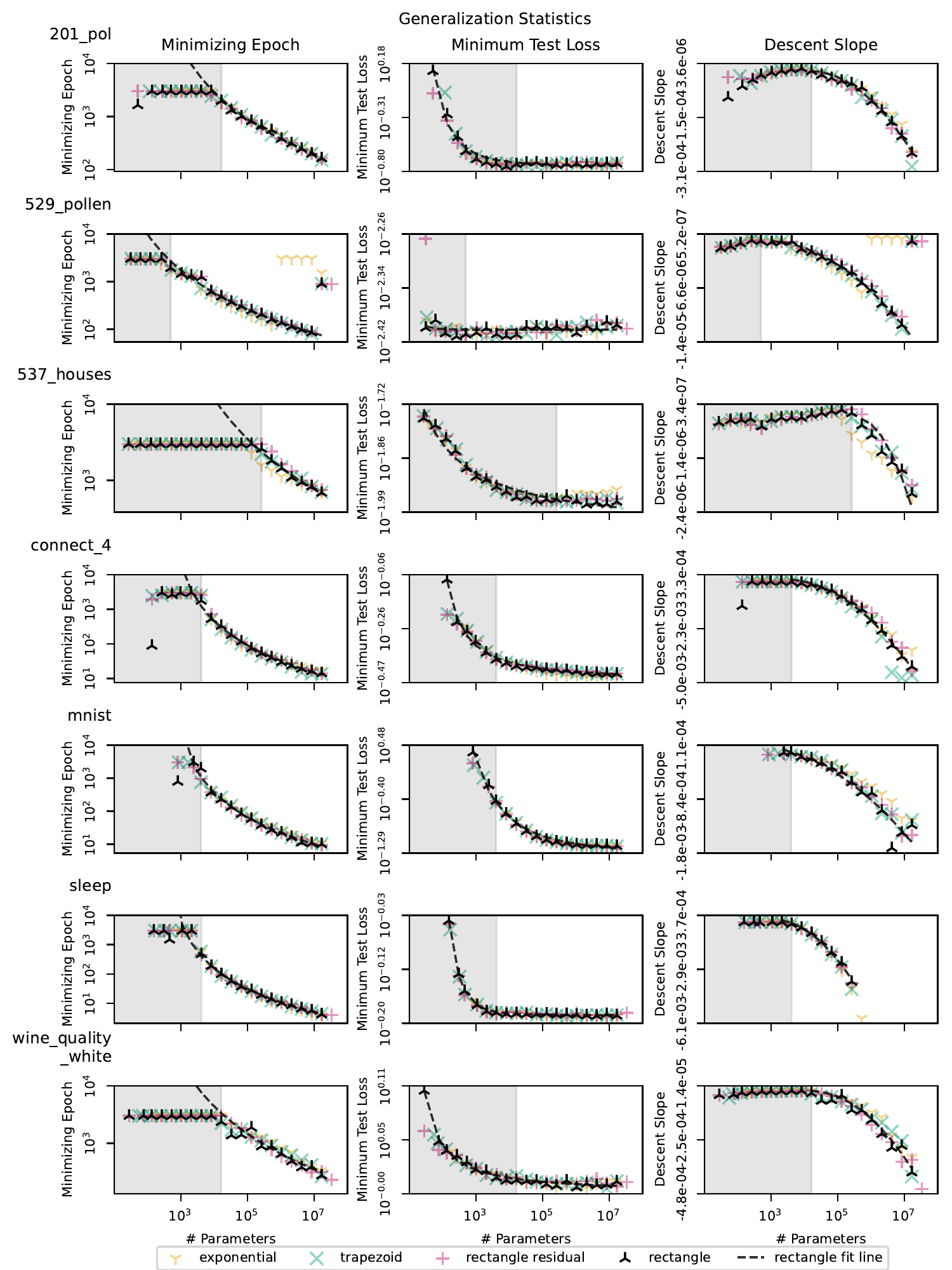}
    \hspace{0.25in}}
    \caption{\textbf{Generalization statistics from the Primary Sweep.}
    Each row depicts statistics for a given task and each column a different generalization statistic.
    For each \# of trainable parameters, the first column plots the number of epochs needed to achieve the minimum test loss, the second column plots the minimum test loss, and the third column the descent slope towards that minimum.
    The dashed line plots Equations \ref{eq:cube_root}, \ref{eq:min_loss_objective}, and \ref{eq:slope_parabola} fit to the rectangular network data points.
    }
    \label{fig:Analysis}
\end{figure*}

\noindent In this section, we seek to establish a relationship between the number of parameters, the smallest loss it can achieve, the number of epochs to do so, and how rapidly it descends to that minimum.
Specifically, we consider the \# of parameters, the median test loss, the epoch at which the median test loss is minimized, and the log-space slope measured between the minimum and the earliest epoch to achieve 10\% greater test loss of the minimum itself.
To illustrate our analysis, we examine 4 shapes and 7 tasks with a depth (number of hidden layers) of 3 from the Primary Sweep and smooth their per-epoch median test loss curves using a linear Savitzky-Golay filter with a window size of 101 \cite{savitzky1964smoothing}.\footnote{For networks trained on the 529\_pollen task we used a window size of 901 and second order polynomial filter to stabilize noisy minimum test losses observed in that task.}
The minimizing epoch, minimum median test loss, and descent slope of the smoothed curves are plotted in Figure \ref{fig:Analysis}.

\newpage
Next, we identified analytic functions which are consistent with this data and can be used to estimate these three values for any parameter count on a given learning task,

\begin{align}
    {\theta_{min}}^{-1/3} &= m\log(\sigma) + b\label{eq:cube_root}\\
    \log(\phi_{min}) &= c(\log(\sigma) - \varsigma_0)^{-p} + \varphi\label{eq:min_loss_objective}\\
    \alpha = \frac{\Delta \log(\phi)}{\Delta \log(\theta)} &= -l (\log(\sigma)-\varsigma_1)^2 + \alpha_0\label{eq:slope_parabola}
\end{align}

where $\theta$ is the epoch; $\theta_{min}$ is the minimizing epoch; $\sigma$ is the number of trainable parameters in the network; $\phi$ is the median test loss; $\phi_{min}$ is the minimum median test loss; $\alpha$ is the log space descent slope approaching $\theta_{min}$; and $m$, $b$, $c$, $p$, $\varphi$, $\varsigma_0$, $\varsigma_1$, and $\alpha_0$ are constant for each task and shape.
Equation \ref{eq:cube_root} estimates that the logarithm of the number of trainable parameters is approximately linearly proportional to the inverse cube root of the minimizing epoch with trade-off slope $m$ and offset $b$.
Equation \ref{eq:min_loss_objective} suggests that log of the minimum test loss has an approximate power law relationship to the log of the parameter count with exponent $-p$, scale $c$, and offset $\varphi$.
$\varsigma_0$ is the log of the minimum number of parameters needed to learn the task, and $\varphi$ is the log of the minimum achievable test loss.
Finally, Equation \ref{eq:slope_parabola} suggests the log slope of test loss as it approaches $\theta_{min}$ is approximately quadratically related to the log of the parameter count where $l$ is a scale factor, and $\alpha_0$ and $\varsigma_1$ are the most gentle slope and the log of the parameter count at which it occurs.

Figure \ref{fig:Analysis} shows non-linear least squares fits as generated with SciPy's least\_squares solver \cite{Scipy} of Equations \ref{eq:cube_root}, \ref{eq:min_loss_objective}, and \ref{eq:slope_parabola} to the rectangular network data points.
To prevent fitting to experiments which might minimize test loss beyond the tested 3000 epochs, we only fit Equation \ref{eq:cube_root} to parameter counts with measured minimums before the 2900th epoch.
The grey region in each plot shades parameter counts with minimums above this threshold.
These generalization performance relationships are similar across all of the network shapes we examined.

\begin{figure*}[p]
    \centering
    \makebox[\textwidth][c]{\includegraphics[width=6in]{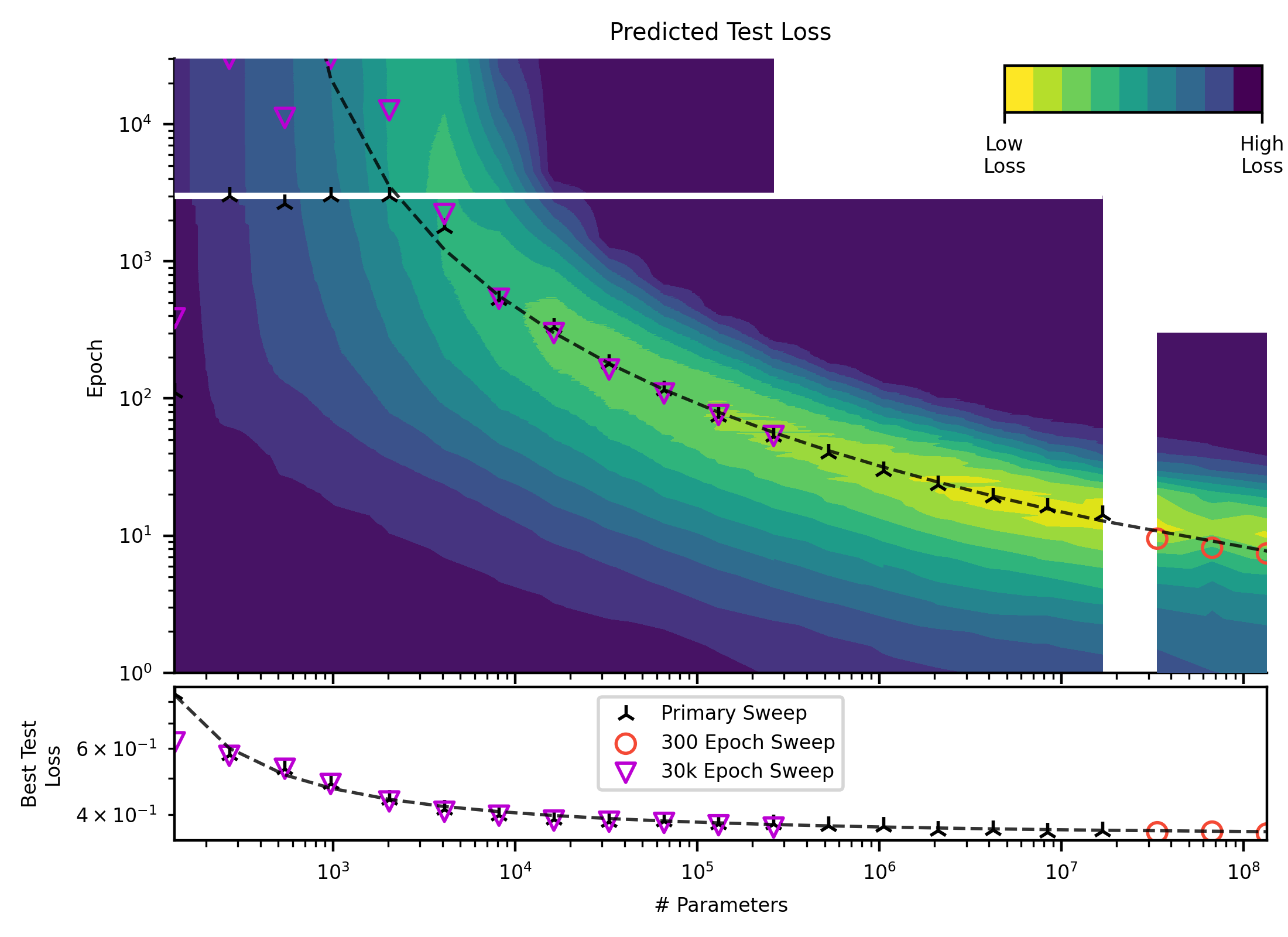}
    \hspace{0.25in}}
    \caption{\textbf{Predictions of generalization statistics in 300 and 30k Epoch sweeps.} 
    The top half of the figure shows log-scaled data from three sweeps for a rectangular network trained on connect\_4. 
    The black dotted line depicts Equation \ref{eq:cube_root} fit to Primary Sweep data.
    Three different icons indicate the locations of test loss minimums in each sweep.
    The bottom plot shows measured test loss minimums for each sweep where the dotted black line represents Equation \ref{eq:min_loss_objective} fit to only the Primary Sweep.
    }
    \label{fig:Prediction}
\end{figure*}

After fitting functions to the generalization statistics of the Primary Sweep, we sought to test our fit functions on data beyond that used to identify these relationships.
Figure \ref{fig:Prediction} shows the outcome of visualizing those tests on the connect\_4 task, with the black dashed line indicating the fit of Equation \ref{eq:cube_root} and \ref{eq:min_loss_objective} to the Primary Sweep data from Figure \ref{fig:Analysis} overlaid on the corresponding 30k Epoch Sweep and 300 Epoch Sweep heatmaps.
Equations \ref{eq:cube_root} and \ref{eq:min_loss_objective} provide promisingly reasonable estimates of minimum test loss locations for networks trained in these additional sweeps.
This visualization serves as a first step in validating Equations \ref{eq:cube_root} and \ref{eq:min_loss_objective} as general trends in the deep learning process.  
\textbf{The preliminary analysis presented here, while promising, only utilizes a small fraction of the dataset and the basis for these relationships is solely empirical.}

\clearpage
\subsection{Sweep Overview Plots}
\label{sec:heatmaps}

\begin{figure*}[p]
    \centering
    \includegraphics[height=8in, width=7in]{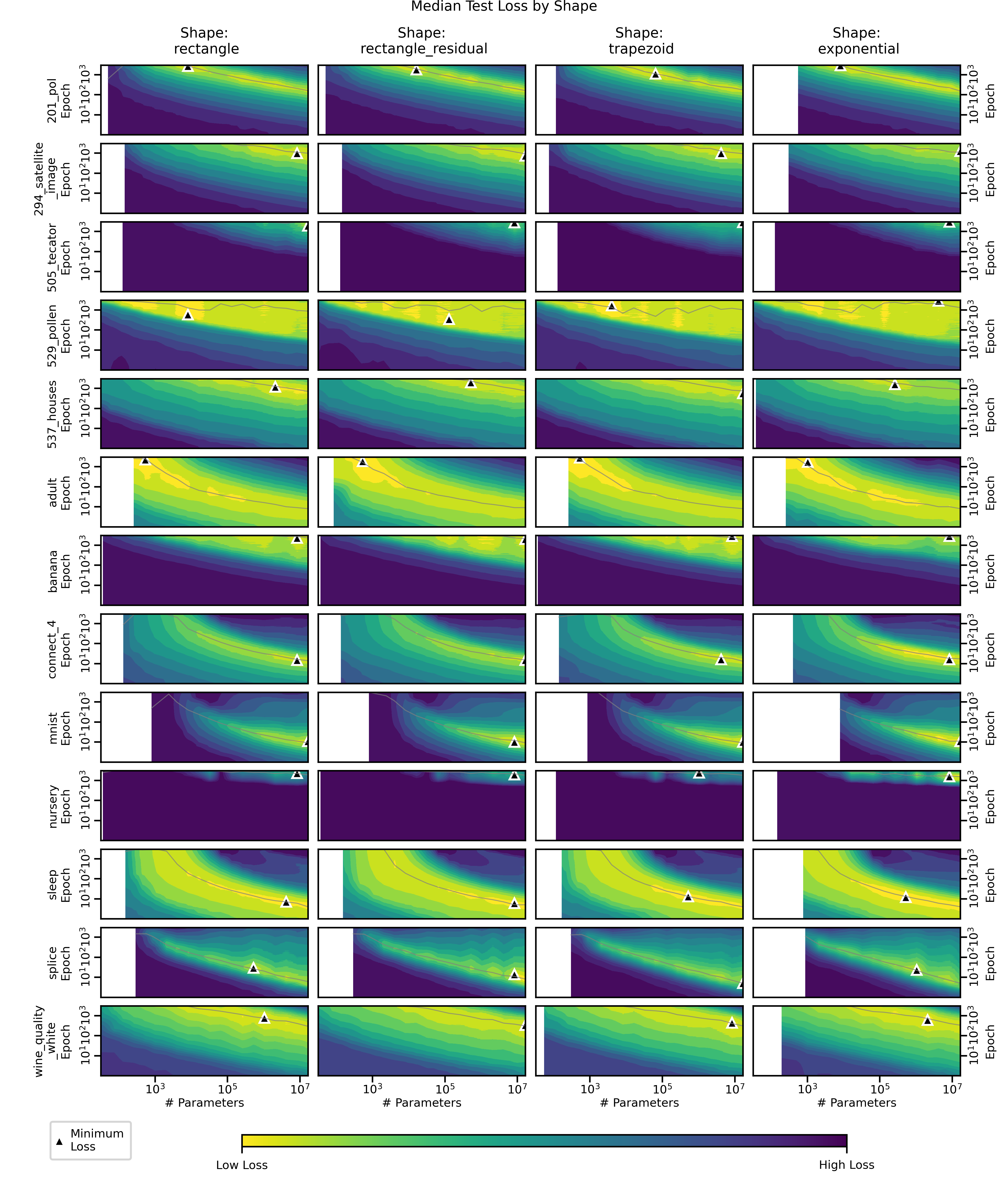} %
    \caption{\textbf{Median test loss for tasks and the first half of shapes in the Primary Sweep.} Patterns are fairly consistent across shapes. The ``\textit{butter-zone}`` of low median test loss forms a curvilinear shape in log-space for all tasks.}
    \label{fig:shape1_median}
\end{figure*}

\begin{figure*}[p]
    \centering
    \includegraphics[height=8.5in, width=7.25in]{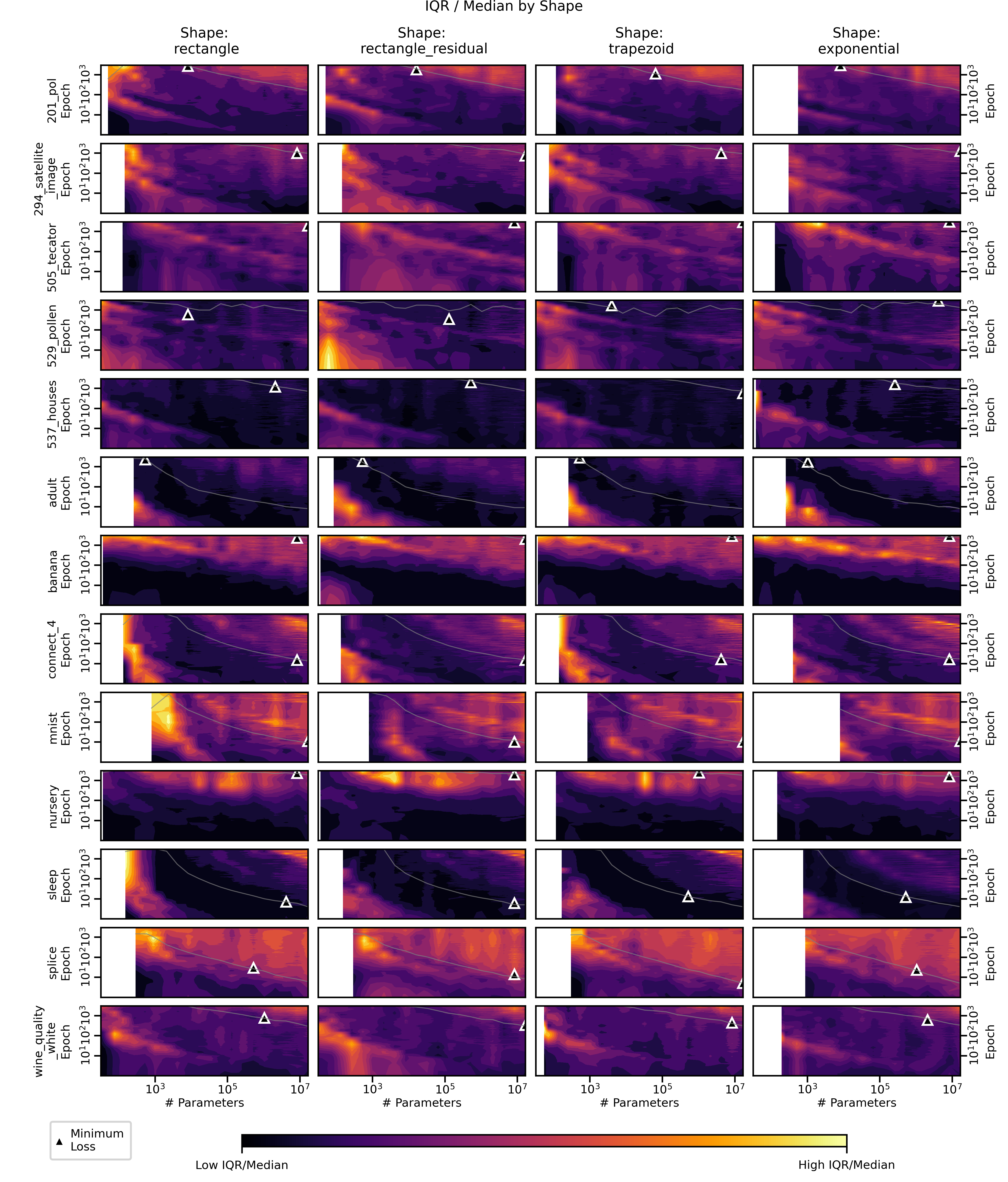}
    \caption{\textbf{Interquartile range / median for tasks and the first half of shapes in the Primary Sweep.} 
    Variation decreased in the \textit{butter-zone} with higher spread in regions of over- and under-fitting. 
    Small parameter counts generally had greater variation in test loss. 
    This pattern may support the lottery ticket hypothesis which predicts that increasing the number of parameters in a network increases the probability that gradient descent converges.}
    \label{fig:shape1_cov}
\end{figure*}

\begin{figure*}[p]
    \centering
    \includegraphics[height=8.5in, width=7.25in]{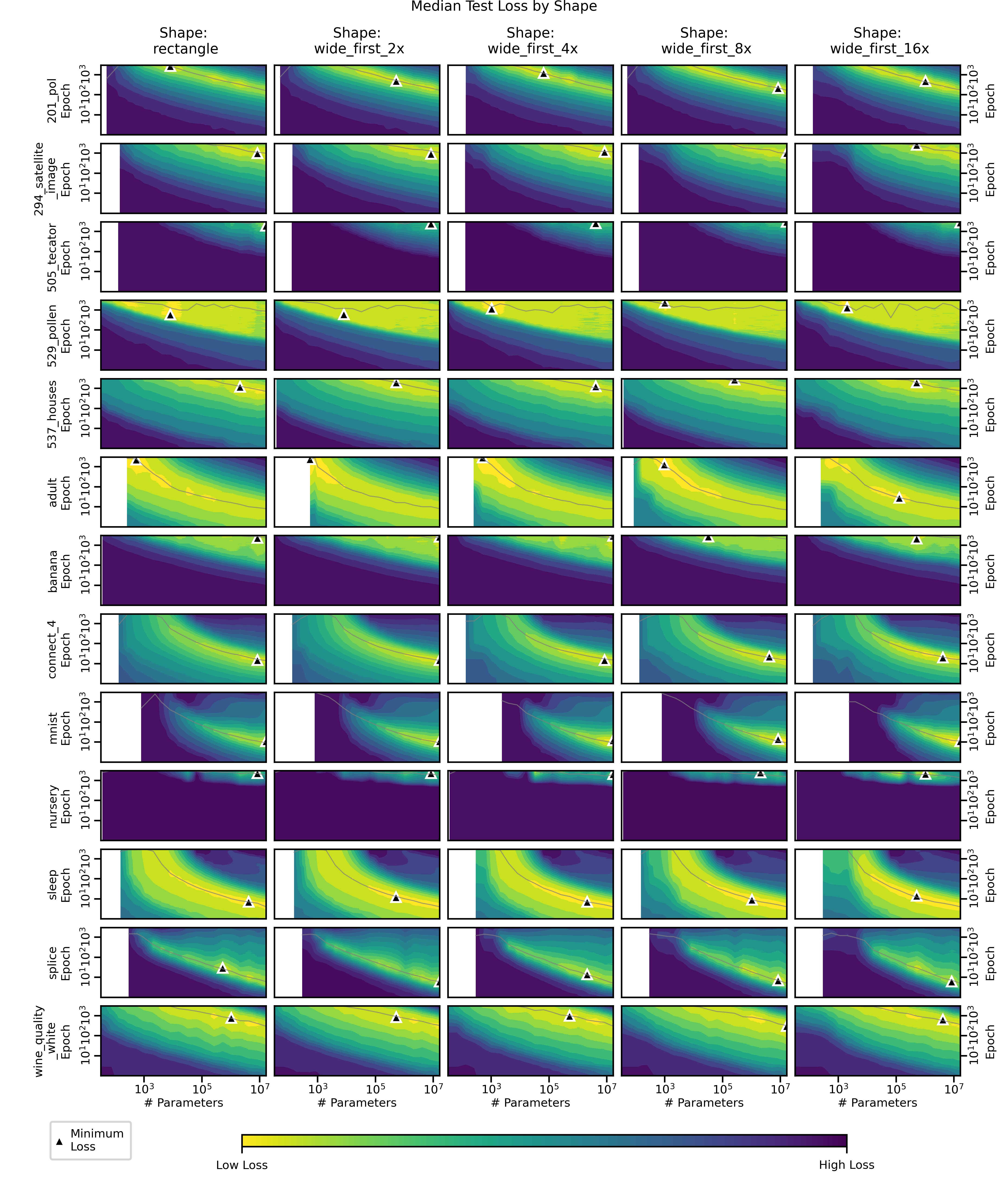}
    \caption{\textbf{Median test loss for tasks and the second half of shapes in the Primary Sweep.} Similar patterns emerge as in Figure \ref{fig:shape1_median}.}
    \label{fig:shape2_median}
\end{figure*}

\begin{figure*}[p]
    \centering
    \includegraphics[height=8.5in, width=7.25in]{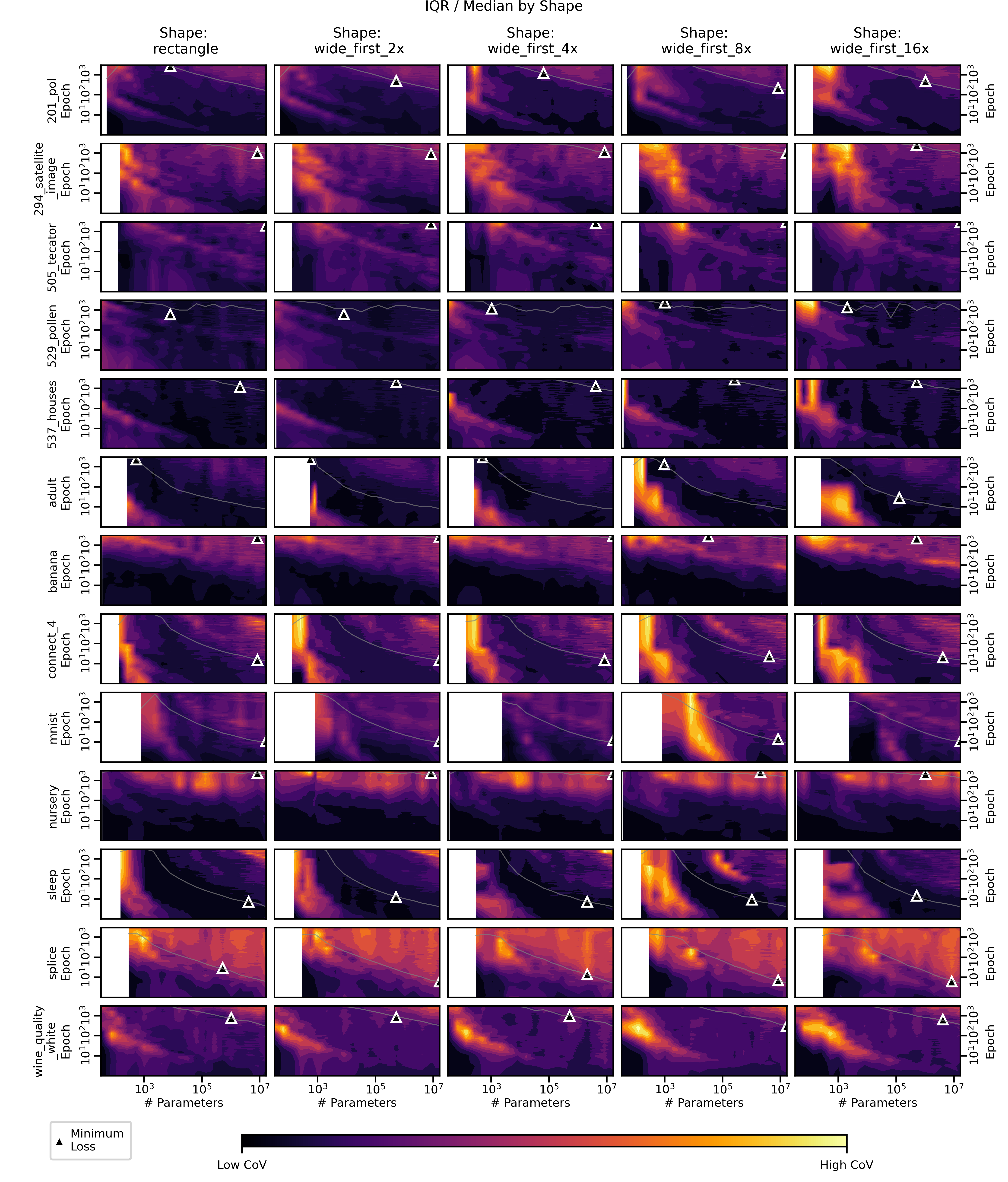}
    \caption{\textbf{Interquartile range / median for tasks and the second half of shapes in the Primary Sweep.} Similar patterns emerge as in Figure \ref{fig:shape1_cov}.}
    \label{fig:shape2_cov}
\end{figure*}

\begin{figure*}[p]
    \centering
    \includegraphics[height=8.5in, width=7.25in]{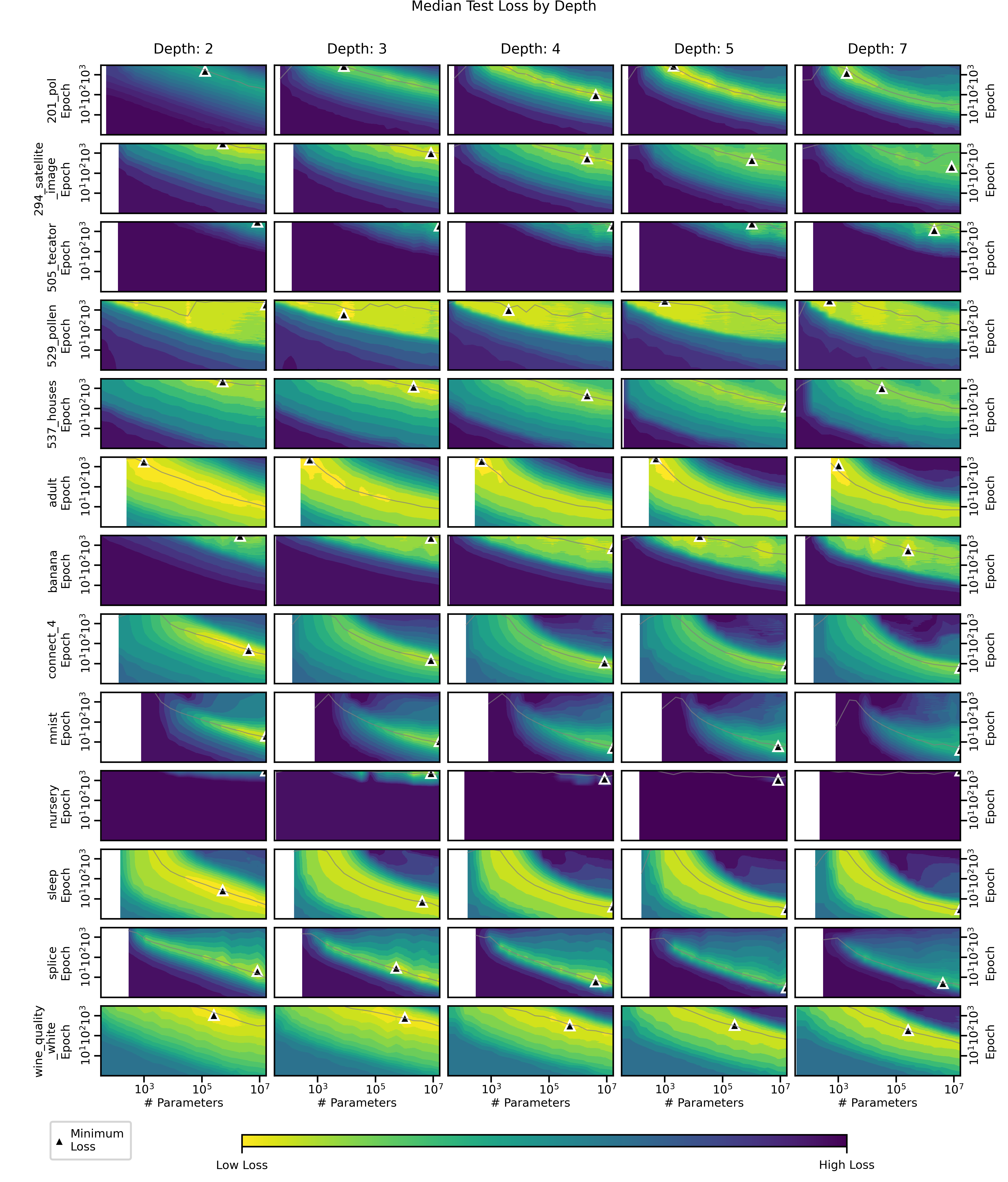}
    \caption{\textbf{Median test loss for tasks and the first half of depths in the Primary Sweep.} 
    The depth at which the lowest median test loss was achieved varied by task.
    For example, very shallow networks trained on mnist generalized best, but the reverse is true for 201\_pol where deeper networks generalized better.}
    \label{fig:depth1_median}
\end{figure*}

\begin{figure*}[p]
    \centering
    \includegraphics[height=8.5in, width=7.25in]{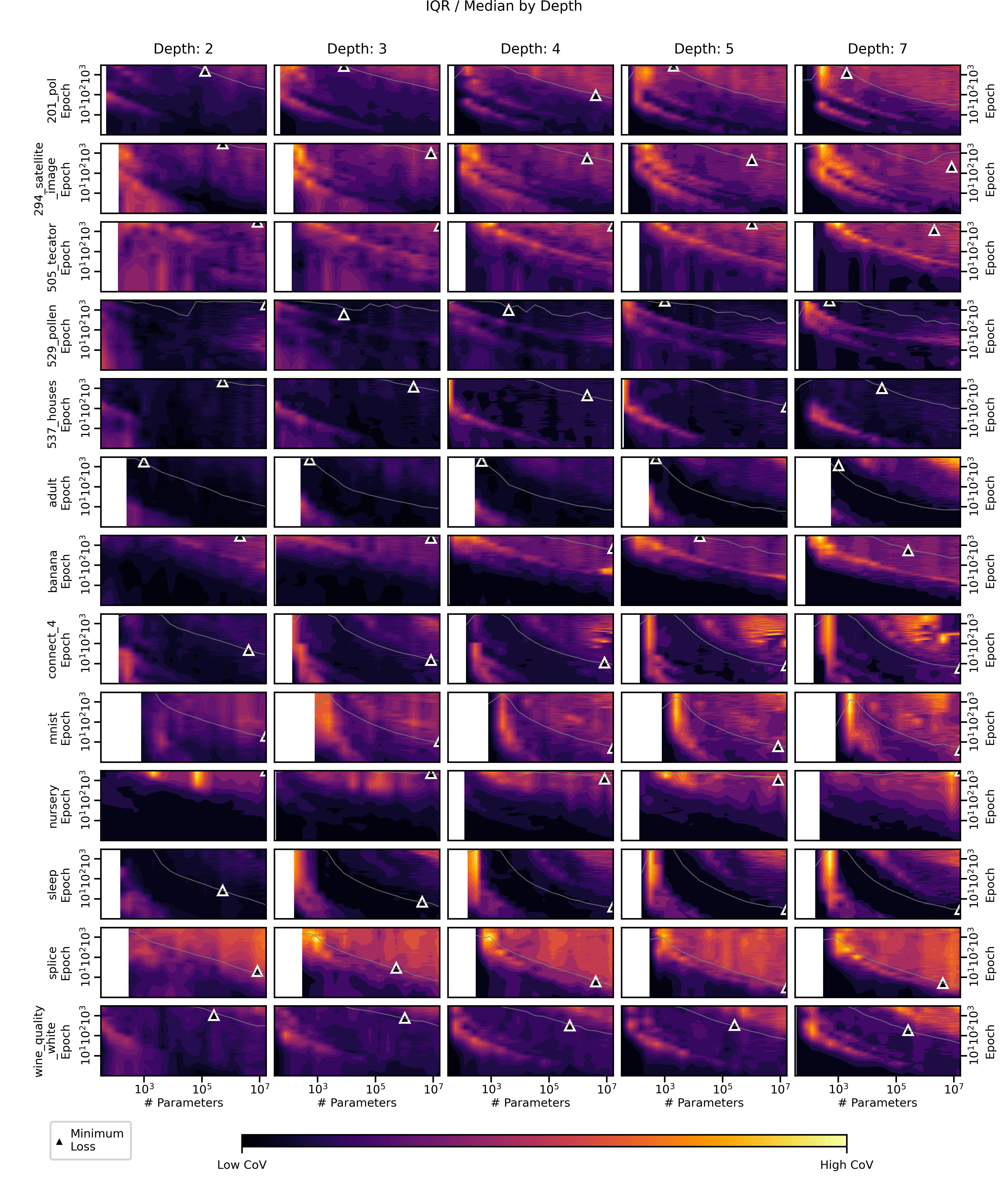}
    \caption{\textbf{Interquartile range / median for tasks and the first half of depths in the Primary Sweep.} 
    Variations in test loss generally increased with increased depth.
    Particularly, the vertical orange band of high spread that covers small networks tends to shift rightward with depth suggesting that more parameters may be needed to stabilize training at higher depths.
    Additionally, variability in overtrained regions increased more for deeper networks.}
    \label{fig:depth1_cov}
\end{figure*}

\begin{figure*}[p]
    \centering
    \includegraphics[height=8.5in, width=7.25in]{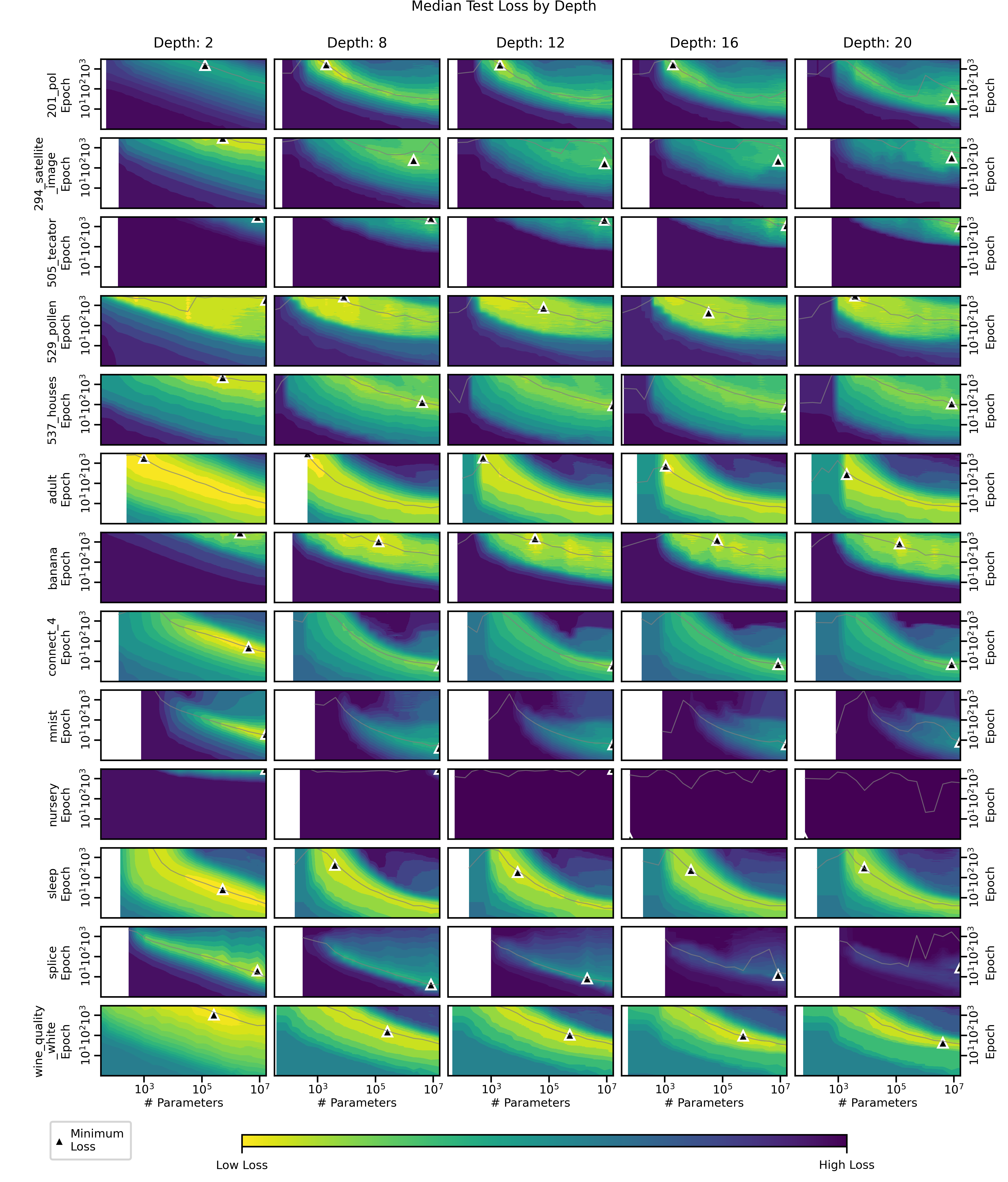}
    \caption{\textbf{Median test loss for tasks and the second half of depths in the Primary Sweep.} Depth two is also shown for comparison. Similar patterns emerge as in Figure \ref{fig:depth1_median}.}
    \label{fig:depth2_median}
\end{figure*}

\begin{figure*}[p]
    \centering
    \includegraphics[height=8.5in, width=7.25in]{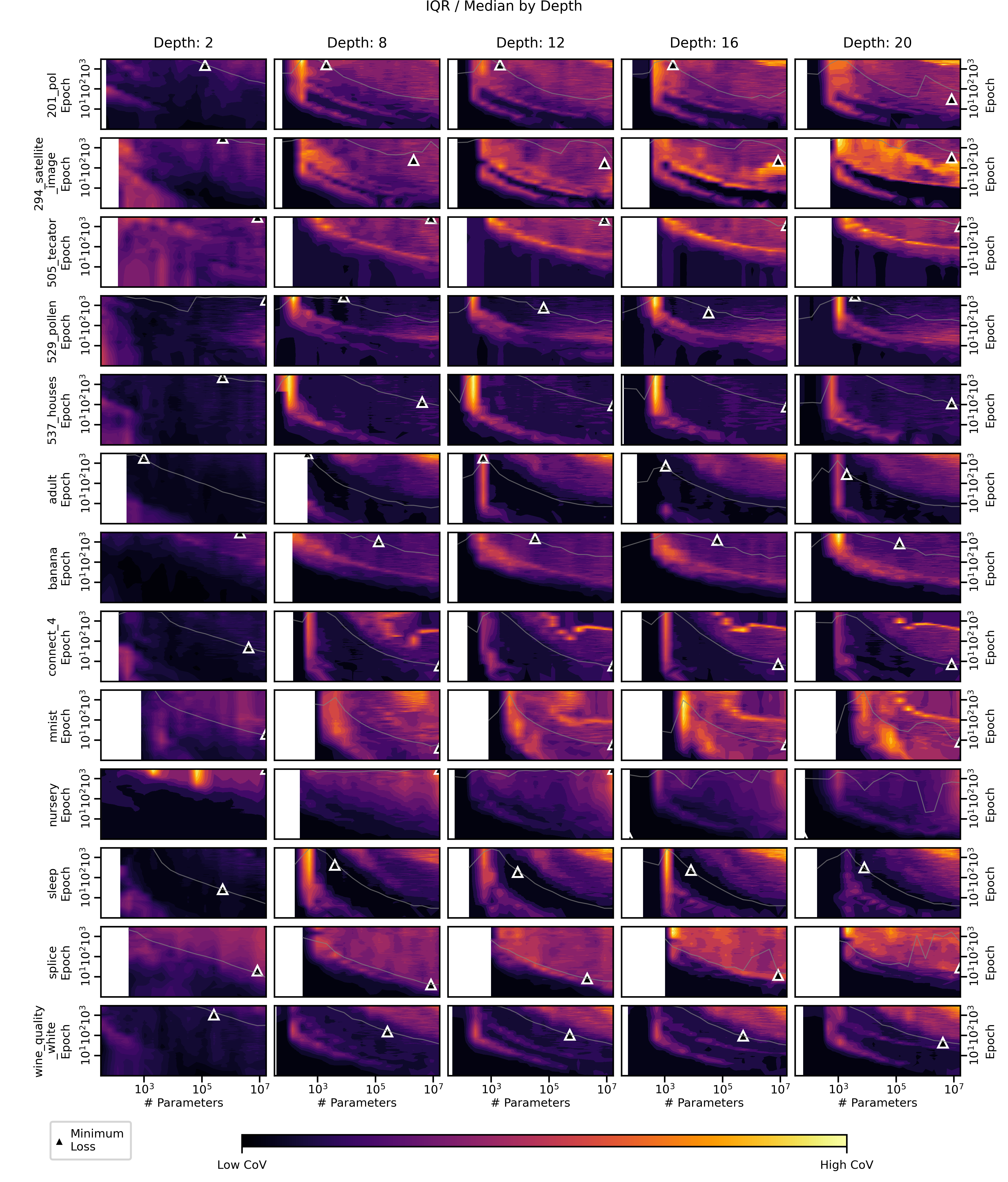}
    \caption{\textbf{Interquartile range / median for tasks and the second half of depths in the Primary Sweep.} Depth two is also shown for comparison. Similar patterns emerge as in Figure \ref{fig:depth1_cov}.}
    \label{fig:depth2_cov}
\end{figure*}

\begin{figure*}[p]
    \centering
    \includegraphics[height=8.5in, width=7.25in]{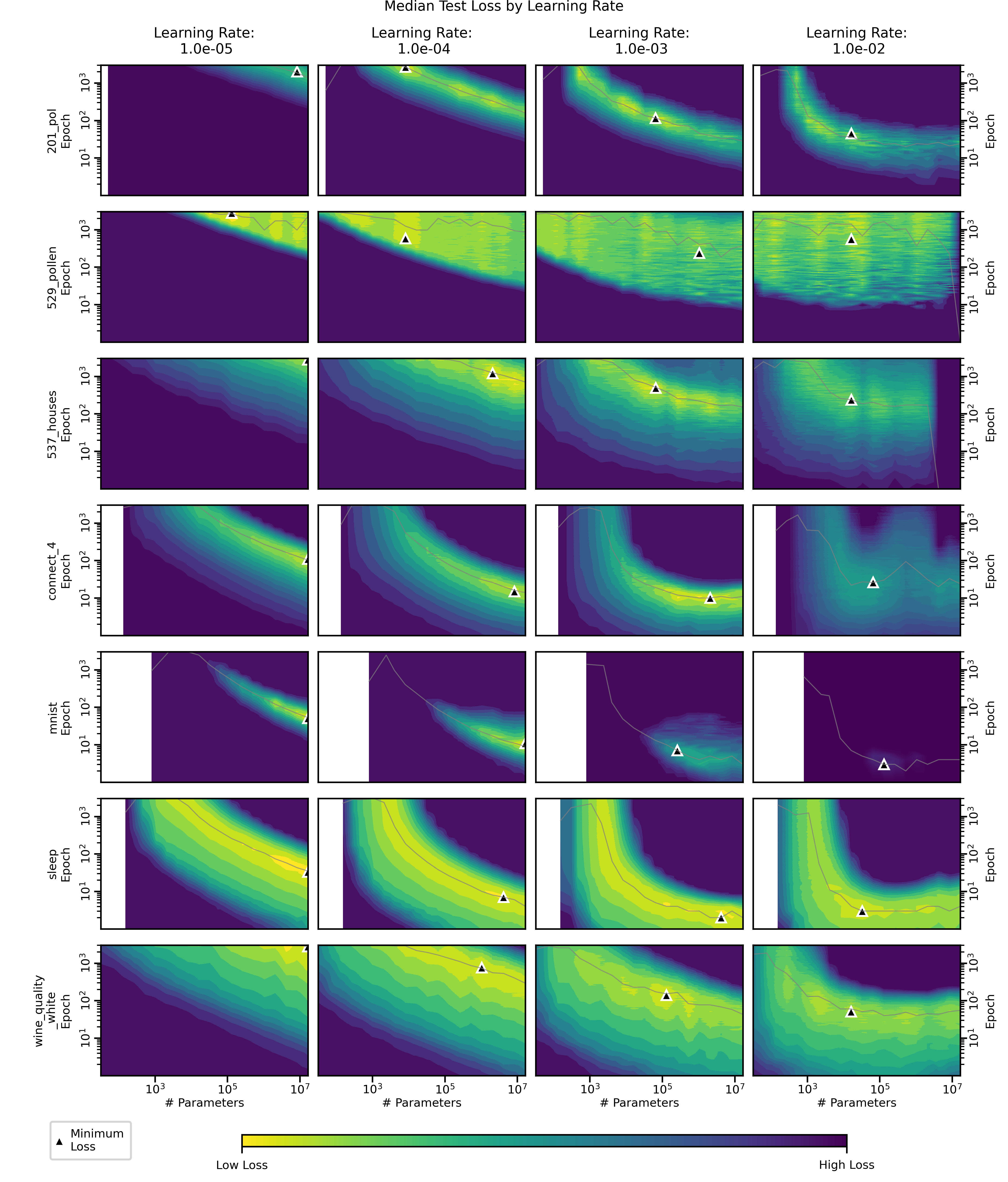}
    \caption{\textbf{Median test loss for tasks and learning rates in the Learning Rate Sweep.} 
    Higher learning rate networks decreased median test loss more rapidly early on, but sometimes did not achieve as low of a minimum median test loss.
    This can be seen in the case of connect\_4 and mnist, where fast learning rates coincided with high minimum losses when compared to those achieved using slower rates.
    Additionally, larger networks showed more sensitivity to the learning rate than did smaller networks.
    We restrict the contours to losses less than 2 times the minimum loss.}
    \label{fig:lr_median}
\end{figure*}

\begin{figure*}[p]
    \centering
    \includegraphics[height=8.5in, width=7.25in]{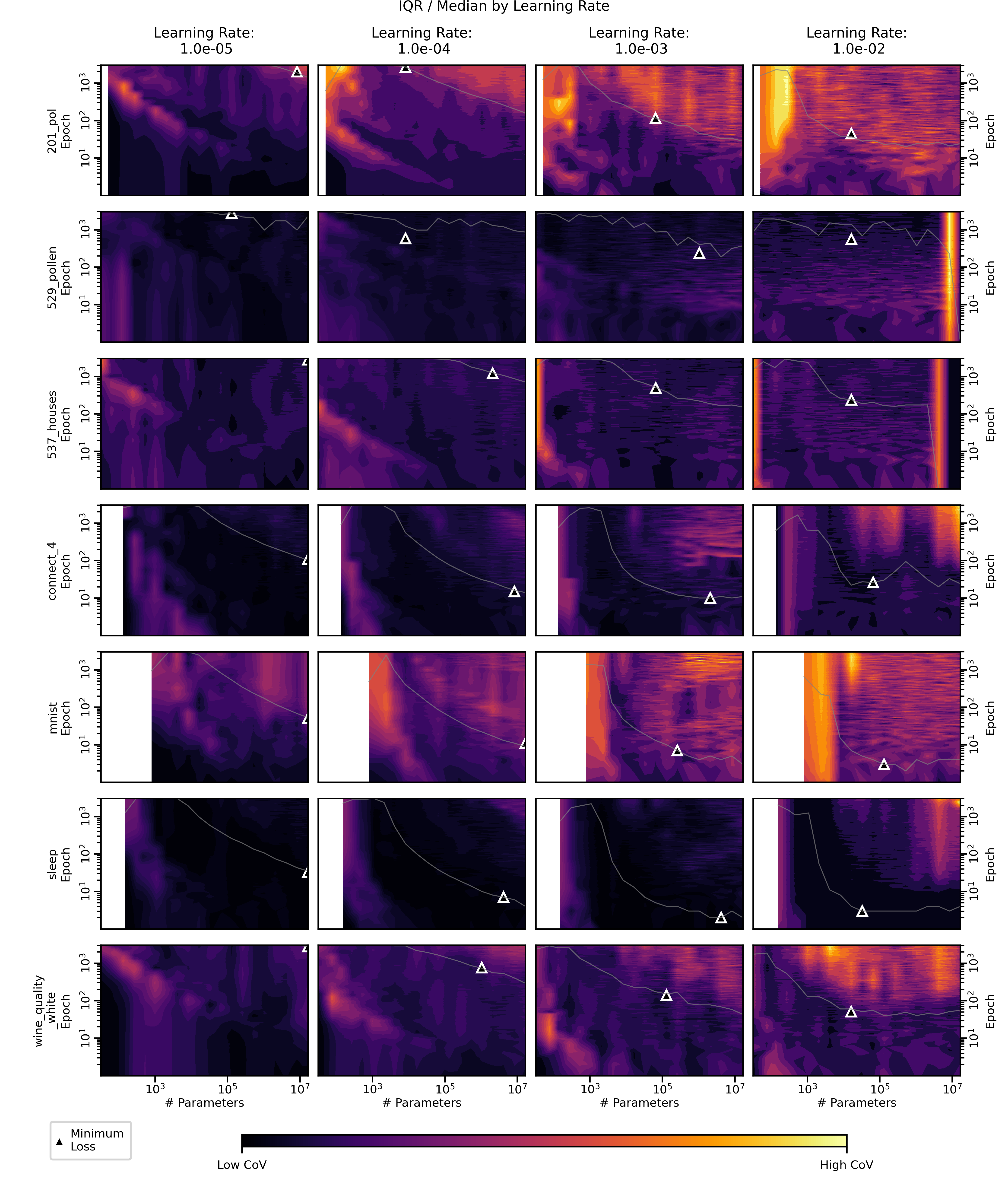}
    \caption{\textbf{Interquartile range / median for tasks and learning rates in the Learning Rate Sweep.} 
    Higher learning rates coincided with elevated variation across experiment repetitions.
    This is especially apparent when networks were overtrained (larger parameter counts and higher epochs). 
    For some tasks (e.g. 529\_pollen and 537\_houses) the highest learning rate prevented training altogether at the largest parameter counts where variations between repetitions become very large.}
    \label{fig:lr_cov}
\end{figure*}

\begin{figure*}[p]
    \centering
    \includegraphics[height=8.5in, width=7.25in]{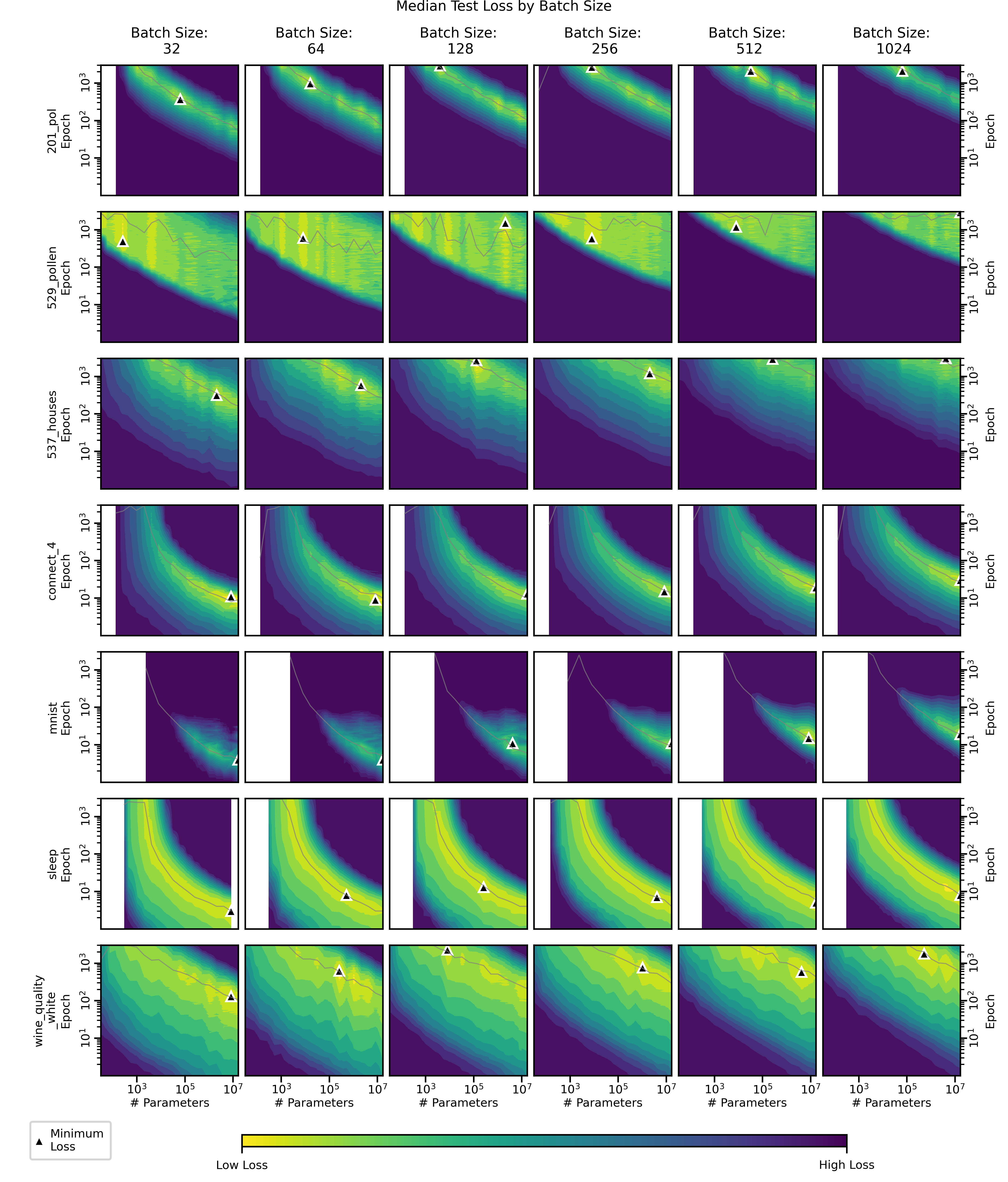}
    \caption{\textbf{Median test loss for tasks and batch sizes in the Batch Size Sweep.} 
    For all tasks shown here, increasing the batch size coincided with an increase in the number of epochs taken to minimize test loss.
    In fact, the entire \textit{butter-zone} tends to shift upwards as the batch size increases. 
    We restrict the contours to losses less than 2 times the minimum loss.}
    \label{fig:batch_size_median}
\end{figure*}

\begin{figure*}[p]
    \centering
    \includegraphics[height=8.5in, width=7.25in]{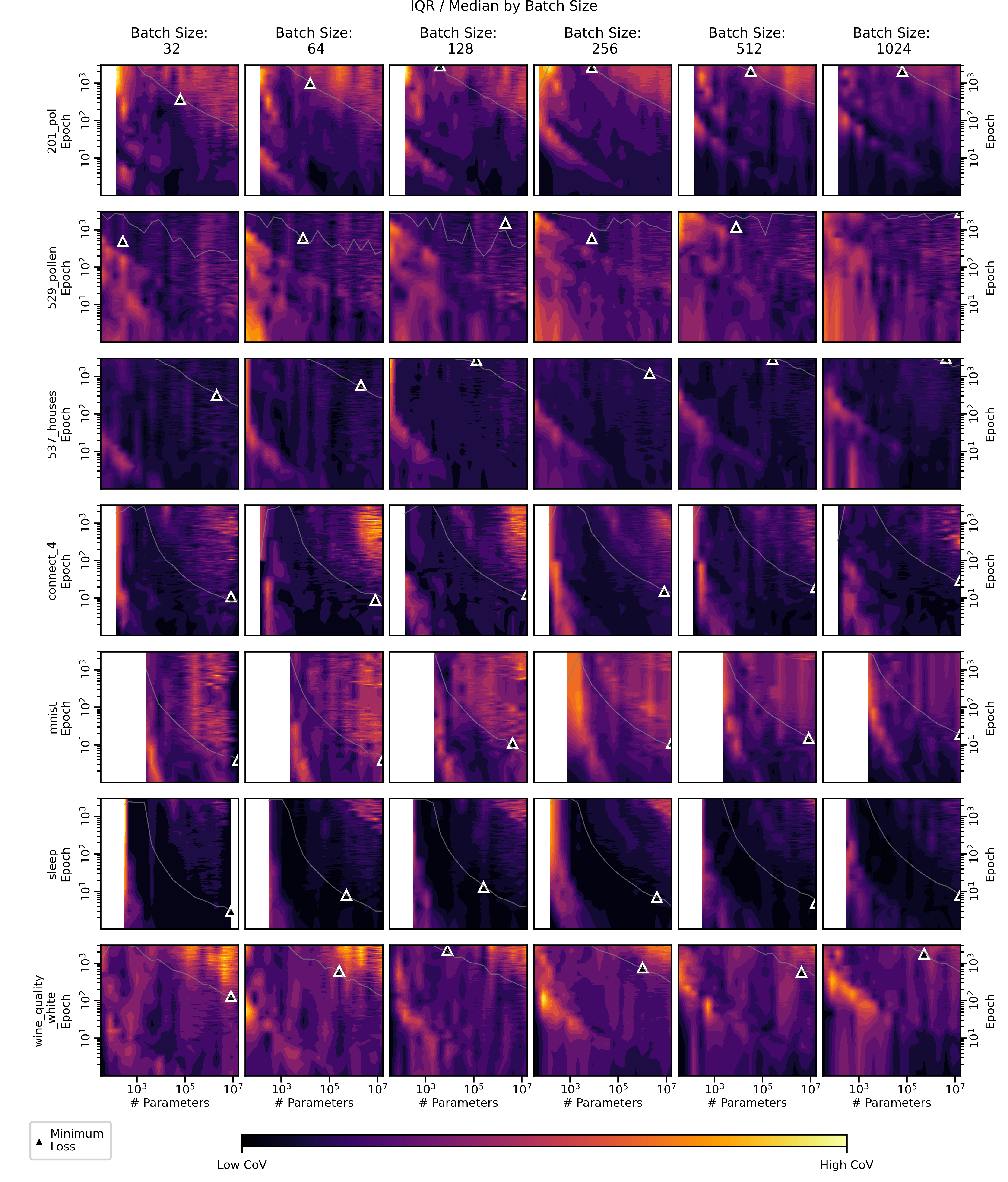}
    \caption{\textbf{Interquartile range / median for tasks and batch sizes in the Batch Size Sweep.} 
    Because of the upward \textit{butter-zone} shifts observed when increasing the batch size, small batch size plots show more overtraining variation and large batch size plots show more undertraining variation.
    }
    \label{fig:batch_size_cov}
\end{figure*}

\begin{figure*}[p]
    \centering
    \includegraphics[height=8.5in, width=7.25in]{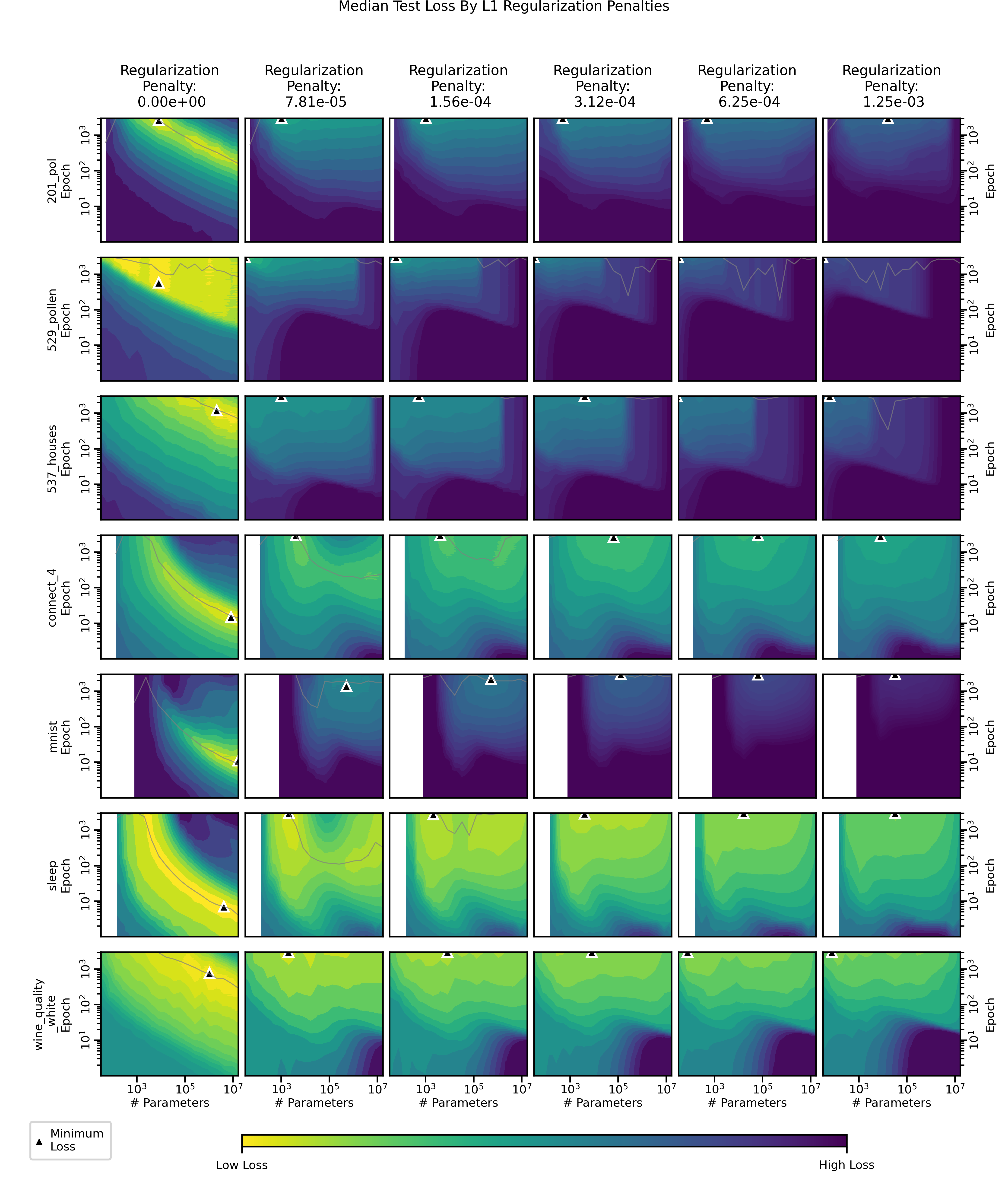}
    \caption{\textbf{Median test loss for tasks and $L^1$ regularization penalty levels in the Regularization Sweep.} 
    Increasing the $L^1$ regularization penalty generally increased the minimum test loss achieved, and increased the number of epochs to achieve a target test loss. 
    High penalties coincided with high test losses in large networks trained for small numbers of epochs.
    $L^1$ regularization reduced or eliminated overtraining in the experiments visualized here. 
    In a few cases, regularization seems to `bend' the butter-zone upwards for higher parameter counts, delaying learning.
    More investigation is warranted to understand this effect.
    }
    \label{fig:l1_median}
\end{figure*}

\begin{figure*}[p]
    \centering
    \includegraphics[height=8.5in, width=7.25in]{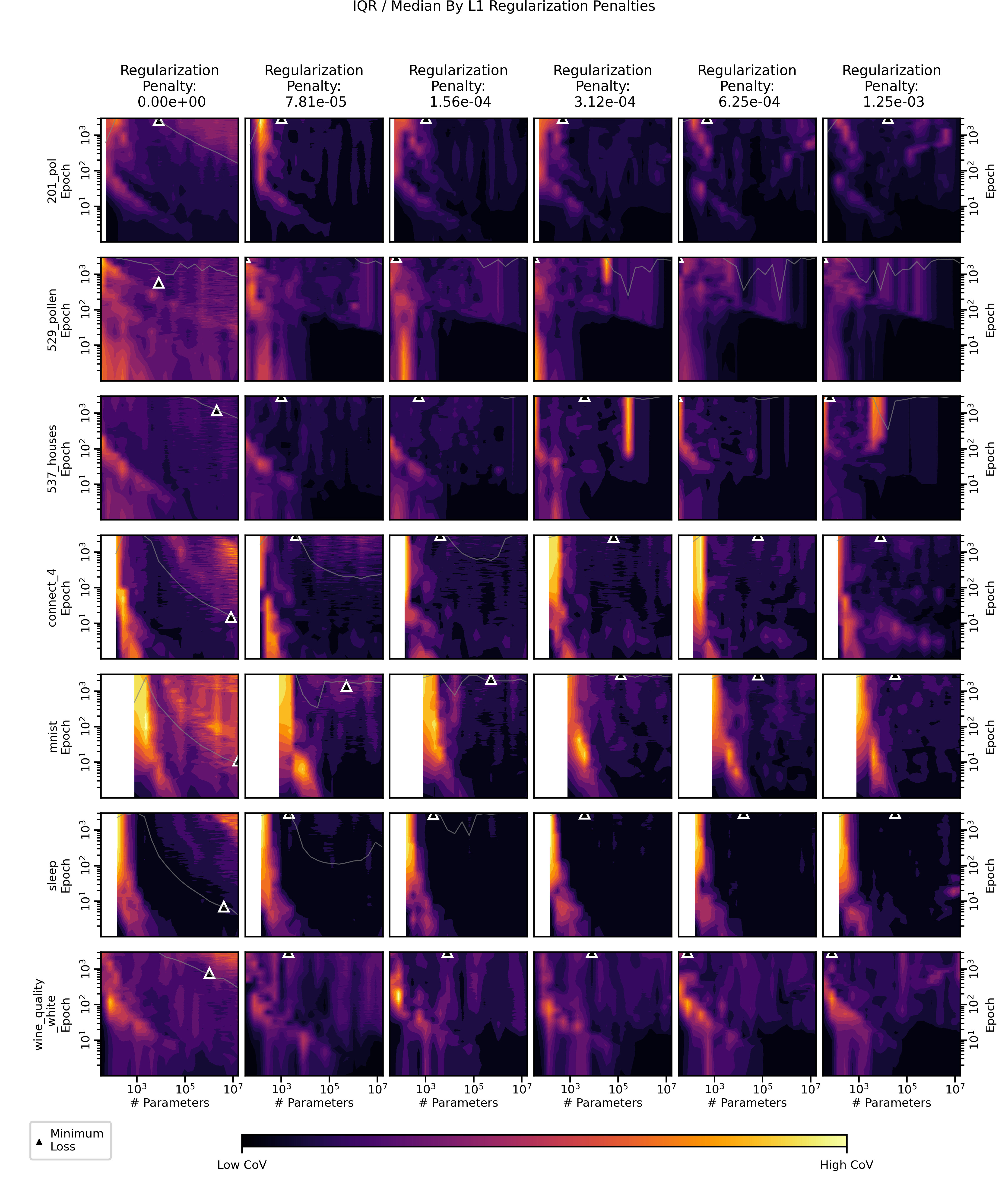}
    \caption{\textbf{Interquartile range / median for tasks and $L^1$ regularization penalties in the Regularization Sweep.} 
    Larger $L^1$ penalties coincided with less variation between repetitions.
    }
    \label{fig:l1_cov}
\end{figure*}

\begin{figure*}[p]
    \centering
    \includegraphics[height=8.5in, width=7.25in]{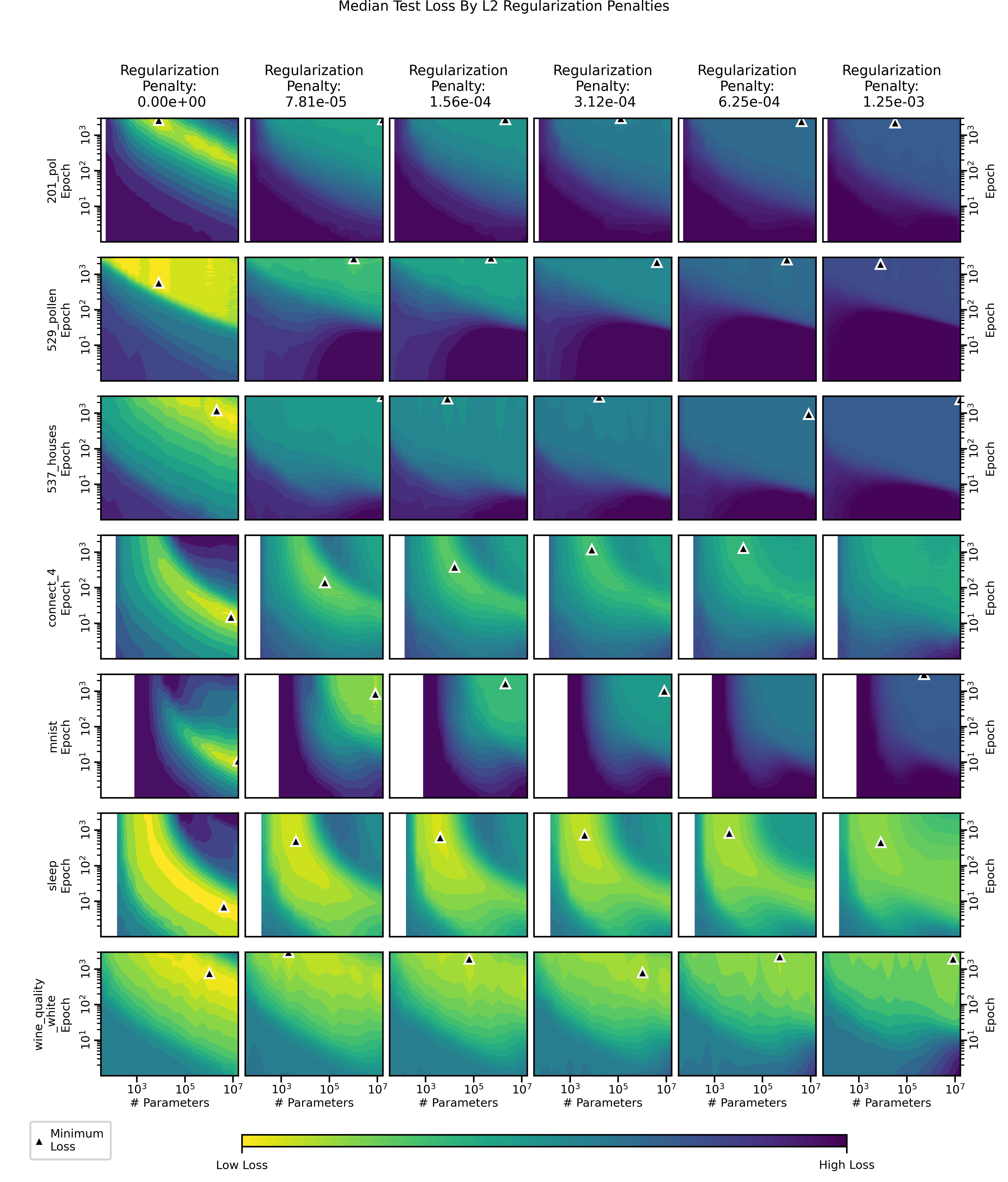}
    \caption{\textbf{Median test loss for tasks and $L^2$ regularization penalty levels in the Regularization Sweep.} 
    Increasing the $L^2$ regularization penalty generally increased the minimum test loss achieved, and increased the number of epochs to achieve a target test loss. 
    Generalization was poor when high penalties were applied on large networks trained for small numbers of epochs.
    However, $L^2$ regularization reduced or eliminated overtraining in the experiments visualized here. 
    Similarly to Figure \ref{fig:l1_median}, this figure is disproportionately affected by the addition of a regularization penalty to the test loss.
    }
    \label{fig:l2_median}
\end{figure*}

\begin{figure*}[p]
    \centering
    \includegraphics[height=8.5in, width=7.25in]{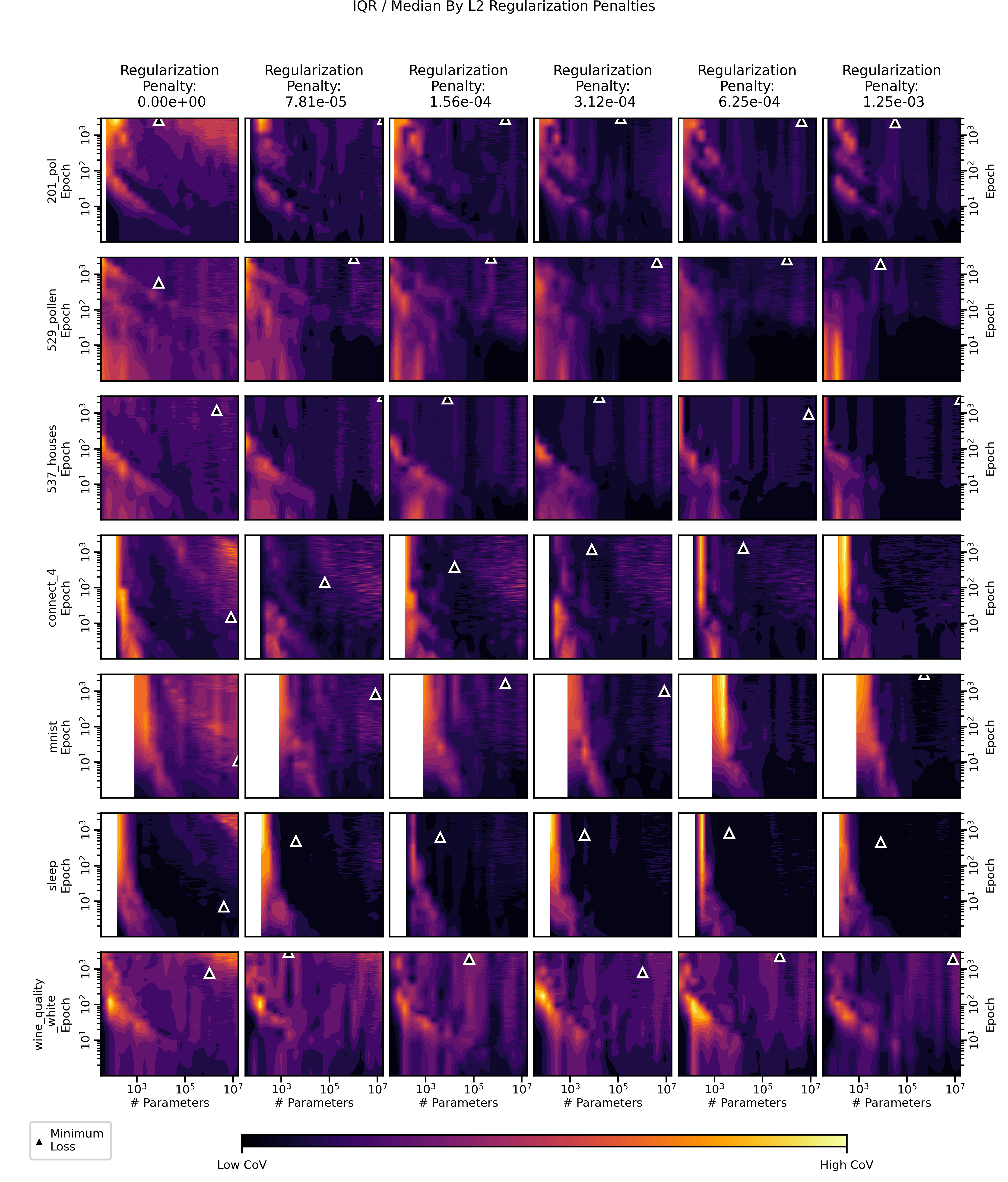}
    \caption{\textbf{Interquartile range / median for tasks and $L^2$ regularization penalties in the Regularization Sweep.} 
    Increased $L^2$ penalties decreased the variation between repetitions.
    }
    \label{fig:l2_cov}
\end{figure*}

\begin{figure*}[p]
    \centering
    \includegraphics[height=8.5in, width=7.25in]{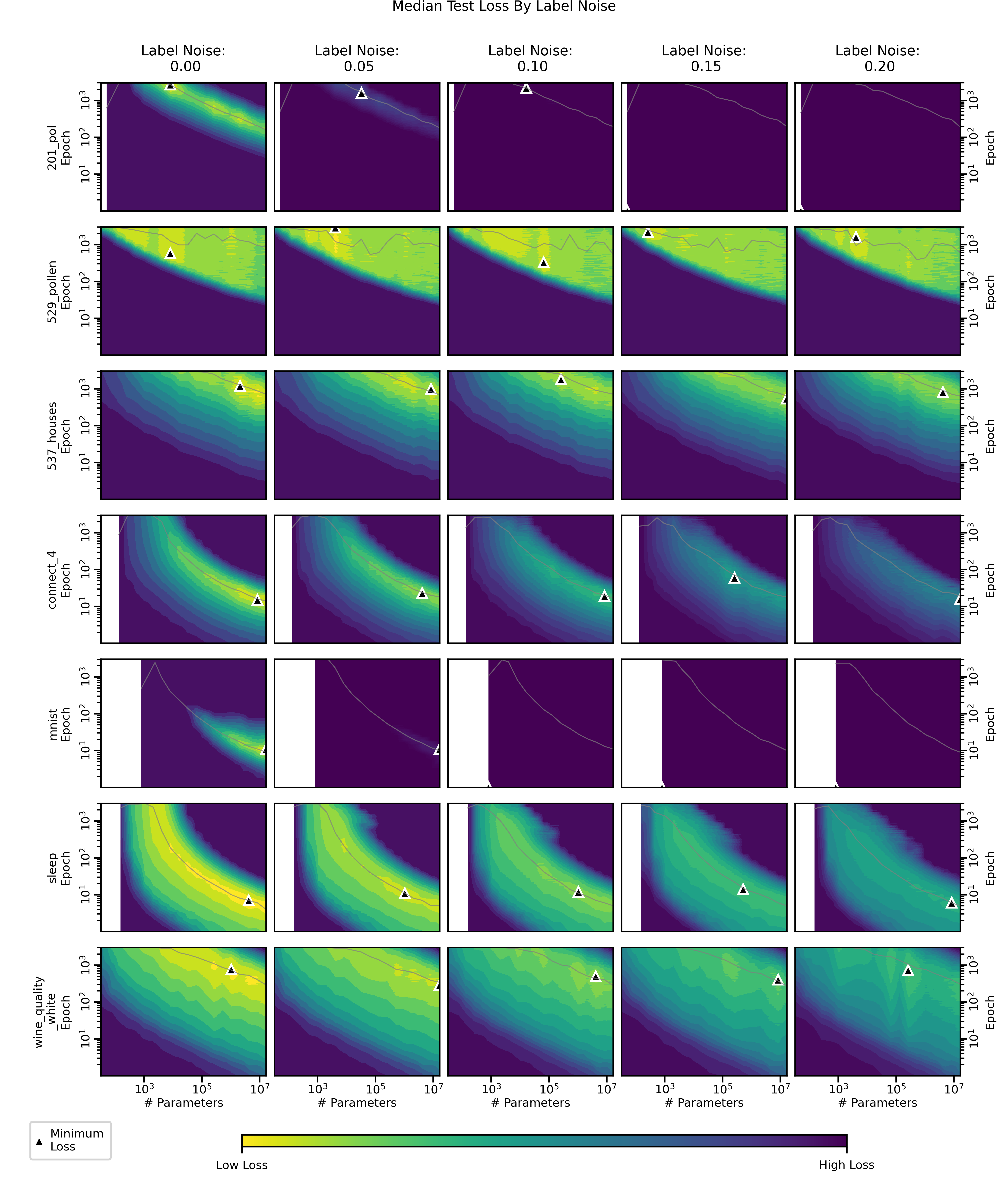}
    \caption{\textbf{Median test loss for tasks and label noise levels in the Label Noise Sweep.} 
    Increased label noise levels affected each task in a different way. 
    Increasing the label noise for mnist, even by 5\% has a large adverse impact on training. 
    However, larger amounts of label noise have little or no effect on test loss for the 529\_pollen task. 
    We restrict the contours to losses less than 2 times the minimum loss.}
    \label{fig:label_noise_median}
\end{figure*}

\begin{figure*}[p]
    \centering
    \includegraphics[height=8.5in, width=7.25in]{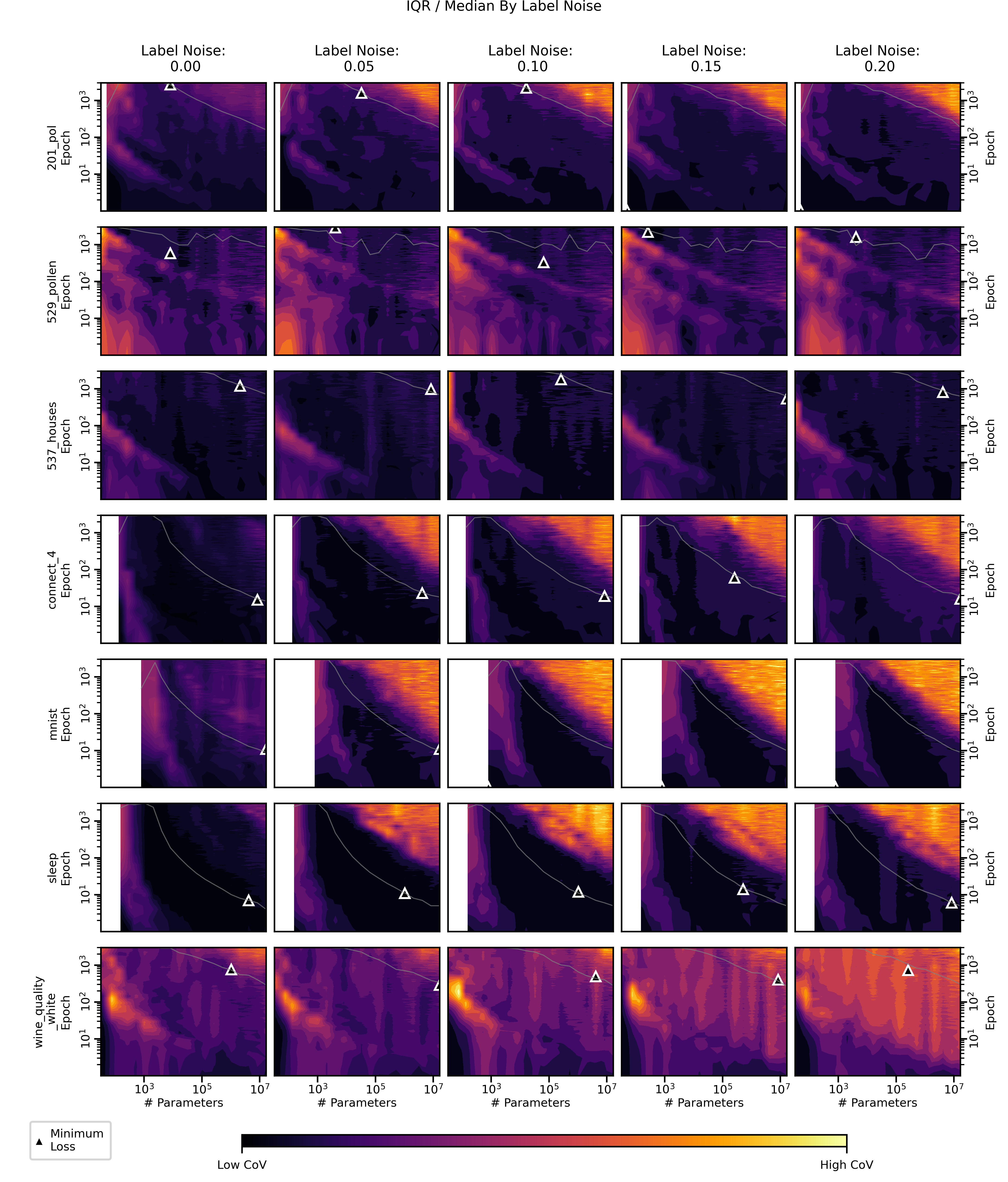}
    \caption{\textbf{Interquartile range / median for tasks and label noise levels in the Label Noise Sweep.} 
    High label noise levels produced high variations in test loss of overtrained networks.}
    \label{fig:label_noise_cov}
\end{figure*}

\begin{figure*}[p]
    \centering
    \includegraphics[height=8.5in, width=7.25in]{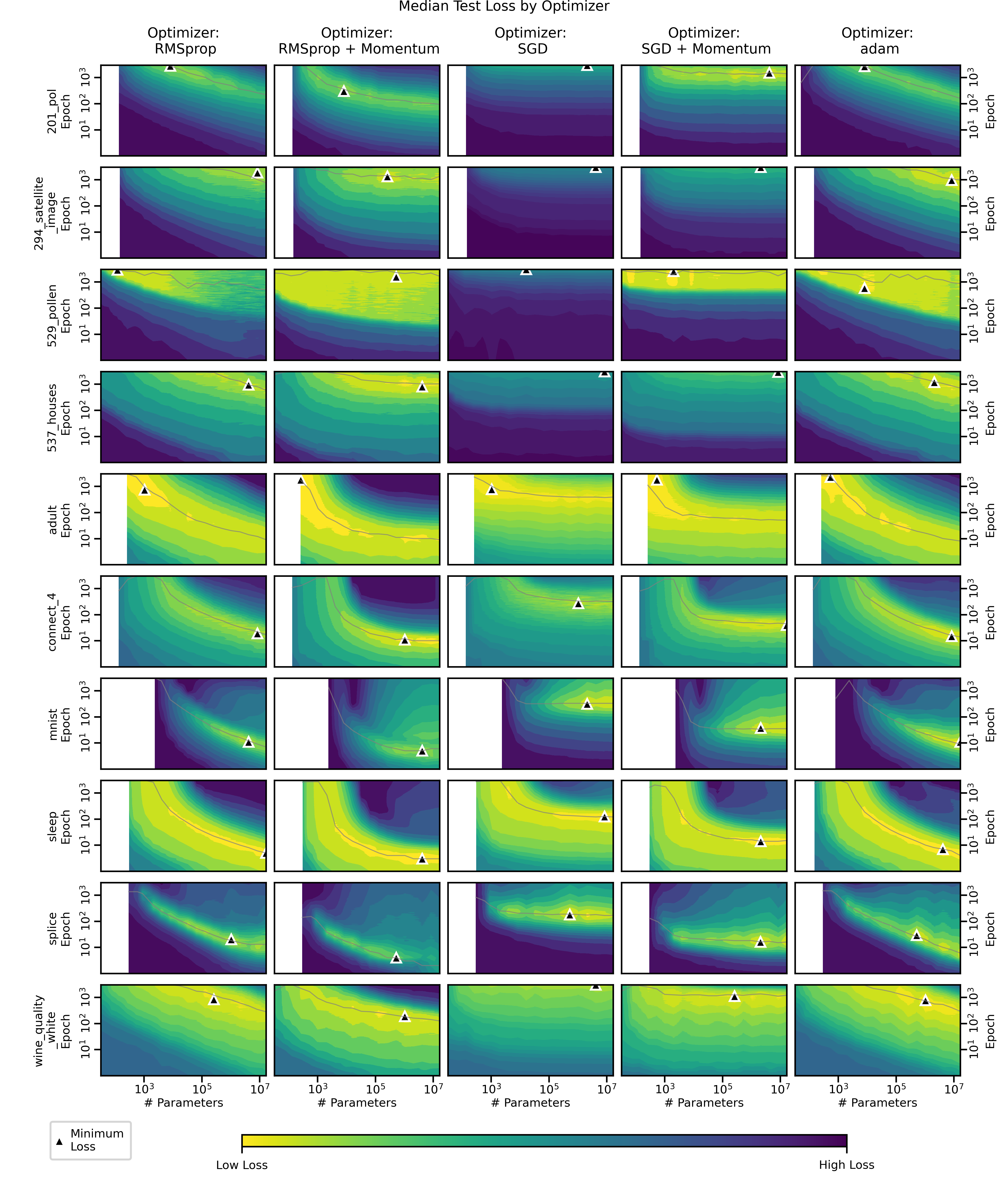}
    \caption{\textbf{Median test loss for tasks and optimizers in the Optimizer Sweep.} 
    RMSProp and SGD both perform similarly to Adam in these experiments.
    Possibly due to a static step-size, SGD has a `flatter' butter-zone beyond the interpolation threshold.
    For SGD we plotted experiments with a learning rate of $0.01$ instead of $0.001$.
    }
    \label{fig:optimizer_median}
\end{figure*}

\begin{figure*}[p]
    \centering
    \includegraphics[height=8.5in, width=7.25in]{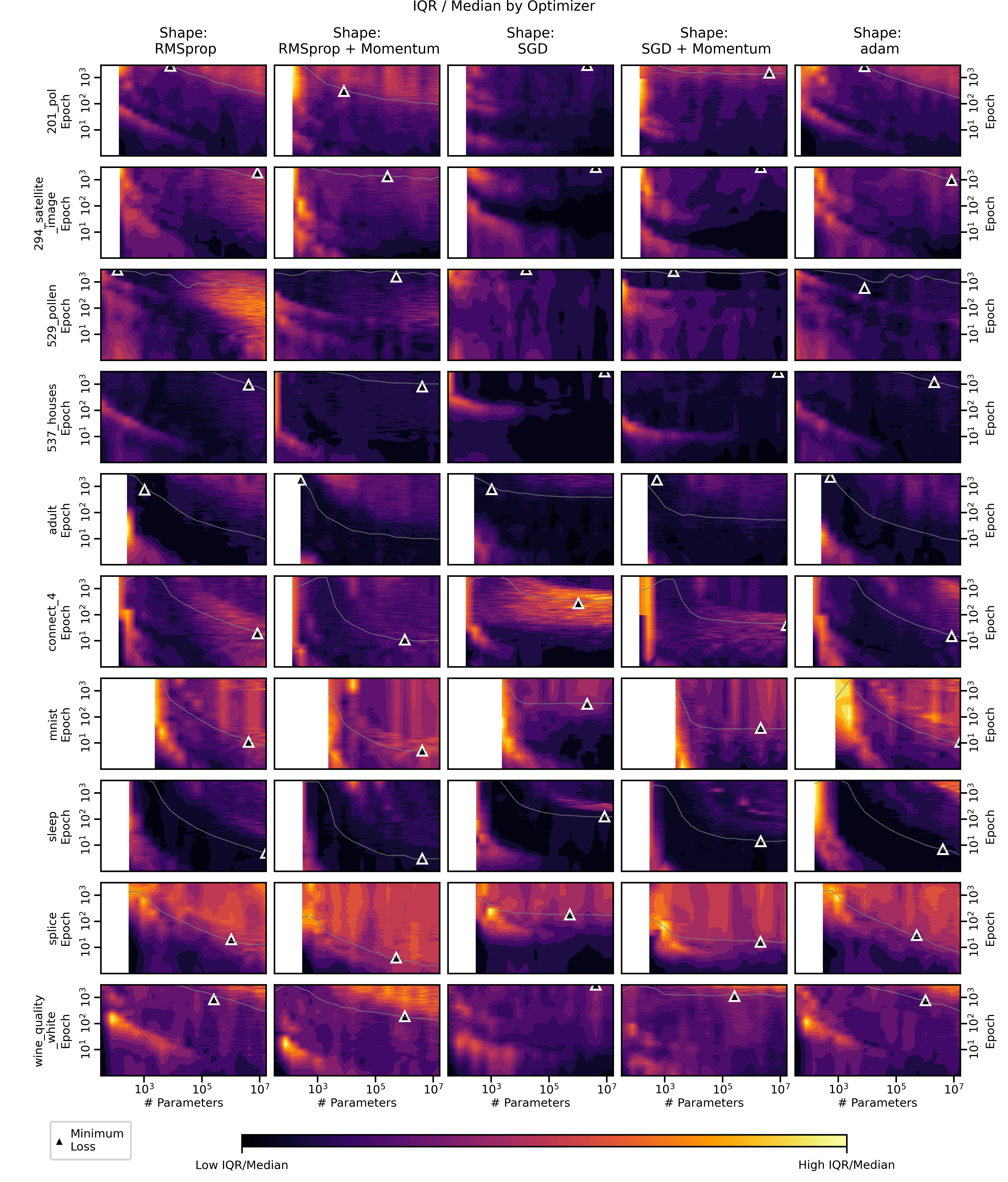}
    \caption{\textbf{Interquartile range / median for tasks and optimizers in the Optimizer Sweep.} 
    We see similar bands of variation before (strongly) and after (diffusely) the bottom of the lowest test loss region.
    Regardless of optimizer, networks sometimes learn faster and sometimes learn slower, particularly near the interpolation threshold.
    }
    \label{fig:optimizer_cov}
\end{figure*}

\begin{itemize}
\item Figures \ref{fig:shape1_median}-\ref{fig:shape2_cov} Network Shape Plots
\item Figures \ref{fig:depth1_median}-\ref{fig:depth2_cov} Network Depth Plots
\item Figures \ref{fig:lr_median}, \ref{fig:lr_cov} Learning Rate Plots
\item Figures \ref{fig:batch_size_median}, \ref{fig:batch_size_cov} Minibatch Size Plots
\item Figures \ref{fig:l1_median}, \ref{fig:l1_cov} $L^1$     Regularization Plots
\item Figures \ref{fig:l2_median}, \ref{fig:l2_cov}  $L^2$ Regularization Plots
\item Figures \ref{fig:label_noise_median}, \ref{fig:label_noise_cov}  Label Noise Plots
\item Figures \ref{fig:optimizer_median}, \ref{fig:optimizer_cov} Optimizer Comparison Plots
\end{itemize}

\FloatBarrier
\clearpage

\subsection{Dataset information}
\label{sec:dataset-information}

\noindent The BUTTER dataset is intended to provide empirical deep learning performance data to inform machine learning researchers and practitioners; to act as a guiding light for theoreticians and engineers in investigating, understanding, and furthering our knowledge of the deep learning phenomenon.

The \href{https://data.openei.org/submissions/5708}{BUTTER dataset} is permanently publicly available through the Open Energy Data Initiative (OEDI) under the \href{https://creativecommons.org/licenses/by-sa/4.0/}{Creative Commons Attribution-ShareAlike 4.0 International Public License}. \footnote{OEDI Dataset Link: \href{https://data.openei.org/submissions/5708}{data.openei.org/submissions/5708}} \footnote{Dataset License: \href{https://creativecommons.org/licenses/by-sa/4.0/}{creativecommons.org/licenses/by-sa/4.0/}}
The entire dataset can be accessed through the \href{https://data.openei.org/s3_viewer?bucket=oedi-data-lake&prefix=butter}{OEDI S3 viewer}.
\footnote{OEDI S3 Viewer:  \href{https://data.openei.org/s3_viewer?bucket=oedi-data-lake&prefix=butter}{data.openei.org/s3\_viewer?bucket=oedi-data-lake\&prefix=butter}}
A \href{https://github.com/openEDI/documentation/blob/main/BUTTER.md}{metadata description} of the storage schema and exact dataset contents is linked from the OEDI dataset page, and can be accessed directly from the OEDI GitHub repository. \footnote{Dataset Readme: \href{https://github.com/openEDI/documentation/blob/main/BUTTER.md}{github.com/openEDI/documentation/blob/main/BUTTER.md}}
Example code that generates figures from this paper using the dataset can be found in the \href{https://github.com/NREL/BUTTER-Better-Understanding-of-Training-Topologies-through-Empirical-Results}{BUTTER visualization repository} is available under the MIT license. \footnote{BUTTER visualization repository:

\href{https://github.com/NREL/BUTTER-Better-Understanding-of-Training-Topologies-through-Empirical-Results}{github.com/NREL/BUTTER-Better-Understanding-of-Training-Topologies-through-Empirical-Results}}
As described in Table \ref{tab:sweeps}, the dataset is composed of 7 hyperparameter sweeps each scanning across a grid of experimental parameters.
Table \ref{tab:shapes} enumerates the network shapes and \ref{tab:tasks} enumerates the tasks referenced in Table \ref{tab:sweeps}.
The number of repetitions listed in Table \ref{tab:sweeps} is typical, however due to overlapping sweeps, in some instances more repetitions are stored in the dataset.
For example, the Learning Rate Sweep has 20 typical repetitions, but in experiments shared with the Primary Sweep, 30 repetitions are recorded.
Additionally, a small number of experiments in a sweep have fewer than the typical number of repetitions.
The exact number of repetitions of each experiment is stored in the summary dataset, and can be calculated from the full dataset by aggregating repetitions over experiment\_id.

The dataset was accumulated from 2021 to 2022 utilizing the four computing systems listed in Table \ref{tab:hpc}.
The \href{https://github.com/NREL/BUTTER-Empirical-Deep-Learning-Experimental-Framework}{software used to accumulate the dataset}, including complete commit history is available under the \href{https://opensource.org/licenses/MIT}{MIT license}. \footnote{Experimental Framework Source:

\href{https://github.com/NREL/BUTTER-Empirical-Deep-Learning-Experimental-Framework}{https://github.com/NREL/BUTTER-Empirical-Deep-Learning-Experimental-Framework}
}
\footnote{Job Queue Source:\href{https://github.com/NREL/E-Queue-HPC}{github.com/NREL/E-Queue-HPC}}
For traceability and reproducibility, each repetition record in the dataset records the start and end time of the training run, along with the hostname, operating system, software version, git hash, Python version, and Tensorflow version used to execute it.

The BUTTER dataset does not contain any offensive or personally identifiable information.
While we have taken measures to mitigate the risk, the dataset may contain incorrect or corrupted data.
The authors bear all responsibility in case of violation of rights.

\subsection{Dataset datasheet}
\label{sec:dataset-datasheet}

\noindent This section contains a dataset datasheet as described in \cite{datasetDatasheets}.

\subsubsection{Motivation}

\textbf{For what purpose was the dataset created?}

The BUTTER dataset is intended to provide empirical deep learning performance data to inform machine learning researchers and practitioners; to act as a guiding light for theoreticians and engineers in investigating, understanding, and furthering our knowledge of the deep learning phenomenon.
We were specifically motivated by the desire to construct evidence-informed priors for efficient hyperparameter search and selection.

\textbf{Who created the dataset?}

The dataset was created by Dr. Charles Tripp, Jordan Perr-Sauer, Lucas Hayne, and Dr. Monte Lunacek as part of an internally funded machine learning project within the Computational Science Center at the National Renewable Energy Laboratory (NREL), a United States Department of Energy (DOE) National Laboratory operated by the Alliance for Sustainable Energy, LLC.

\textbf{Who funded the dataset?}

Creation of the dataset was funded using NREL laboratory directed research and development funds.

\subsubsection{Composition}

\textbf{What do the instances that comprise the dataset represent?}

As described in the \href{https://github.com/openEDI/documentation/blob/main/BUTTER.md}{dataset readme}, the raw dataset contains records of deep learning training runs, repeating each distinct experiment multiple times using different random seeds.

Each row of the full dataset represents a single repetition of an experiment (that is, a single training run). Each repetition was instrumented to record various statistics at the end of each training epoch, and those statistics are stored as epoch-indexed arrays in each row. Runs are labeled with an 'experiment\_id' which was used to aggregate repetitions of the same experimental parameters together in the summary dataset. An experiment\_id in the experiment dataset correspond to the same experiment\_id in the summary dataset.

For convenience, a summary dataset is also provided which contains statistics of these measurements aggregated over every repetition of each distinct experiment.

\textbf{How many instances are there in total?}
The full dataset contains 11.2 million total training runs, and the summary dataset contains 483 thousand distinct experiment records covering 40 billion training epochs.

\textbf{Does the dataset contain all possible instances or is it a sample of instances from a larger set?}
The dataset contains all of the training runs we executed.

\textbf{What data does each instance consist of?}
As described in the \href{https://github.com/openEDI/documentation/blob/main/BUTTER.md}{dataset readme}:

\paragraph{Data Format}
 
The complete raw dataset is available in the /all\_runs/ partitioned parquet dataset. Each row in this dataset is a record of one training repetition of a network. Several statistics were recorded at the end of each training epoch, and those records are stored in this row as arrays indexed by training epoch. For convenience, we also provide the /complete\_summary/ partitioned parquet dataset which contains statistics aggregated over all repetitions of the same experiment including average and median test and training losses at the end of each training epoch. Distinct experiments are uniquely and consistently identified in both datasets by an 'experiment\_id'. Additionally, we have created separate full (containing all repetitions) and per-experiment summary datasets for each experimental sweep so that they can be downloaded and queried separately if the entire dataset is not needed. The schemas of summary and full datasets are the same for every sweep.
 
\paragraph{File Hierarchy and Descriptions}

\begin{itemize}
    \item  \textbf{/complete\_executive\_summary/} \textit{This file is intended to provide a simple, small dataset that can be more easily and quickly downloaded, queried, and analyzed than the summary or run datasets and provides an easy starting point for using this dataset.} It contains a minimal set of per-experiment statistics aggregated over every repetition of each distinct experiment for all sweeps. This file has the same schema as the summary datasets, except it does not include any per-epoch statistic columns except for test\_loss\_q1, test\_loss\_median, and test\_loss\_q3.
    \item \textbf{/complete\_executive\_summary.tar.xz} is a compressed tarball of /complete\_executive\_summary/

    \item  \textbf{/complete\_summary/} contains per-experiment statistics aggregated over every repetition of each distinct experiment for all sweeps. This complete summary dataset is designed for easy analysis of aggregate statistics without the need to query individual repetitions. If the executive summary doesn't have everything you need (for example you want to see the KL-Divergence history for regularized experiments), the summary dataset likely is what you want to use.
    \item  /complete\_summary.tar.xz is a compressed tarball of /complete\_summary/

    \item  **/all\_repetitions/** contains all of the repetition records in all sweeps. This dataset contains all of the raw data logged during every repetition of every experiment. It is well-partitioned, but even so, it can be cumbersome to work with. Use the executive summary or summary datasets if you can. If you need the raw data, though, it's in here.
    \item  /all\_repetitions.tar is a tarball of /all\_repetitions/

    \item  /primary\_sweep\_summary/ contains summary experiment statistics for the Primary Sweep
    \item  /primary\_sweep\_summary.tar.xz is a compressed tarball of /primary\_sweep\_summary/

    \item  /300\_epoch\_sweep\_summary/ contains summary experiment statistics for the 300 epoch sweep
    \item  /300\_epoch\_sweep\_summary.tar.xz is a compressed tarball of /300\_epoch\_sweep\_summary/

    \item  /30k\_epoch\_sweep\_summary/ contains summary experiment statistics for the 30k epoch sweep
    \item  /30k\_epoch\_sweep\_summary.tar.xz is a compressed tarball of /30k\_epoch\_sweep\_summary/

    \item  /learning\_rate\_sweep\_summary/ contains summary experiment statistics for the learning rate sweep
    \item  /learning\_rate\_sweep\_summary.tar.xz is a compressed tarball of /learning\_rate\_sweep\_summary/

    \item  /label\_noise\_sweep\_summary/ contains summary experiment statistics for the label noise sweep
    \item  /label\_noise\_sweep\_summary.tar.xz is a compressed tarball of /label\_noise\_sweep\_summary/

    \item  /batch\_size\_sweep\_summary/ contains summary experiment statistics for the batch size sweep
    \item  /batch\_size\_sweep\_summary.tar.xz is a compressed tarball of /batch\_size\_sweep\_summary/

    \item  /regularization\_sweep\_summary/ contains summary experiment statistics for the regularization sweep
    \item  /regularization\_sweep\_summary.tar.xz is a compressed tarball of /regularization\_sweep\_summary/

    \item  /optimizer\_sweep\_summary/ contains summary experiment statistics for the optimizer sweep
    \item  /optimizer\_sweep\_summary.tar.xz is a compressed tarball of /optimizer\_sweep\_summary/

\end{itemize}

\paragraph{Experiment Summary Schema}
 
For preliminary analysis, we recommend using the summary dataset as it is smaller and more convenient to work with than the full repetition dataset. However, the entire record of every repetition of every experiment is stored in the full dataset, allowing other statistics to be computed from the raw data. Each experiment has a unique experiment\_id value, which matches repetition records in the runs dataset. Summary data is partitioned by dataset, shape, learning rate, batch size, kernel regularizer, label noise, depth, and number of training epochs.

\textbf{Columns Describing Each Experiment}
    \begin{itemize}
    \item  experiment\_id the unique id for this experiment
    \item  primary\_sweep: bool, true iff this experiment is part of the Primary Sweep
    \item  300\_epoch\_sweep: bool, true iff this experiment is part of the 300 epoch sweep
    \item  30k\_epoch\_sweep: bool, true iff this experiment is part of the 30k epoch sweep
    \item  learning\_rate\_sweep: bool, true iff this experiment is part of the learning rate sweep
    \item  label\_noise\_sweep: bool, true iff this experiment is part of the label noise sweep
    \item  batch\_size\_sweep: bool, true iff this experiment is part of the batch size sweep
    \item  regularization\_sweep: bool, true iff this experiment is part of the regularization sweep
    \item  optimizer\_sweep: bool, true iff this experiment is part of the optimizer sweep
    \item  activation: string, the activation function used for hidden layers
    \item  batch: string, a nickname for the experimental batch this experiment belongs to
    \item  batch\_size: uint32, minibatch size
    \item  dataset: string, name of the dataset used
    \item  depth: uint8, number of layers
    \item  early\_stopping: string, early stopping policy
    \item  epochs: uint32, number of training epochs in this experiment
    \item  input\_activation: string, input activation function
    \item  kernel\_regularizer: string, null if no regularizer is used
    \item  kernel\_regularizer\_l1: float32, L1 regularization penalty coefficient
    \item  kernel\_regularizer\_l2: float32, L2 regularization penalty coefficient
    \item  kernel\_regularizer\_type: string, name of kernel regularizer used (null if none used)
    \item  label\_noise: float32, amount of label noise applied to dataset before training (.05 means 5%
    \item  learning\_rate: float32, learning rate used
    \item  optimizer: string, name of the optimizer used
    \item  output\_activation: string, activation function for output layer
    \item  shape: string, network shape
    \item  size: uint64, approximate number of trainable parameters used
    \item  task: string, name of training task
    \item  test\_split: float32, test split proportion
    \item  num\_free\_parameters: uint64, exact number of trainable parameters
    \item  widths: [uint32], list of layer widths used
    \item  network\_structure: string, marshaled json representation of network structure used
    \item  num\_runs: uint8, number of runs aggregated in this summary record  
    \end{itemize}
\textbf{Columns Describing Overall Training Trajectories for Each Experiment:}
  \begin{itemize}
  \item  num: [uint8], number of runs aggregated in this summary record at each epoch
  \item  Test Loss Trajectory Statistics:
      \begin{itemize}
        \item  test\_loss\_num\_finite: [uint8], number of finite test losses at each epoch
        \item  test\_loss\_min: [float32], minimum test loss at each epoch
        \item  test\_loss\_q1: [float32], first quartile test loss at each epoch
        \item  test\_loss\_median: [float32], median test loss at each epoch
        \item  test\_loss\_q3: [float32], third quartile test loss at each epoch
        \item  test\_loss\_max: [float32], maximum test loss at each epoch
        \item  test\_loss\_avg: [float32], average test loss at each epoch
        \item  test\_loss\_stddev: [float32], standard deviation of the test loss at each epoch
      \end{itemize}
  \item  Training Loss Trajectory Statistics:
      \begin{itemize}
        \item  train\_loss\_num\_finite: [uint8], number of finite training losses at each epoch
        \item  train\_loss\_min: [float32], minimum training loss at each epoch
        \item  train\_loss\_q1: [float32], first quartile training loss at each epoch
        \item  train\_loss\_median: [float32], median training loss at each epoch
        \item  train\_loss\_q3: [float32], third quartile training loss at each epoch
        \item  train\_loss\_max: [float32], maximum training loss at each epoch
        \item  train\_loss\_avg: [float32], average training loss at each epoch
        \item  train\_loss\_stddev: [float32], standard deviation of training losses at each epoch
      \end{itemize}
  \item  Test Accuracy Trajectory Statistics:
      \begin{itemize}
        \item  test\_accuracy\_q1: [float32], first quartile test accuracy at each epoch
        \item  test\_accuracy\_median: [float32], median test accuracy at each epoch
        \item  test\_accuracy\_q3: [float32], third quartile test accuracy at each epoch
        \item  test\_accuracy\_avg: [float32], average test accuracy at each epoch
        \item  test\_accuracy\_stddev: [float32], test accuracy standard deviation accuracy at each epoch
      \end{itemize}
  \item  Training Accuracy Trajectory Statistics:
      \begin{itemize}
        \item  train\_accuracy\_q1: [float32], first quartile training accuracy at each epoch
        \item  train\_accuracy\_median: [float32], median training accuracy at each epoch
        \item  train\_accuracy\_q3: [float32], third quartile training accuracy at each epoch
        \item  train\_accuracy\_avg: [float32], average training accuracy at each epoch
        \item  train\_accuracy\_stddev: [float32], standard deviation of the training accuracy at each epoch
      \end{itemize}
  \item  Test Mean Squared Error (MSE) Trajectory Statistics:
        \begin{itemize}
        \item  test\_mean\_squared\_error\_q1: [float32], first quartile test MSE at each epoch
        \item  test\_mean\_squared\_error\_median: [float32], median test MSE at each epoch
        \item  test\_mean\_squared\_error\_q3: [float32], third quartile test MSE at each epoch
        \item  test\_mean\_squared\_error\_avg: [float32], test MSE at each epoch
        \item  test\_mean\_squared\_error\_stddev: [float32], test MSE standard deviation at each epoch
        \end{itemize}
  \item  Training Mean Squared Error (MSE) Trajectory Statistics:
  \begin{itemize}
    \item  train\_mean\_squared\_error\_q1: [float32], first quartile training MSE at each epoch
    \item  train\_mean\_squared\_error\_median: [float32], median training MSE at each epoch
    \item  train\_mean\_squared\_error\_q3: [float32], third quartile training MSE at each epoch
    \item  train\_mean\_squared\_error\_avg: [float32], average training MSE at each epoch
    \item  train\_mean\_squared\_error\_stddev: [float32], training MSE standard deviation at each epoch
  \end{itemize}
  \item  Test Kullback-Leibler Divergence (KL-Divergence) Trajectory Statistics:
  \begin{itemize}
    \item  test\_kullback\_leibler\_divergence\_q1: [float32], first quartile test KL-Divergence at each epoch
    \item  test\_kullback\_leibler\_divergence\_median: [float32], median test KL-Divergence at each epoch
    \item  test\_kullback\_leibler\_divergence\_q3: [float32], third quartile test KL-Divergence at each epoch
    \item  test\_kullback\_leibler\_divergence\_avg: [float32], average test KL-Divergence at each epoch
    \item  test\_kullback\_leibler\_divergence\_stddev: [float32], test KL-Divergence standard deviation at each epoch
  \end{itemize}
  \item  Training Kullback-Leibler Divergence (KL-Divergence) Trajectory Statistics:
  \begin{itemize}
    \item  train\_kullback\_leibler\_divergence\_q1: [float32], first quartile training KL-Divergence at each epoch
    \item  train\_kullback\_leibler\_divergence\_median: [float32], median training KL-Divergence at each epoch
    \item  train\_kullback\_leibler\_divergence\_q3: [float32], third quartile training KL-Divergence at each epoch
    \item  train\_kullback\_leibler\_divergence\_avg: [float32], average training KL-Divergence at each epoch
    \item  train\_kullback\_leibler\_divergence\_stddev: [float32], training KL-Divergence standard deviation at each epoch  
  \end{itemize}
  \end{itemize}
\textbf{Columns Describing Performance at the Epoch that Minimized a Statistic (e.x.: the epoch that achieved the lowest test loss)}
\begin{itemize}
  \item  Statistics about Per-Repetition Points of Minimum Test Loss
  \begin{itemize}
    \item  Distribution of the Epoch of Minimum Test Loss of each Repetition:
    \begin{itemize}
      \item  test\_loss\_min\_epoch\_min: [float32], earliest epoch at which test loss was minimized among all repetitions of the experiment
      \item  test\_loss\_min\_epoch\_q1: [float32], first quartile of the epoch at which test loss was minimized 
      \item  test\_loss\_min\_epoch\_median: [float32], median epoch at which test loss was minimized 
      \item  test\_loss\_min\_epoch\_q3: [float32], third quartile of the epoch at which test loss was minimized 
      \item  test\_loss\_min\_epoch\_max: [float32], latest epoch at which test loss was minimized among all repetitions of the experiment
      \item  test\_loss\_min\_epoch\_avg: [float32], average epoch at which test loss was minimized over all repetitions of the experiment  
    \end{itemize}
    \item  Distribution of the Value of the Minimum Test Loss of each Repetition:
    \begin{itemize}
      \item  test\_loss\_min\_value\_min: [float32], lowest minimum test loss
      \item  test\_loss\_min\_value\_q1: [float32], first quartile minimum test loss 
      \item  test\_loss\_min\_value\_median: [float32], median minimum test loss 
      \item  test\_loss\_min\_value\_q3: [float32], third quartile minimum test loss
      \item  test\_loss\_min\_value\_max: [float32], highest minimum test loss 
      \item  test\_loss\_min\_value\_avg: [float32], average minimum test loss 
    \end{itemize}
  \item  Statistics about Per-Repetition Points of Maximum Test Accuracy:
  \begin{itemize}
    \item  Distribution of the Epoch of Maximum Test Accuracy of each Repetition:
      \begin{itemize}
      \item  test\_accuracy\_max\_epoch\_min: [float32], earliest epoch at which test loss was minimized among all repetitions of the experiment
      \item  test\_accuracy\_max\_epoch\_q1: [float32], first quartile of the epoch at which test loss was minimized 
      \item  test\_accuracy\_max\_epoch\_median: [float32], median epoch at which test loss was minimized 
      \item  test\_accuracy\_max\_epoch\_q3: [float32], third quartile of the epoch at which test loss was minimized 
      \item  test\_accuracy\_max\_epoch\_max: [float32], latest epoch at which test loss was minimized among all repetitions of the experiment
      \item  test\_accuracy\_max\_epoch\_avg: [float32], average epoch at which test loss was minimized over all repetitions of the experiment
      \end{itemize}
    \item  Distribution of the Value of the Maximum Test Accuracy of each Repetition:
      \begin{itemize}
      \item  test\_accuracy\_max\_value\_min: [float32], lowest minimum test loss
      \item  test\_accuracy\_max\_value\_q1: [float32], first quartile minimum test loss 
      \item  test\_accuracy\_max\_value\_median: [float32], median minimum test loss 
      \item  test\_accuracy\_max\_value\_q3: [float32], third quartile minimum test loss 
      \item  test\_accuracy\_max\_value\_max: [float32], highest minimum test loss 
      \item  test\_accuracy\_max\_value\_avg: [float32], average minimum test loss 
      \end{itemize}
    \end{itemize}
  \item  Statistics about Per-Repetition Points of Minimum Test MSE:
  \begin{itemize}
    \item  Distribution of the Epoch of Minimum Test MSE of each Repetition:
      \begin{itemize}
      \item  test\_mean\_squared\_error\_min\_epoch\_min: [float32], earliest epoch at which test MSE was minimized among all repetitions of the experiment
      \item  test\_mean\_squared\_error\_min\_epoch\_q1: [float32], first quartile of the epoch at which test MSE was minimized 
      \item  test\_mean\_squared\_error\_min\_epoch\_median: [float32], median epoch at which test MSE was minimized 
      \item  test\_mean\_squared\_error\_min\_epoch\_q3: [float32], third quartile of the epoch at which test MSE was minimized 
      \item  test\_mean\_squared\_error\_min\_epoch\_max: [float32], latest epoch at which test MSE was minimized among all repetitions of the experiment
      \item  test\_mean\_squared\_error\_min\_epoch\_avg: [float32], average epoch at which test MSE was minimized over all repetitions of the experiment  
      \end{itemize}
    \item  Distribution of the Value of the Minimum MSE of each Repetition:
      \begin{itemize}
      \item  test\_mean\_squared\_error\_min\_value\_min: [float32], lowest minimum test MSE
      \item  test\_mean\_squared\_error\_min\_value\_q1: [float32], first quartile minimum test MSE 
      \item  test\_mean\_squared\_error\_min\_value\_median: [float32], median minimum test MSE 
      \item  test\_mean\_squared\_error\_min\_value\_q3: [float32], third quartile minimum test MSE 
      \item  test\_mean\_squared\_error\_min\_value\_max: [float32], highest minimum test MSE 
      \item  test\_mean\_squared\_error\_min\_value\_avg: [float32], average minimum test MSE 
      \end{itemize}
    \end{itemize}
  \item  Statistics about Per-Repetition Points of Minimum KL-Divergence:
  \begin{itemize}
    \item  Distribution of the Epoch of Minimum Test KL-Divergence of each Repetition:
    \begin{itemize}
      \item  test\_kullback\_leibler\_divergence\_min\_epoch\_min: [float32], earliest epoch at which test KL-divergence was minimized among all repetitions of the experiment
      \item  test\_kullback\_leibler\_divergence\_min\_epoch\_q1: [float32], first quartile of the epoch at which test KL-divergence was minimized 
      \item  test\_kullback\_leibler\_divergence\_min\_epoch\_median: [float32], median epoch at which test KL-divergence was minimized 
      \item  test\_kullback\_leibler\_divergence\_min\_epoch\_q3: [float32], third quartile of the epoch at which test KL-divergence was minimized 
      \item  test\_kullback\_leibler\_divergence\_min\_epoch\_max: [float32], latest epoch at which test KL-divergence was minimized among all repetitions of the experiment
      \item  test\_kullback\_leibler\_divergence\_min\_epoch\_avg: [float32], average epoch at which test KL-divergence was minimized over all repetitions of the experiment  
    \end{itemize}
    \item  Distribution of the Value of the Minimum KL-Divergence of each Repetition:
    \begin{itemize}
      \item  test\_kullback\_leibler\_divergence\_min\_value\_min: [float32], lowest minimum test KL-divergence
      \item  test\_kullback\_leibler\_divergence\_min\_value\_q1: [float32], first quartile minimum test KL-divergence 
      \item  test\_kullback\_leibler\_divergence\_min\_value\_median: [float32], median minimum test KL-divergence 
      \item  test\_kullback\_leibler\_divergence\_min\_value\_q3: [float32], third quartile minimum test KL-divergence 
      \item  test\_kullback\_leibler\_divergence\_min\_value\_max: [float32], highest minimum test KL-divergence 
      \item  test\_kullback\_leibler\_divergence\_min\_value\_avg: [float32], average minimum test KL-divergence 
     \end{itemize} 
    \end{itemize}
  \end{itemize}
\end{itemize}

\paragraph{Repetition Schema}
 
Each row of the repetition dataset represents a single training run. Each repetition was instrumented to record various statistics at the end of each training epoch, and those statistics are stored as epoch-indexed arrays in each row. Runs are labeled with an 'experiment\_id' which was used to aggregate repetitions of the same experimental parameters together in the summary dataset. An experiment\_id in the experiment dataset correspond to the same experiment\_id in the summary dataset.
The columns of the full dataset are:

\begin{itemize}
    \item experiment\_id: uint32, id of the experiment this run was a repetition of
    \item run\_id: string, unique id of this repetition
    \item primary\_sweep: bool, true iff this experiment is part of the Primary Sweep
    \item 300\_epoch\_sweep: bool, true iff this experiment is part of the 300 epoch sweep
    \item 30k\_epoch\_sweep: bool, true iff this experiment is part of the 30k epoch sweep
    \item learning\_rate\_sweep: bool, true iff this experiment is part of the learning rate sweep
    \item label\_noise\_sweep: bool, true iff this experiment is part of the label noise sweep
    \item batch\_size\_sweep: bool, true iff this experiment is part of the batch size sweep
    \item regularization\_sweep: bool, true iff this experiment is part of the regularization sweep
    \item optimizer\_sweep: bool, true iff this experiment is part of the optimizer sweep
    \item activation: string, the activation function used for hidden layers
    \item batch: string, a nickname for the experimental batch this experiment belongs to
    \item batch\_size: uint32, minibatch size
    \item dataset: string, name of the dataset used
    \item depth: uint8, number of layers
    \item early\_stopping: string, early stopping policy
    \item epochs: uint32, number of training epochs in this repetition
    \item input\_activation: string, input activation function
    \item kernel\_regularizer: string, null if no regularizer is used
    \item kernel\_regularizer\_l1: float32, L1 regularization penalty coefficient
    \item kernel\_regularizer\_l2: float32, L2 regularization penalty coefficient
    \item kernel\_regularizer\_type: string, name of kernel regularizer used (null if none used)
    \item label\_noise: float32, amount of label noise applied to dataset before training (.05 means 5%
    \item learning\_rate: float32, learning rate used
    \item optimizer: string, name of the optimizer used
    \item output\_activation: string, activation function for output layer
    \item python\_version: string, python version used
    \item shape: string, network shape
    \item size: uint64, approximate number of trainable parameters used
    \item task: string, name of training task
    \item task\_version: uint16, major version of training task
    \item tensorflow\_version: string, tensorflow version
    \item test\_split: float32, test split proportion
    \item test\_split\_method: string, test split method
    \item num\_free\_parameters: uint64, exact number of trainable parameters
    \item widths: [uint32], list of layer widths used
    \item network\_structure: string, marshaled json representation of network structure used
    \item platform: string, platform repetition executed on
    \item git\_hash: string, git hash of version of experimental framework used
    \item hostname: string, hostname of system that executed this repetition
    \item seed: int64, random seed used to initialize this repetition
    \item start\_time: int64, unix timestamp of start time of this repetition
    \item update\_time: int64, unix timestamp of completion time of this repetition
    \item command: string, complete marshaled json representation of this repetition's settings
    \item network\_structure: string, marshaled json representation of this repetition's network configuration
    \item widths: uint32, list of layer widths for the network used
    \item num\_free\_parameters: uint64, exact number of free parameters in the tested network
    \item val\_loss: [float32], test loss at the end of each epoch
    \item loss: [float32], training loss at the end of each epoch
    \item val\_accuracy: [float32], test accuracy at the end of each epoch
    \item accuracy: [float32], training accuracy at the end of each epoch
    \item val\_mean\_squared\_error: [float32], test MSE at the end of each epoch
    \item mean\_squared\_error: [float32], training MSE at the end of each epoch
    \item val\_mean\_absolute\_error: [float32], test MAE at the end of each epoch
    \item mean\_absolute\_error: [float32], training MAE at the end of each epoch
    \item val\_root\_mean\_squared\_error: [float32], test RMS error at the end of each epoch
    \item root\_mean\_squared\_error: [float32], training RMS error at the end of each epoch
    \item val\_mean\_squared\_logarithmic\_error: [float32], test MSLE at the end of each epoch
    \item mean\_squared\_logarithmic\_error: [float32], training MSLE at the end of each epoch
    \item val\_hinge: [float32], test hinge loss at the end of each epoch
    \item hinge: [float32], training hinge loss at the end of each epoch
    \item val\_squared\_hinge: [float32], test squared hinge loss at the end of each epoch
    \item squared\_hinge: [float32], training squared hinge loss at the end of each epoch
    \item val\_cosine\_similarity: [float32], test cosine similarity at the end of each epoch
    \item cosine\_similarity: [float32], training cosine similarity at the end of each epoch
    \item val\_kullback\_leibler\_divergence: [float32], test KL-divergence at the end of each epoch
    \item kullback\_leibler\_divergence: [float32], training KL-divergence at the end of each epoch
\end{itemize}

\textbf{Is there a label or target associated with each instance?}
Not specifically.
Each record is annotated with columns indicating which sweeps it belongs to as well as its defining parameters.

\textbf{Is any information missing from individual instances?}
No.
In some cases a parameter value is NULL; this does not indicate missing data, but that the value is not applicable to that repetition or experiment.
For example, runs with no regularization report NULL for $L^1$ regularization penalty because not only was the penalty $0$, no regularization was applied what-so-ever. 
This allows us to distinguish between the case of activating the regularization code with zero penalty (the recorded penalty would be $0$) and not activating the regularization code (the recorded penalty would be NULL).

\textbf{Are relationships between individual instances made explicit?}
Yes.
The repetition records corresponding to a summary record share the same experiment\_id.

\textbf{Are there recommended data splits?}
No.
However, we organized the dataset by sweep (see Table \ref{tab:sweeps}) through the parameter space.
Salient dimensions, such as learning rate, shape, size, depth, label noise, and regularization level can be used to split the data in useful ways.
Per-sweep slices of the summary dataset are available for download.

\textbf{Are there any errors, sources of noise, or redundancies in the
dataset?}
The dataset was collected over a period of time and multiple systems.
Each system had different hardware and software configurations and the environment might have changed between runs.
The experimental framework itself was also changed during the collection of these runs to enable greater flexibility and compatibility.
Run records include many execution environment parameters including software version numbers, the git hash of the framework version, the hostname, and the operating system used.
If differing operating environments introduced noise into the data, this metadata should help in detecting and filtering such noise or errors from the dataset.
However, we have not detected any such variations.

Experimental repetitions were repeated multiple times with varying random seeds; these matching repetitions will have the same experiment\_id values.
The number of repetitions of a given experiment is also present in its experiment summary record. This number can vary between experiments.

\textbf{Is the dataset self-contained, or does it link to or otherwise rely on
external resources?}
The dataset does not rely on external resources, but the training tasks utilized datasets from the \href{https://epistasislab.github.io/pmlb/}{Penn Machine Learning Benchmarks} collection of datasets.

\textbf{Does the dataset contain data that might be considered confidential?}
No.

\textbf{Does the dataset contain data that, if viewed directly, might be offensive, insulting, threatening, or might otherwise cause anxiety?}
No.

\subsubsection{Collection process}

\textbf{How was the data associated with each instance acquired?}
The worker process which executed the experiment recorded the data for that repetition in the database using the same unique id associated with the repetition task stored in the task queue.
That same unique id is also stored in each run record of the dataset.

\textbf{What mechanisms or procedures were used to collect the data?}
Our \href{https://github.com/NREL/BUTTER-Empirical-Deep-Learning-Experimental-Framework}{experimental framework} was used to collect the data using the systems listed in Table \ref{tab:hpc}.

\textbf{If the dataset is a sample from a larger set, what was the sampling strategy?} The data was not sampled from a larger set.

\textbf{Who was involved in the data collection process and how were they compensated?}
The authors, as paid employees of the Alliance for Sustainable Energy, LLC collected the data.

\textbf{Over what timeframe was the data collected?}
The data was collected from 2021 to 2022. 
Collection start and end dates for each run are recorded in the dataset.

\textbf{Were any ethical review processes conducted?}
No.

\subsubsection{Preprocessing/cleaning/labeling}

\textbf{Was any preprocessing/cleaning/labeling of the data done?}
Runs and summary records were annotated with sweep set membership flags.
No preprocessing or cleaning was done to the run table.

\textbf{Was the “raw” data saved in addition to the preprocessed/cleaned/labeled data?}
The full dataset is composed primarily of the raw data which is unmodified and unfiltered.

\textbf{Is the software that was used to preprocess/clean/label the data available?}
No. 
Sweep labeling was accomplished using simple database queries which marked experiments based on set membership of the parameters of each experiment in each sweep. 
Sweep labels are derived directly from the raw data and are provided only for convenience; they are not a fundamental component of the dataset.

\subsubsection{Uses}

\textbf{Has the dataset been used for any tasks already?}
No.

\textbf{Is there a repository that links to any or all papers or systems that use the dataset?}
No. However, additional resources can be found on the Framework's Open Energy Data Initiative (OEDI) Record \cite{BUTTERDataset} and the Framework's DOE Code record \cite{EmpiricalFramework}.

\textbf{What (other) tasks could the dataset be used for?}
Informing hyperparameter initialization and search is one possibility.

\textbf{Is there anything about the composition of the dataset or the way it was collected and preprocessed/cleaned/labeled that might impact future uses?}
No.

\textbf{Are there tasks for which the dataset should not be used?}
The dataset should not be used in an unethical or unreasonable way.
The dataset only tests a finite set of hyperparameter combinations and training tasks.
Therefore, generalizing findings from this dataset beyond the bounds of the specific experiments recorded in it must be done with care and an acknowledgment that such extrapolations lie beneath a veil of speculative uncertainty.

\subsubsection{Distribution}

\textbf{Will the dataset be distributed to third parties outside of the entity on behalf of which the dataset was created?}
Yes. It is publicly available.

\textbf{How will the dataset will be distributed?}
The dataset is posted as part of the Open Energy Data Initiative (OEDI) dataset repository: \href{https://data.openei.org/submissions/5708}{data.openei.org/submissions/5708} \cite{BUTTERDataset}.

\textbf{When will the dataset be distributed?}
It is currently available.

\textbf{Will the dataset be distributed under a copyright or other intellectual
property (IP) license, and/or under applicable terms of use (ToU)?}
The dataset is distributed under the \href{https://creativecommons.org/licenses/by-sa/4.0/}{Creative Commons Attribution-ShareAlike 4.0 International Public License}. \footnote{Dataset License: \href{https://creativecommons.org/licenses/by-sa/4.0/}{creativecommons.org/licenses/by-sa/4.0/}}.

\textbf{Have any third parties imposed IP-based or other restrictions on the
data associated with the instances?}
No.

\textbf{Do any export controls or other regulatory restrictions apply to the
dataset or to individual instances?}
No.

\subsubsection{Maintenance}

\textbf{Who will be supporting/hosting/maintaining the dataset?}
Dr. Charles Tripp will maintain the dataset.
It will be hosted by the United States Department of Energy (DOE) Open Energy Data Initiative (OEDI).

\textbf{How can the owner/curator/manager of the dataset be contacted?}
Dr. Charles Tripp can be contacted at \texttt{charles.tripp@nrel.gov}.

\textbf{Is there an erratum?}
\vskip-\parskip
Eratta and details are available here:
\href{https://github.com/openEDI/documentation/blob/main/BUTTER.md}{github.com/openEDI/documentation/blob/main/BUTTER.md}.

\textbf{Will the dataset be updated?}
Updates will be posted in the dataset readme.

\textbf{Will older versions of the dataset continue to be supported/hosted/maintained?}
If updated, older versions will remain accessible but may not be maintained.
The versioning system and how to access previous versions will be described in the readme when new versions are released.

\textbf{If others want to extend/augment/build on/contribute to the dataset, is there a mechanism for them to do so?}
Please contact Charles Tripp, \texttt{charles.tripp@nrel.gov}, if you would like to augment the dataset.
Any additions will be replicable, verified for consistency, and include instructions and software for doing so.

\end{document}